\documentclass[11pt]{article}

\usepackage[top=1in, bottom=1in, left=0.75in, right=0.75in]{geometry}

\usepackage{amssymb}
\usepackage{bm}
\usepackage{amsmath}
\usepackage[ruled]{algorithm2e}
\usepackage[usenames,dvipsnames]{color}
\usepackage{array}
\usepackage{subcaption}
\usepackage{graphicx}
\usepackage{tikz}
\usetikzlibrary{shapes, arrows.meta, positioning, calc}
\usepackage{booktabs}
\usepackage{float}
\usepackage{caption}
\usepackage{placeins}
\usepackage{longtable}
\usepackage{natbib}
\usepackage{tabularx}

\usepackage{hyperref}
\hypersetup{
    colorlinks,
    linkcolor={magenta},
    citecolor={blue},
    urlcolor={blue!80!black},
    breaklinks=true,
    plainpages=true
}

\newcommand{\cref}[3]{\hyperref[#2]{#1~\ref*{#2}{#3}}}
\newcommand{\crefs}[3]{\hyperref[#2]{#1~\ref*{#2}-\ref*{#3}}}
\newcommand{\colref}[2]{\hyperref[#2]{#1~\ref*{#2}}}
\newcommand{\eqnref}[1]{\colref{Eq.}{#1}}
\newcommand{\figref}[1]{\colref{Figure}{#1}}

\newcommand{\secref}[1]{\colref{Section}{#1}}

\newcommand{\tabref}[1]{\colref{Table}{#1}}

\newcommand{\doi}[1]{\textsc{doi}: \href{http://dx.doi.org/#1}{\nolinkurl{#1}}}

\title{Procedural Generation of 3D Maize Plant Architecture from LIDAR Data}

\author{
    Mozhgan Hadadi\textsuperscript{1}, Mehdi Saraeian\textsuperscript{1}, Jackson Godbersen\textsuperscript{1}, \\
    Talukder Jubery\textsuperscript{3}, Yawei Li\textsuperscript{2}, Lakshmi Attigala\textsuperscript{2}, \\
    Aditya Balu\textsuperscript{3}, Soumik Sarkar\textsuperscript{1,3,4}, Patrick S. Schnable\textsuperscript{2,4}, \\
    Adarsh Krishnamurthy\textsuperscript{1,3,4}, Baskar Ganapathysubramanian\textsuperscript{1,3,4} \\
}

\date{
    \textsuperscript{1}Department of Mechanical Engineering, Iowa State University, Ames, IA, USA \\
    \textsuperscript{2}Department of Agronomy, Iowa State University, Ames, IA, USA \\
    \textsuperscript{3}Translational AI Center, Iowa State University, Ames, IA, USA \\
    \textsuperscript{4}Plant Science Institute, Iowa State University, Ames, IA, USA \\
}

\begin{document}
\maketitle

\begin{abstract}
This study introduces a robust framework for generating procedural 3D models of maize (\textit{Zea mays}) plants from LiDAR point cloud data, offering a scalable alternative to traditional field-based phenotyping. Our framework leverages Non-Uniform Rational B-Spline (NURBS) surfaces to model the leaves of maize plants, combining Particle Swarm Optimization (PSO) for an initial approximation of the surface and a differentiable programming framework for precise refinement of the surface to fit the point cloud data. In the first optimization phase, PSO generates an approximate NURBS surface by optimizing its control points, aligning the surface with the LiDAR data, and providing a reliable starting point for refinement. The second phase uses NURBS-Diff, a differentiable programming framework, to enhance the accuracy of the initial fit by refining the surface geometry and capturing intricate leaf details. Our results demonstrate that, while PSO establishes a robust initial fit, the integration of differentiable NURBS significantly improves the overall quality and fidelity of the reconstructed surface. This hierarchical optimization strategy enables accurate 3D reconstruction of maize leaves across diverse genotypes, facilitating the subsequent extraction of complex traits like phyllotaxy. We demonstrate our approach on diverse genotypes of field-grown maize plants. All our codes are open-source to democratize these phenotyping approaches.
\end{abstract}


\begin{figure}[t]
    \centering
    \includegraphics[trim=5cm 0cm 4cm 0cm, clip, width=0.6\linewidth]{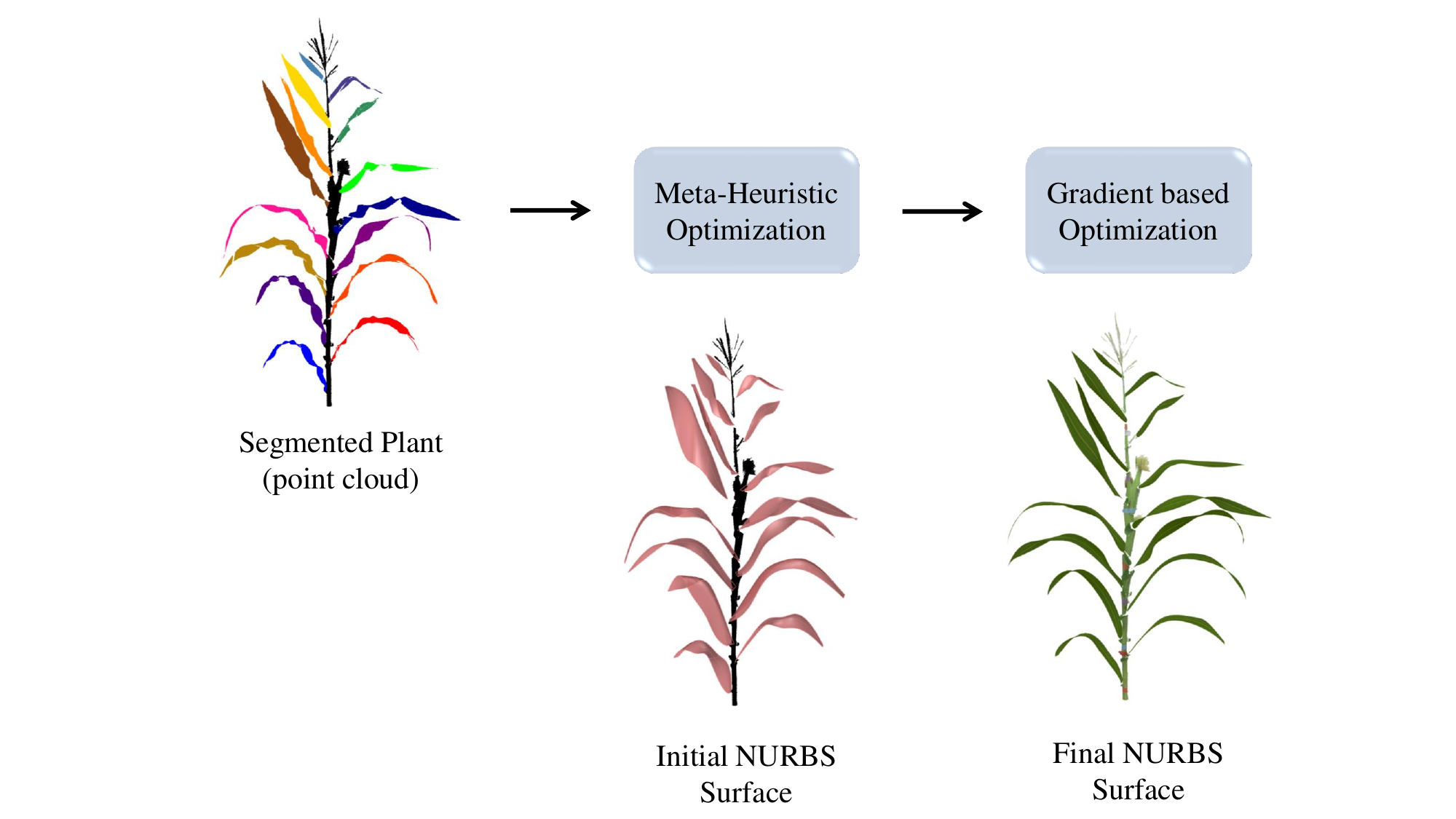}
    \caption{The workflow outlining the proposed two-step optimization process for reconstructing maize leaves from LiDAR point cloud data. The first phase involves a meta-heuristic optimization (PSO) to initialize a NURBS surface, which captures the overall shape and structure of the leaves. The second phase applies a gradient-based optimization (NURBS-Diff) to refine the surface and achieve higher fidelity, particularly in capturing intricate details along the leaf edges and tips.} 
    \label{fig:project_objective}
\end{figure}

\section{Introduction}
\label{sec:intro}

Point clouds serve as a versatile representation of 3D data, encapsulating the spatial structure of objects as a collection of unstructured points in Euclidean space, often enriched with surface normal information. These data are commonly produced by geometric acquisition methods such as laser-based scanners (LiDAR, Light Detection and Ranging) and passive methods such as multi-view stereo imaging~\citep{berger2017survey}. In agriculture, point clouds are increasingly being used for plant phenotyping to analyze traits such as leaf morphology, biomass, plant height, and canopy structure. Techniques including terrestrial laser scanning, airborne LiDAR, and Structure from Motion~\citep{westoby2012structure, jay2015field, hasheminasab2020gnss} allow for the detailed capture of 3D point clouds at the plot and plant scales, which are used to monitor crop health and growth dynamics~\citep{Photoneo2024, fernandez2022estimation}. These advances have significantly enhanced the ability to acquire rich 3D data for downstream modeling and analysis, which form the basis for many applications in diverse scientific fields ranging from reverse engineering to environmental mapping~\citep{Qin2023}.

Despite their utility, point clouds remain a raw and unstructured representation, which poses challenges for direct analysis or additional downstream modeling and simulation. Extracting meaningful phenotypic traits from plants such as leaf area, stem diameter, canopy architecture, or canopy volume requires a structured representation that can capture the intricate geometry of plants. However, point clouds often suffer from issues such as incomplete data, occlusions, noise~\citep{Guerrero2018}, and sparsity~\citep{Li2019}. These challenges complicate the accurate measurement of plant traits, which are critical for understanding plant health, productivity, and responses to environmental conditions.

To address these limitations and facilitate more sophisticated analyses, reconstructing coherent 3D models (essentially computer-aided design or CAD models) from point cloud data has become an essential step in modern phenotyping workflows. For instance, \citet{Sodhi2018} attempted to address these issues with a model-based optimization framework for phenotyping plant structures from noisy data. Similarly, \citet{Ando2021} proposed a surface reconstruction method specifically designed for plant leaves. Accurate 3D reconstructions are critical for extracting meaningful crop traits, such as biomass estimation, canopy volume analysis, and creating digital twins of plants for simulation purposes~\citep{Benes2023Roots}. 3D models, derived from point cloud data, offer structured and analyzable representations of three-dimensional objects or environments, addressing many of the limitations inherent in raw point clouds. These models serve as a bridge between unstructured spatial data and actionable insights, enabling more sophisticated analyses and simulations. They are widely used in various domains, including engineering, architecture, and plant science~\citep{Remondino2006, Paulus2019, bomer20243d, harandi2023make}. In agriculture, 3D models are essential for studying crop morphology, analyzing growth patterns, and simulating environmental interactions~\citep{gibbs2016approaches, okura20223d}. They enable detailed visualization, realistic simulations, and cost-effective experimentation~\citep{Sun2023}. Such 3D models offer several advantages over raw point clouds: 

\begin{enumerate} 
\item \textbf{Enhanced trait extraction}: 3D models enable straightforward, simple and precise measurements of traits such as leaf angles and their distribution within a plant, internode lengths, and biomass estimation, which are crucial for phenotyping and optimizing crop improvement. 
\item \textbf{Improved visualization}: Reconstructed 3D models provide an intuitive representation of plant structures, aiding the interpretation of complex morphological features. 
\item \textbf{Simulation capabilities}: Accurate 3D models support advanced simulations of plant-environment interactions, including light interception, water use, and nutrient uptake. This is the central thesis of a thriving sub-field in plant sciences called Functional-structural plant modeling (FSPM) \cite{vos2007functional,sievanen2014functional,louarn2020two,evers2018computational}.  
\item \textbf{Temporal analysis}: Reconstructing models from time-series data facilitates the tracking of plant growth and development with greater precision (again, a key consideration in FSPM). 
\item \textbf{Data compression}: Structured 3D models offer a more compact representation compared to raw point clouds, enabling efficient storage and processing of large-scale phenotyping datasets.  \end{enumerate}

Generating accurate 3D models from agriculturally relevant point clouds is a non-trivial task. Various reconstruction methods have been developed to address these issues so far~\citep{Arshad2022}, including learning-based methods~\citep{Williams2019, ben2019nesti, Guerrero2018, Chen1992, Yu1999, Junior2004, Lim2013, Boulch2016, Lee2017, Lenssen2020, Groueix2018, Park2019, Mescheder2019, Atzmon2020, yuan2022ssrnet}, traditional approximation~\citep{tang2018multi, Hoppe1992, Curless1996, Calakli2011, Carr2001, Liu2020, Ohtake2005, Kazhdan2006, Kazhdan2013, Manson2008, Ren2018, Galvez2008, Galvez2012, Lim2014, Iglesias2018, Yang2008} and interpolation techniques~\citep{Amenta1998, Amenta2000, Amenta2001, Dey2001, Dey2003, Dey2011, Boltcheva2017}. In addition, soft computing techniques have been introduced to enhance reconstruction performance~\citep{Galvez2008, Galvez2012, Iglesias2018, Forkan2008, Wang2019, Wang2019cuckoo, Wang2020}. Recently, learning-based methods, particularly deep learning approaches, have shown promise in reconstructing 3D surfaces. These methods decode point clouds into explicit shapes~\citep{Groueix2018, Luo2021, Sharp2020} or implicit fields, where zero-level isosurfaces are extracted~\citep{Park2019, Mescheder2019, Jiang2020}. Although effective for many applications, deep learning methods often (still) struggle to generalize when reconstructing complex agricultural structures~\citep{Huang2024}. 

A complementary strategy, called procedural 3D modeling, which uses algorithms and rules to generate 3D structures automatically, offers significant advantages over traditional manual modeling approaches. These techniques enable the efficient creation of complex models, flexibility through parameter adjustments, scalability for large-scale simulations, and consistency across generated models~\citep{Fischer2021}. In agriculture, procedural modeling has been used to simulate tree canopies and root systems, aiding the study of plant-environment interactions under varying conditions~\citep{Benes2023Roots}. In recent years, advancements in functional–structural plant models (FSPMs) have provided powerful tools for understanding crop morphology and simulating plant-environment interactions. For maize, FSPMs such as the 3DPhytomer-based reconstruction framework enable detailed trait analysis, including leaf angles and phyllotaxy, while offering robust simulations of plant growth dynamics under varying environmental conditions~\citep{3DPhytomerBased}. Similarly, procedural approaches like CropCraft leverage inverse procedural modeling combined with neural radiance fields and Bayesian optimization to reconstruct realistic 3D crop canopies, addressing challenges like occlusion and morphological variability~\citep{CropCraft}. Additionally, the broader application of FSPMs, as discussed in functional–structural plant modeling literature, integrates physiological processes with 3D structural models to enhance crop production simulations, including light interception and carbon partitioning~\citep{FunctionalStructuralModeling}.

The 3D modeling approaches mentioned above have emerged as vital tools in many domains of engineering design, especially precision agriculture~\citep{Zarei2024}. 3D computational models of plants could provide a viable, cost-effective alternative, especially for evaluating what-if scenarios and for extracting diverse traits. However, there is currently a lack of streamlined, automated frameworks that efficiently and automatically convert raw point cloud data to precise, modifiable 3D plant geometries. Our work addresses this challenge by developing a procedural model fitting approach to automatically fit maize plant models to LiDAR-scanned point cloud data. Specifically, we use Non-Uniform Rational B-Spline (NURBS) surfaces to represent maize leaves and employ a two-step optimization process combining Particle Swarm Optimization (PSO) with the differentiable NURBS framework (NURBS-Diff). More details about NURBS are provided in \secref{subsec:nurbs-diff}. This pipeline bridges an existing gap while also providing precise control over plant structures, enabling detailed analysis, simulation, and adaptation to diverse genotypes. An example of the resulting procedural 3D model is shown in \figref{fig:project_objective}. 

The main contributions of this work are as follows:
\begin{enumerate}
    \item An automated procedural framework for generating accurate 3D models of maize plant architecture (specifically leaves) from point cloud data, addressing scalability and genotype adaptability.
    \item A two-step optimization process that combines PSO with NURBS-Diff to fit NURBS surfaces to leaf point clouds. This two-step approach allows fast fitting of the model.    
    \item A robust and flexible framework for creating accurate 3D models of maize plants that can be easily adjusted to match different genotypes with minimal user intervention. We illustrate by fitting individualized procedural models to a diverse set of genotypes. 
\end{enumerate}

The rest of the paper is organized as follows: In \secref{subsec:dp}, we describe the dataset used in this study, including the acquisition of 3D maize point clouds using LiDAR scanning and the segmentation of point clouds into individual leaves and stalks. \secref{sec:methods} details our two-step optimization framework, combining PSO and NURBS-Diff for reconstructing precise 3D maize leaf surfaces. This section includes a comprehensive explanation of NURBS, the optimization process, and the loss functions used for surface refinement. \secref{sec:results} presents the results of our method, demonstrating its accuracy and robustness across diverse maize genotypes. It includes quantitative metrics such as Chamfer distance reduction and visual comparisons of reconstructed surfaces. Finally, in \secref{sec:conclusion}, we summarize the contributions of this study and discuss potential avenues for future research. We provide additional visual examples of reconstructed maize genotypes in the Appendix to further illustrate the performance of the proposed method.

\clearpage

\section{Dataset}
\label{subsec:dp}

The maize plants used in this study were grown during the summer of 2021 on Iowa State University’s Curtis Farm in Ames, IA under agronomically relevant crop management. Plants representing distinct genotypes from the SAM (Shoot Apical Meristem) diversity panel~\citep{thompson2015diversity, leiboff2015genetic} were cut at ground level and harvested from the field around the time of anthesis and brought indoors for LiDAR imaging. The 3D point cloud data of diverse genotypes were obtained using a Faro Focus S350 Scanner. With a 10 m scanning range and an angular resolution of 0.011 degrees, this high-precision scanner produces a point spacing of 1.5 mm. At a rate of one million points per second, the scanner can record up to 700 million points.

\begin{figure}[h!]
    \centering

    \begin{subfigure}[b]{0.20\linewidth}
        \centering
        \includegraphics[trim=0cm 7cm 0cm 2.1cm, clip, width=0.99\linewidth]{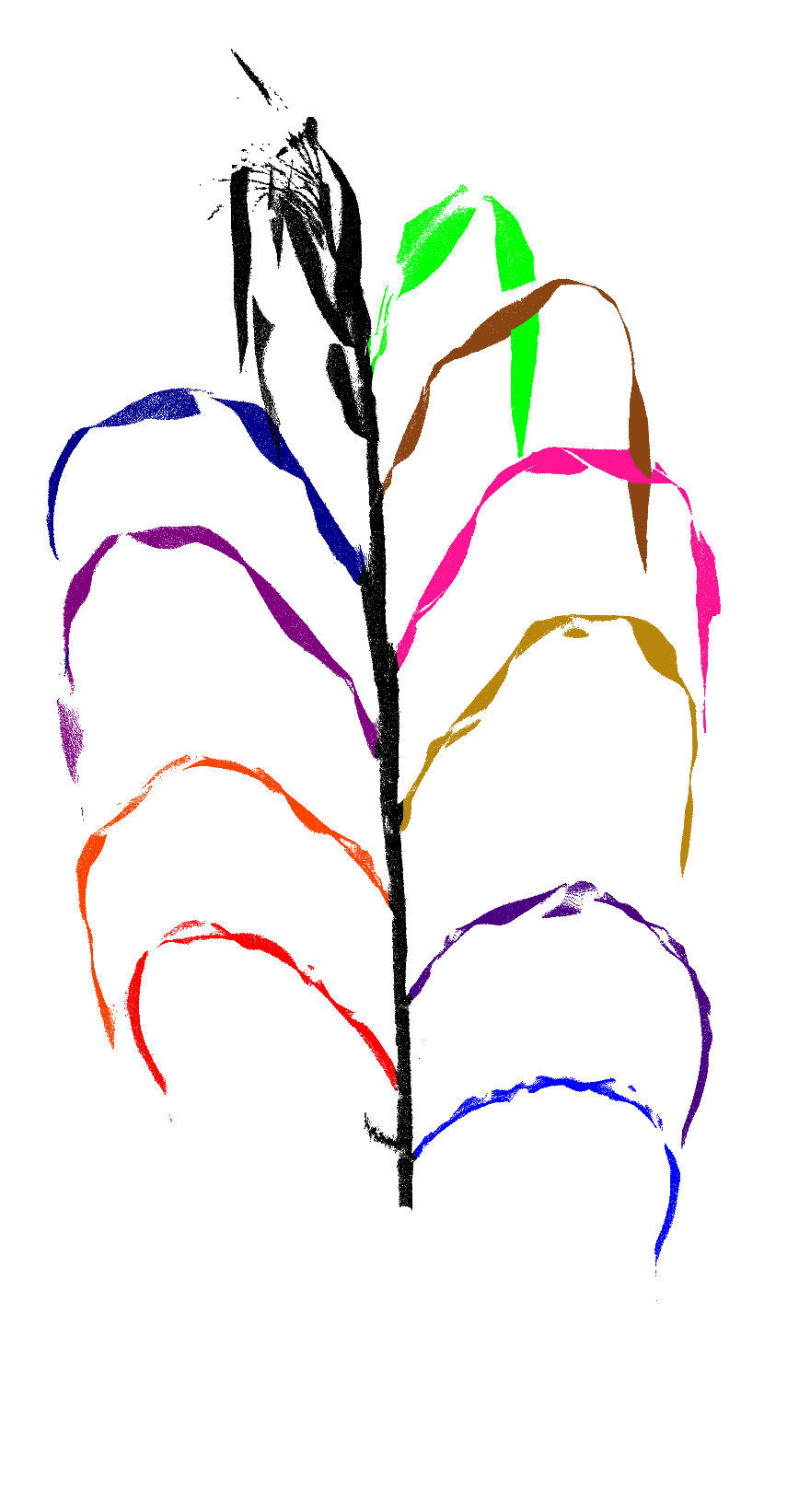}
        \caption{CML238}
    \end{subfigure}
    \begin{subfigure}[b]{0.20\linewidth}
        \centering
        \includegraphics[trim=2cm 9cm 2cm 2.1cm, clip, width=0.99\linewidth]{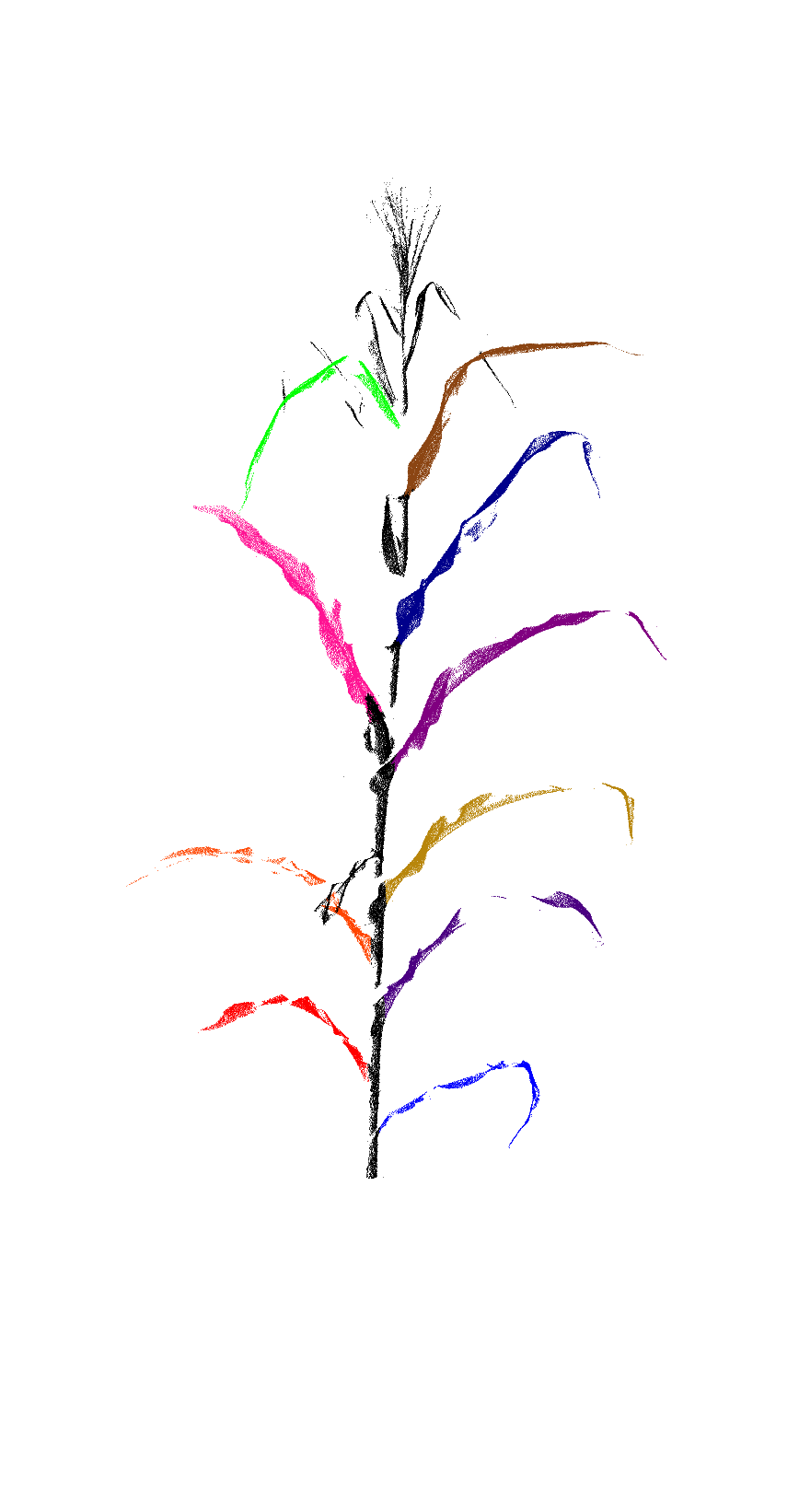}
        \caption{T8}
    \end{subfigure}
    \begin{subfigure}[b]{0.20\linewidth}
        \centering
        \includegraphics[trim=2cm 7cm 2cm 10cm, clip, width=0.99\linewidth]{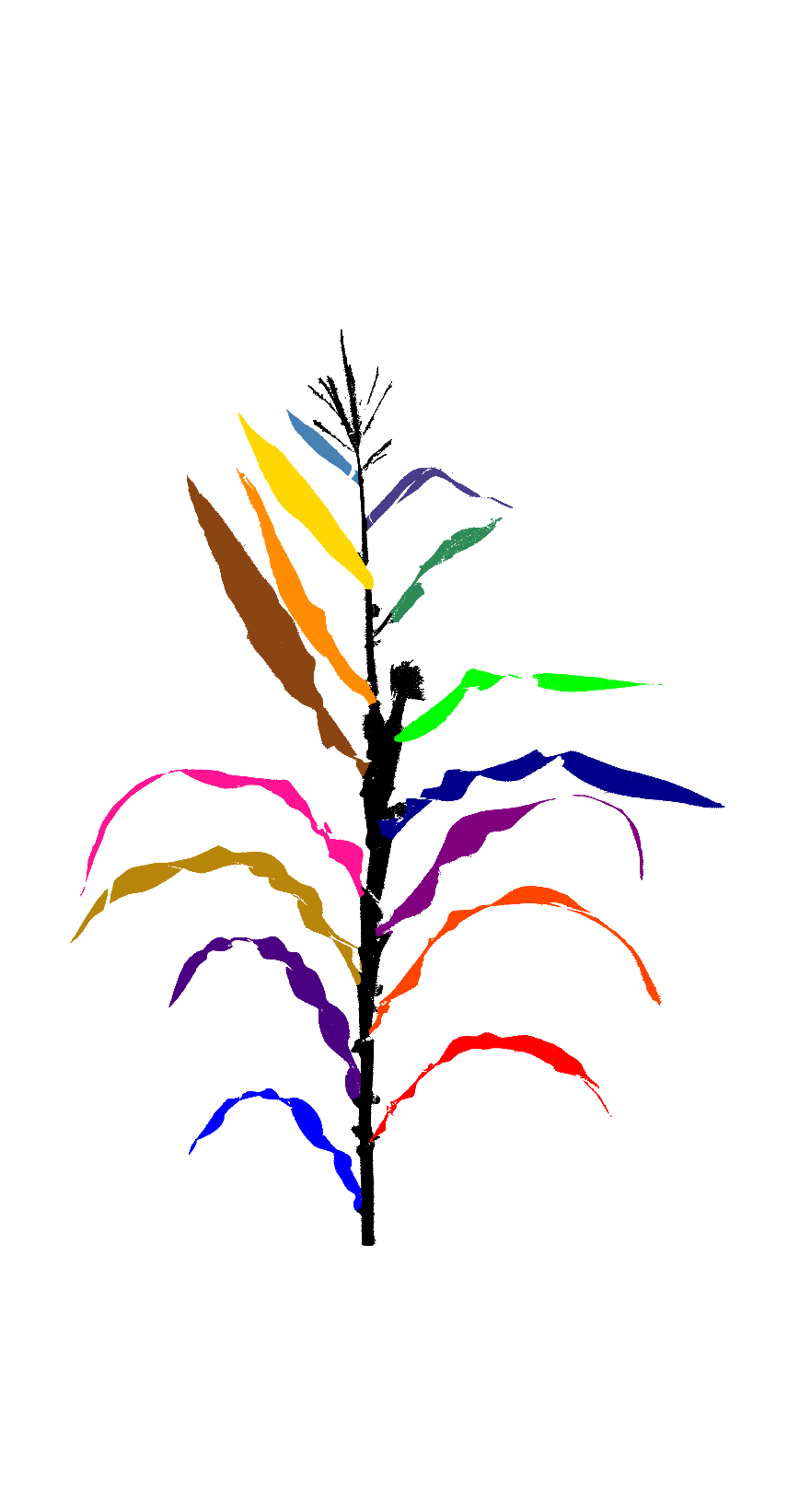}
        \caption{M162W}
    \end{subfigure}
    \begin{subfigure}[b]{0.20\linewidth}
        \centering
        \includegraphics[trim=2cm 3cm -2cm 7cm, clip, width=0.99\linewidth]{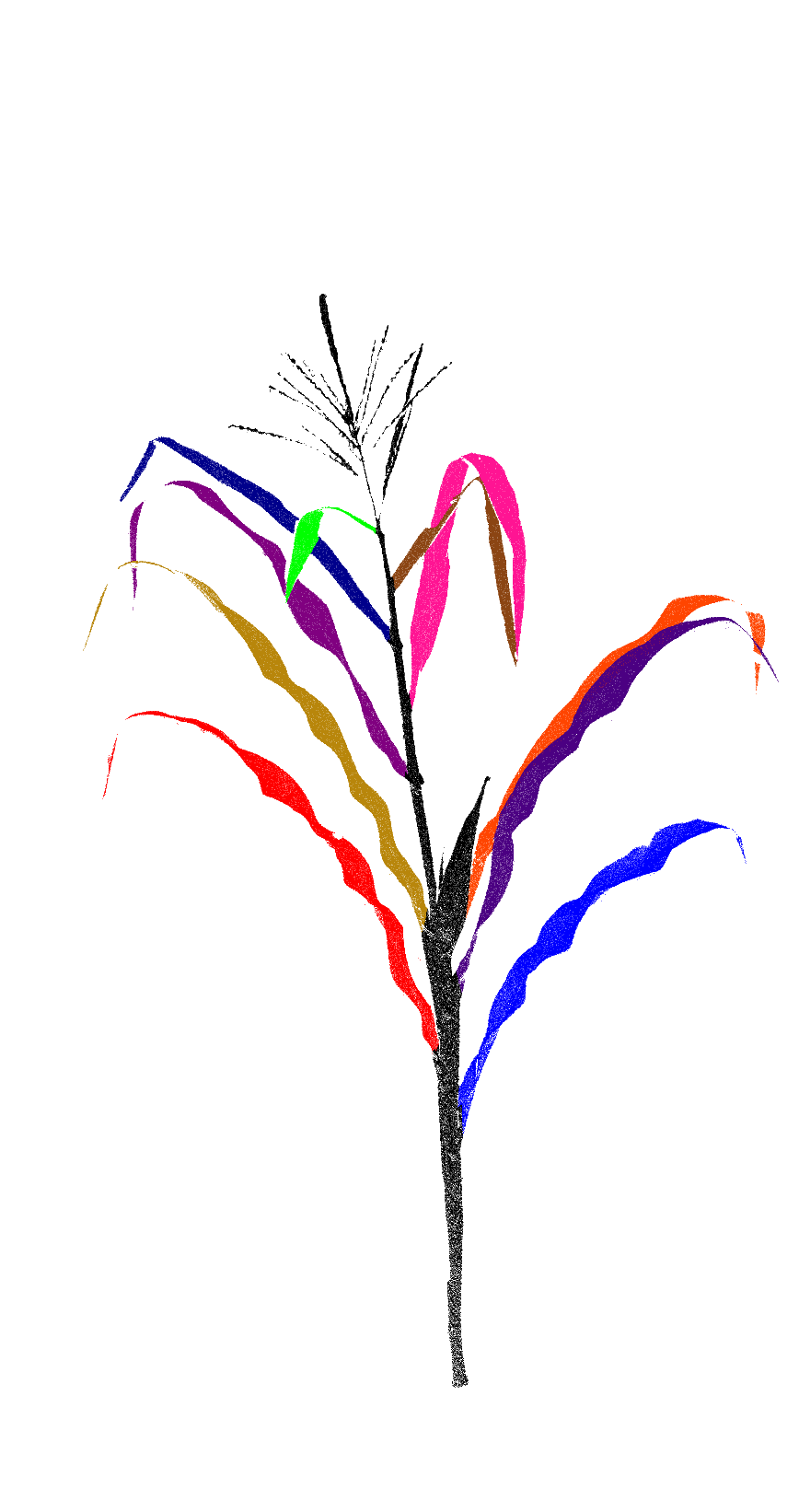}
        \caption{CI90C}
    \end{subfigure}
    \caption{LiDAR point cloud data visualized for four maize genotypes. These point clouds capture the intricate structural details of each genotype and form the basis for reconstructing 3D maize leaf models.}
    \label{fig:segmented_GT}
\end{figure}

Each point cloud data set associated with an individual plant was visually checked for quality using Cloud Compare (v2.13.2, 2024). Individual point clouds were segmented to separate leaves and stalks. \figref{fig:segmented_GT} showcases four diverse genotypes from the dataset, highlighting the segmented plant point clouds with leaves and stalks color-coded for clarity.

\section{Methods}
\label{sec:methods}

\begin{figure}[b!]
    \centering
    \includegraphics[trim=0cm 6cm 0cm 6cm, clip, width=0.7\linewidth]{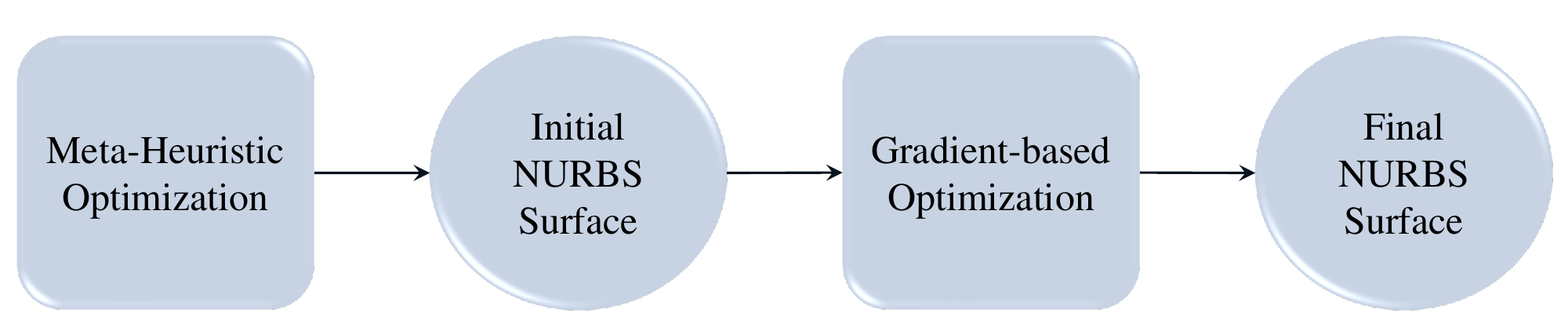}
    \caption{Schematic representation of the optimization workflow used for fitting NURBS surfaces to maize leaf point clouds. The process begins with an initial approximation using PSO, followed by refinement using NURBS-Diff to generate accurate and smooth 3D reconstructions.} 
    \label{fig:Workflow_diagram}
\end{figure}

Our approach establishes an automated workflow that progresses from input point cloud data to the final output of fitted NURBS surfaces. This workflow employs a two-step optimization method. First, we use particle swarm optimization (PSO) as a meta-heuristic approach to estimate the NURBS parameters (control points) that provide an initial approximation of the leaf shape. Subsequently, NURBS-Diff utilizes gradient-based optimization to refine this surface, achieving precise alignment with the point cloud data (\figref{fig:Workflow_diagram}). This automated pipeline enables fast, accurate modeling of maize plant leaves, minimizing user intervention and ensuring efficient and reliable surface fitting. This section outlines our pipeline, emphasizing its capability to handle unstructured point cloud data and produce high-fidelity NURBS surfaces.

\subsection{Overview of Non-Uniform Rational B-Splines (NURBS)}

\begin{figure}[t!]
    \centering
    \begin{subfigure}[b]{0.48\linewidth}
    \includegraphics[trim=8cm 3cm 8cm 3cm, clip, width=0.95\linewidth]{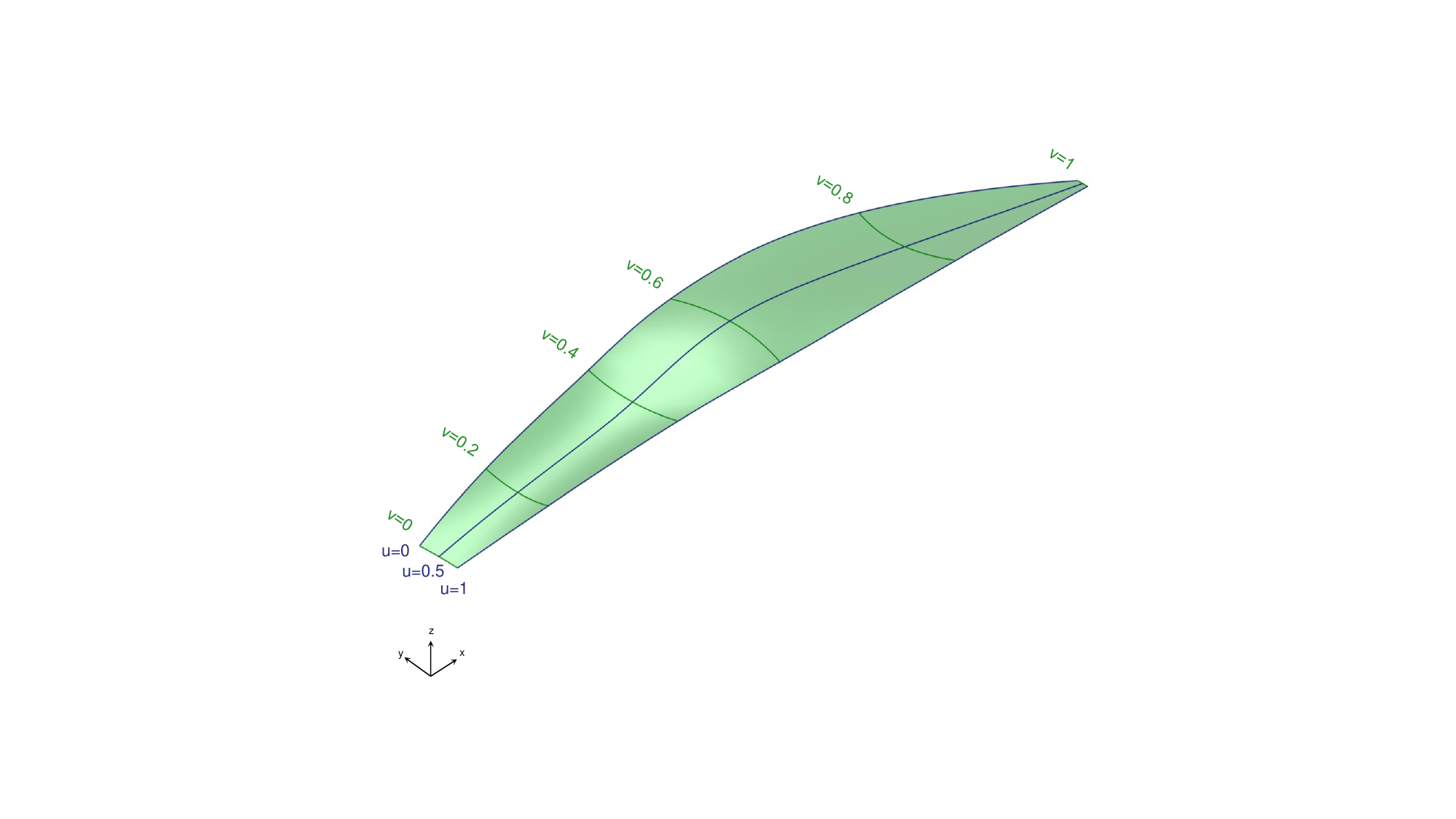} 
    \caption{}
    \label{fig:surface_param_a}
    \end{subfigure}
    \begin{subfigure}[b]{0.48\linewidth}
    \includegraphics[trim=8cm 3cm 8cm 4cm, clip, width=0.95\linewidth]{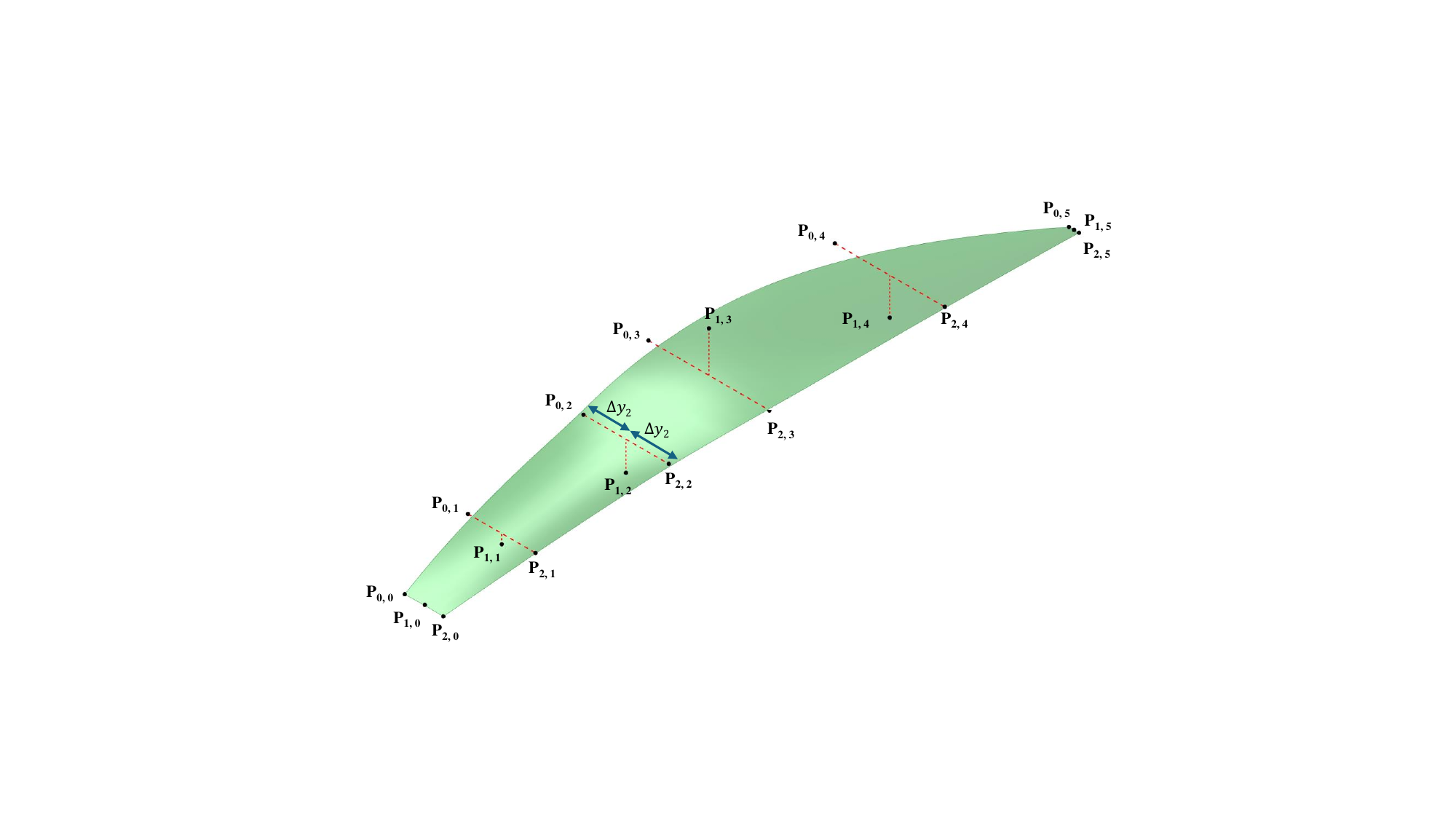} 
    \caption{}
    \label{fig:surface_param_b}
    \end{subfigure}
    \caption{Visualization of the NURBS surface parameters, including the control points ($P$), knot vectors ($U$ and $V$), and associated values that define the leaf's curvature and tapering. These parameters are optimized to replicate the natural shape and structure of maize leaves.}
    \label{fig:surface_param}
\end{figure}

NURBS are a widely used mathematical tool for representing curves and surfaces in computer-aided design (CAD) and computer graphics, see \figref{fig:surface_param}. At the core of NURBS are B-splines, which are piecewise polynomial curves constructed as linear combinations of B-spline basis functions. These basis functions are defined over a knot vector, denoted as:
\begin{equation}
    \mathbf{U}=\{u_0,\dots,u_{n+p}\}.
\end{equation}
Here, $p$ represents the degree of the B-spline, and $n$ is the number of control points. The knot vector contains non-decreasing real numbers that partition the parametric space of the curve. The B-spline basis functions $N_{i,p}(u)$ are defined recursively. Starting with piecewise constant functions for $p=0$:
\begin{equation}
N_{i,0}(u)=
    \begin{cases}
      1 & \text{if $u_{i}\leq u<u_{i+1}$}\\
      0 & \text{otherwise}
    \end{cases}.
\end{equation}
For higher degrees $p = 1 , 2 , 3 , \dots$, the basis functions are computed using the following recursive formula:
\begin{equation}
N_{i,p}(u)=\frac{u-u_i}{u_{i+p}-u_i}N_{i,p-1}(u)+\frac{u_{i+p+1}-u}{u_{i+p+1}-u_{i+1}}N_{i+1,p-1}(u).
\end{equation}
These basis functions are used to construct B-spline curves in combination with the control points. B-spline surfaces can be generated by extending the concept of B-spline curves into two dimensions using tensor products. Given two knot vectors $\mathbf{U}=\{u_0,\dots,u_{n+p}\}$ and $\mathbf{V}=\{v_0,\dots,v_{m+q}\}$, along with an $n \times m $ grid of control points $\mathbf{P}_{i,j}$, the B-spline surface is defined as:
\begin{equation}
\mathbf{S}(u,v)=\sum_{i=0}^{n}\sum_{j=0}^{m}N_{i,p}(u)N_{j,q}(v)\mathbf{P}_{i,j}
\end{equation}
where $N_{i,p}(u)$ and $N_{j,q}(v)$ are one-dimensional basis functions corresponding to the respective knot vectors. NURBS extend B-splines by introducing weights for each control point. Each control point $\mathbf{P}_{i,j}$ is augmented with a weight $w_{i,j}$. The surface is then defined as: 
\begin{equation}
\mathbf{S}(u,v)=\frac{\sum\limits_{i=0}^{n}\sum\limits_{j=0}^{m}N_{i,p}(u)N_{j,q}(v)w_{i,j}\mathbf{P}_{i,j}}{\sum\limits_{i=0}^{n}\sum\limits_{j=0}^{m}N_{i,p}(u)N_{j,q}(v)w_{i,j}} \ \ \ \ \ 0\leq u,v\leq1.
\end{equation}
The rational basis functions provide additional flexibility in shaping curves and surfaces. This allows NURBS to accurately represent both standard geometric shapes (such as circles and ellipses) and complex free-form surfaces.

\subsection{Finding initial NURBS surface using particle swarm optimization (PSO)}
\label{subsec:pso}

\begin{figure}[t!]
    \centering
    \includegraphics[trim=8cm 0cm 8cm 0cm, clip, width=0.75\linewidth]{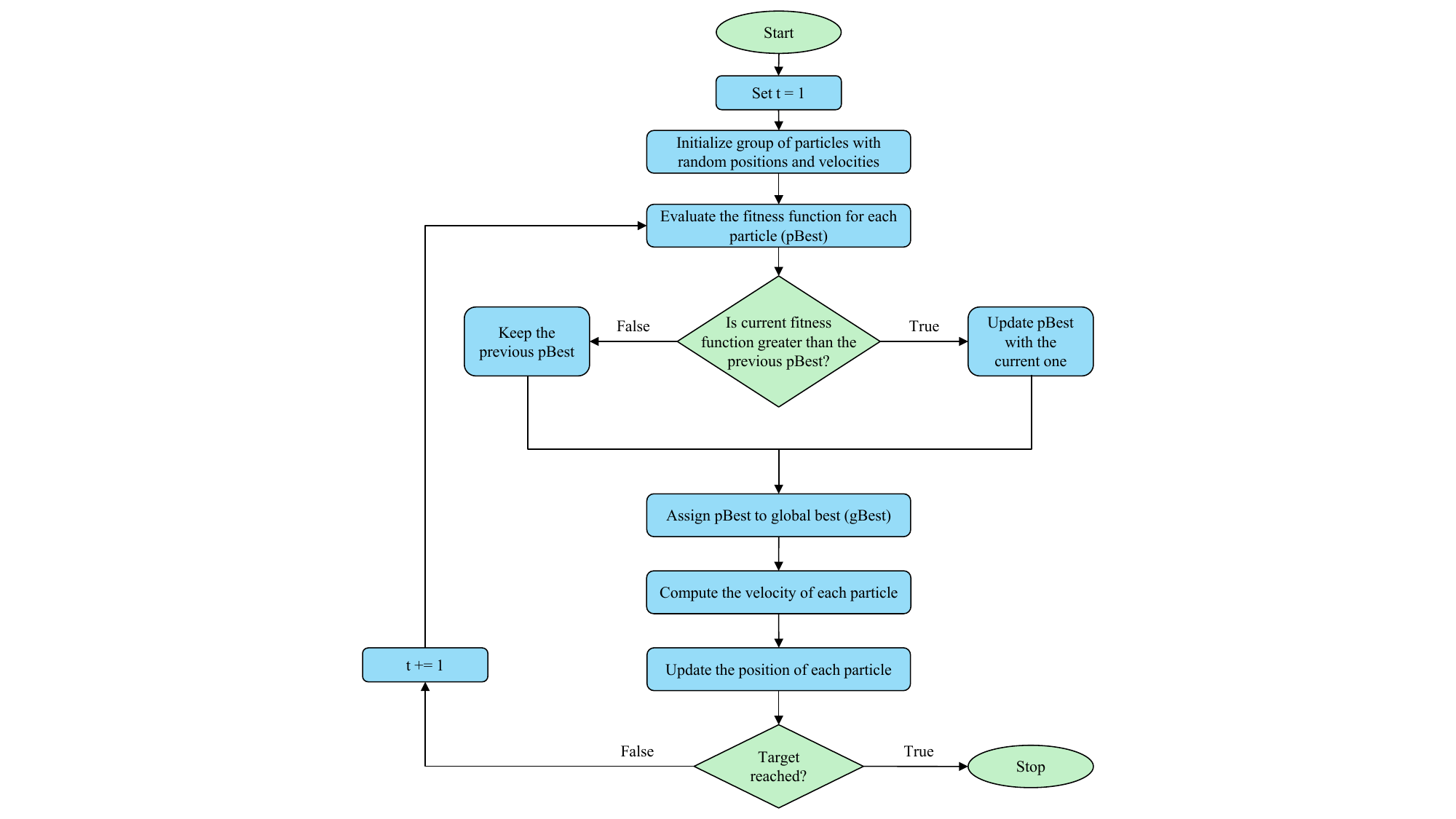}
    \caption{Flowchart illustrating the PSO algorithm used for initializing the NURBS surface. Key steps include initializing particle positions and velocities, evaluating fitness, updating global and personal best positions, and iteratively refining the swarm until convergence is achieved.} 
    \label{fig:pso_algorithm}
\end{figure}

The first step of our optimization process is PSO, which is used to determine the initial control points for the NURBS surface that represents each leaf. PSO is a population-based stochastic optimization technique inspired by the social behavior of bird flocking or fish schooling. As illustrated in \figref{fig:pso_algorithm}, PSO works by iteratively improving candidate solutions with regard to a given measure of quality, which is the fitness function. This process involves particles, each representing potential solutions, that explore the parameter space collectively to converge on an optimal configuration.

At each iteration \( t \), the position of each particle \( m_i(t) \) is updated using its velocity \( v_i(t) \), as defined by:
\begin{equation}
m_i(t + 1) = m_i(t) + v_i(t + 1)
\end{equation}
where the computed velocity at \( t + 1 \) is used to update the position at the current timestep \( t \). What is meant by the velocity update is:
\begin{equation}
v_{ij}(t + 1) = w * v_{ij}(t) + c_1 r_{1j}(t) [pbest_{ij}(t) - m_{ij}(t)] + c_2 r_{2j}(t) [gbest_j(t) - m_{ij}(t)]
\end{equation}
Here:

- \( w \) is the inertia weight, which controls the influence of the swarm's movement,

- \( c_1 \) and \( c_2 \) present cognitive and social parameters, respectively,

- \( r_1 \) and \( r_2 \) are random numbers between 0 and 1, ensuring stochastic behavior,

- \( pbest_{ij} \) is the personal best position of the particle \( i \),

- \( gbest_j \) is the global best position found by any particle in the swarm.\\
These parameters are used to control particle movements. It may move towards its personal best or follow the global best solution found by the swarm. Whether the swarm will behave in an exploratory or exploitative manner depends on the inertia weight \( w \). Exploration is encouraged with a greater inertia weight, whereas exploitation is encouraged by a smaller one~\cite{pyswarmsJOSS2018}. 
In our implementation, we used the parameter values shown in \tabref{tab:PSOhyperparameters} for the PSO algorithm. The inertia weight \( w \) follows an exponential decay strategy, while the cognitive (\( c_1 \)) and social (\( c_2 \)) components vary linearly over iterations to balance exploration and exploitation.

\begin{table}[t!]
\centering
\caption{Hyperparameter used in the PSO algorithm.}
\label{tab:PSOhyperparameters}
\setlength{\extrarowheight}{5pt}
    \begin{tabularx}{\textwidth}{|c|c|X|X|} 
    \hline
    \textbf{Parameter} & \textbf{Value} & \textbf{Hyperparameter Strategy} & \textbf{Description} \\ \hline
    $c_1$ & 2.5 & Linear variation & Cognitive component; controls the particle's tendency to move towards its own best position. \\ \hline
    $c_2$ & 0.5 & Linear variation & Social component; controls the particle's tendency to move towards the global best position. \\ \hline
    $w$ & 0.9 & Exponential decay & Inertia weight; controls the influence of the previous velocity on the current velocity. \\ \hline
    \end{tabularx}
\end{table}

\subsubsection*{Design Variables: Leaf Control Points}\label{subsec:pso-initialization}

In our model, PSO optimizes 32 parameters that define each leaf's NURBS surface control points. Aspect ratio normalization is used in each leaf point cloud data to ensure that our model can handle leaves of any size. The \(x\), \(y\), and \(z\) coordinates are specifically normalized by dividing them by the greatest \(x\) value in each leaf. This method maintains the aspect ratio and allows our model to function consistently over a range of leaf sizes by scaling the data proportionately to the \(x\) dimension. We also obtain a size-invariant representation by completing this normalization, which enables the PSO to optimize the NURBS surface regardless of the leaf's absolute dimensions. We rescale the optimal NURBS surface to its original dimensions following the PSO process, which restores the leaf's actual size and creates a precise geometric representation in relation to the point cloud data.

The parameters are organized into a \(3 \times 6\) matrix, forming 18 control points that define the NURBS surface grid in the u and v direction based on specific constraints to create a NURBS surface similar to the shape of the maize leaf as shown in \figref{fig:surface_param}. Each control point is represented in three-dimensional space ( \(x\), \(y\), and \(z\)), yielding 54 coordinate values. Additionally, the bounds of the parameters are carefully selected to enable efficient exploration and convergence. The parameters are structured as follows:
\begin{itemize}
    \item \textbf{\(x\)-coordinates:} Six \(x\)-coordinates are generated and shared across all three rows to ensure a consistent length for the maize leaf. These parameters are constrained within bounds from 0.2 to 1.2, where the lower bound of 0.2 helps avoid the generation of points that are too close together during initialization. To maintain an ordered sequence and prevent folding in the NURBS surface, we apply a cumulative sum (\texttt{cumsum}) operation on the six \(x\)-coordinates. Afterward, we normalize these values by dividing them by the mean of the normal distribution, which scales them appropriately within the target bounds and preserves the desired aspect ratio.
    
    \item \textbf{\(y\)-coordinates:} The middle row is defined by six \(y\)-values, and the first and third rows are produced by modifying these values using six matchings \(\Delta y\) values. An example of \(\Delta y\) is shown in \figref{fig:surface_param_b} for \(v_2\) which is \(\Delta y_2\). This results in a uniform profile for the width of the leaf. Each Y-value is limited to the range of -0.1 to 0.1. To ensure that the base (where the leaf attaches to the stalk) and the tip of the leaf remain narrow, we set boundaries of 0.0 to 0.1 for the \(\Delta y\) values in the first and last columns. This reduced width at the ends reflects the natural tapering of the leaf. We use a broader range of 0.05 to 0.2 for \(\Delta y\) in the middle columns, allowing for a wider midsection that better represents the leaf's natural shape.
    
    \item \textbf{\(z\)-coordinates:} The leaf's natural curvature is accurately modeled by 14 \(z\)-values, which capture the vertical variation along its length and width. To ensure smooth transitions at the leaf's base, the initial control points in each row are set to the same \(z\)-values. Similarly, to create a consistent alignment of the endpoint that mimics the natural tapering of the leaf, the last control points in each row are also set to the same values in the \(z\) direction. The bounds for these values are between 0.0 and 1.0.
\end{itemize}

Ultimately, the final control points matrix can be represented as a function of the following 32 variables:

$$\begin{bmatrix} P_{0,0} & P_{0,1} & P_{0,2} & P_{0,3} & P_{0,4} & P_{0,5} \\ P_{1,0} & P_{1,1} & P_{1,2} & P_{1,3} & P_{1,4} & P_{1,5} \\ P_{2,0} & P_{2,1} & P_{2,2} & P_{2,3} & P_{2,4} & P_{2,5} \\ \end{bmatrix} = $$

\setlength{\arraycolsep}{2pt}
$$\begin{bmatrix} x_0, y_0 + \Delta y_0, z_0 & x_1, y_1 + \Delta y_1, z_{10} & x_2, y_2 + \Delta y_2, z_{11} & x_3, y_3 + \Delta y_3, z_{12} & x_4, y_4 + \Delta y_4, z_{13} & x_5, y_5 + \Delta y_5, z_5 \\ x_0, y_0, z_0 & x_1, y_1, z_6 & x_2, y_2, z_7 & x_3, y_3, z_8 & x_4, y_4, z_9 & x_5, y_5, z_5 \\ x_0, y_0 - \Delta y_0, z_0 & x_1, y_1 - \Delta y_1, z_1 & x_2, y_2 - \Delta y_2, z_2 & x_3, y_3 - \Delta y_3, z_3 & x_4, y_4 - \Delta y_4, z_4 & x_5, y_5 - \Delta y_5, z_5 \\ \end{bmatrix}$$

In our approach, we utilize PSO in two distinct steps:

\begin{enumerate}
    \item \textbf{Phyllotaxy Extraction:} 
    During the optimization process, aligning the point cloud data and the reconstructed surface of each leaf is essential to accurately compare distances between them. To achieve this alignment, we first extract the rotation angle (phyllotaxy) of each leaf, which helps approximate the leaf's natural orientation. Additionally, we align the centroid of the surface mesh with the centroid of the point cloud data, ensuring consistent spatial positioning.

    To obtain this rotation angle, we perform a single iteration of PSO. In this step, PSO is initialized with a group of 300 random particles, each representing a candidate solution in a 33-dimensional space (parameters). From these parameters, 32 of them are used to define the \(3 \times 6\) control point coordinates for the NURBS surface, as previously explained, and one extra parameter is dedicated to extracting the rotation angle. This setup provides an optimal initial alignment for the NURBS surface relative to the point cloud data.

    \item \textbf{Initial Surface Fitting:} 
    Once the optimal rotation angle is determined, we proceed to fit an initial NURBS surface to the point cloud data data using PSO. This second step uses the previously determined rotation angle to ensure proper alignment of the generated NURBS surface with the input point cloud data, allowing for precise comparison. We set the number of iterations to 50 based on experimental observations. Fewer iterations do not produce an adequately accurate initial NURBS surface for further alignment, while more than 50 iterations show diminishing returns in minimizing the distance, as the cost function tends to flatten after 50 iterations. Therefore, 50 iterations provide an effective balance, ensuring an accurate initial surface fit without unnecessary computational overhead. In this step, we use only 32 parameters, as the rotation angle is now fixed. By the end of this process, the initial NURBS surface reasonably approximates the input point cloud data.

\end{enumerate}

To solve our 32 (or 33) dimensional problem in this study, we use the PSO method with a population size of 300 particles. Recent studies on population sizing in high-dimensional PSO applications are consistent with this choice. \citet{guo2023} noted that larger swarm sizes are often beneficial for complex, high-dimensional problems, as they allow for better exploration of the search space. Our selection of 300 particles falls within the range suggested by \citet{Yao2024}, who recommended population sizes between 200 and 500 for problems with more than 30 dimensions. This aligns with earlier work by \citet{Sengupta2018}, who suggested similar population sizes for high-dimensional problems. In addition, our selection closely resembles the heuristic put forward by \citet{Xu2019}, which suggests that for high-dimensional optimization challenges, the population size should be roughly ten times the issue dimensionality. This heuristic is consistent with the earlier recommendation by \citet{Clerc2010}, who proposed a similar sizing strategy. For our 32 (or 33) dimensional problem, this heuristic would indicate a swarm size of 320-330 particles, validating our selection of 300. This larger swarm size allows for better exploration of the high-dimensional search space, which is crucial for avoiding premature convergence to local optima. \citet{Zheng2023} observed that while increasing the swarm size beyond a certain point may offer diminishing returns, it can still improve solution quality for complex problems. This observation is in line with the findings of \citet{Bonyadi2017}, who highlighted the benefits of larger swarms in complex, high-dimensional spaces.

The fitness function for our PSO implementation combines two distance metrics to balance between overall fit and avoiding large discrepancies, which is defined as:
\begin{equation}
    \mathcal{L}_{PSO} = d_{CD}(X, Y) + \lambda_{HD} d_{HD}(X, Y)
\end{equation}
where:
\begin{itemize}
    \item \( d_{CD}(X, Y) \): Chamfer Distance;
    \item \( d_{HD}(X, Y) \): Hausdorff Distance;
    \item \( \lambda_{HD} \): Weight for Hausdorff Distance (e.g., 0.1).
\end{itemize}

\noindent\textbf{Chamfer Distance:} Chamfer distance measures the similarity between two point sets by summing the squared distances between each point in one set and its nearest point in the other set. It computes this in both directions: from points in set $X$ to their nearest points in set $Y$ and from points in set $Y$ to their nearest points in set $X$. This makes it a useful metric for comparing shapes or surfaces without requiring them to have identical points.
\begin{equation}
    d_{CD}(X, Y) = \sum_{x \in X} \min_{y \in Y} \| x - y \|_2^2 + \sum_{y \in Y} \min_{x \in X} \| x - y \|_2^2
\end{equation}

\noindent\textbf{Hausdorff Distance:} Hausdorff distance measures the maximum distance between two point sets. It identifies the point in one set that is farthest from any point in the other set and takes that distance as the Hausdorff distance. This metric is useful for capturing the largest deviation between two shapes or surfaces, making it sensitive to outliers. The Hausdorff distance between two sets \( X \) and \( Y \) is defined as:
\begin{equation}
d_H(X, Y) = \max(H(X, Y), H(Y, X));
\end{equation}
where the directed Hausdorff distance \( H(X, Y) \) is:
\begin{equation}
H(X, Y) = \max_{x \in X} \min_{y \in Y} \|x - y\|
\end{equation}
The two-sided Hausdorff distance is:
\begin{equation}
d_H(X, Y) = \max \left( \max_{x \in X} \min_{y \in Y} \|x - y\|, \max_{y \in Y} \min_{x \in X} \|y - x\| \right)
\end{equation}

This combination of Chamfer and Hausdorff distances allows for a balanced evaluation of the similarity between the point clouds. Chamfer distance provides an average measure of how close the point clouds are, while the Hausdorff distance captures the largest discrepancy. By weighting the Hausdorff distance, in our case, multiplying by 0.1, we can ensure that large outliers do not dominate the optimization, allowing for a robust fit while still accounting for extreme deviations.

\subsection{Differentiable NURBS-based optimization}
\label{subsec:nurbs-diff}

We use NURBS-Diff, a differentiable spline module designed to integrate NURBS CAD models with deep learning methods~\citep{prasad2022nurbs}, to perform the gradient-based fitting of the final NURBS surface starting from the NURBS control points predicted by PSO. NURBS-Diff module mathematically defines the derivatives of NURBS curves and surfaces with respect to input parameters (i.e., control points, weights, and knot vectors). These derivatives are then utilized to construct an approximate Jacobian, which enables the backward evaluation necessary for training deep learning models. Here, to streamline the procedure, the weights and knot vectors are maintained as constants, focusing solely on optimizing the positions of the control points. The derivative of the NURBS surface with respect to the control points is provided in \eqnref{eq:S,P_ij}:
\begin{equation}
\label{eq:S,P_ij}
\mathbf{S}_{,\mathbf{P}_{i,j}}(u,v) = \frac{N_{i,p}(u)N_{j,q}(v)w_{i,j}}{\sum\limits_{k=0}^{n}\sum\limits_{l=0}^{m}N_{k,p}(u)N_{l,q}(v)w_{k,l}}.
\end{equation}

To fit the original NURBS surface from PSO to the unstructured LIDAR data, we use the loss function:
\begin{equation}
\label{eq:loss_NURBS-Diff}
    \mathcal{L}_{\text{NURBS-Diff}} = d_{CD}^{\text{one-sided}}(X, Y) + \lambda_{\text{curv}_{\text{11}}} \mathcal{L}_{\text{curv}_{\text{11}}} + \lambda_{\text{curv}_{\text{12}}} \mathcal{L}_{\text{curv}_{\text{12}}} + \lambda_{\text{proximity}} \mathcal{L}_{\text{proximity}}
\end{equation}
where $\mathcal{L}_{\text{curv}_{\text{11}}}$ and $\mathcal{L}_{\text{curv}_{\text{12}}}$ are incorporated terms that minimize the curvature of the surface, thereby preventing sharp twists and contributing to a smoother surface representation and $\mathcal{L}_{\text{proximity}}$ is added to ensure that the specified control points stay closely grouped, keeping the first control points and last control points in the v-direction close to one another. These terms are mathematically expressed as follows:
\begin{gather}
    \mathcal{L}_{\text{curv}_{\text{11}}}=\sum_{\mathbf{S_i}\in \mathbf{S}}\frac{\partial^2\mathbf{S_i}}{\partial u^2} \\
    \mathcal{L}_{\text{curv}_{\text{12}}}=\sum_{\mathbf{S_i}\in \mathbf{S}}\frac{\partial^2\mathbf{S_i}}{\partial u\partial v} \\
    \mathcal{L}_{\text{proximity}}=\sum_{j=1,v}(\| \mathbf{P}_{1,j} - \mathbf{P}_{2,j} \|_2^2+\| \mathbf{P}_{1,j} - \mathbf{P}_{3,j} \|_2^2+\| \mathbf{P}_{2,j} - \mathbf{P}_{3,j} \|_2^2).
\end{gather}
Weighting factors, denoted as $\lambda_{\text{curv}_{\text{11}}}={10}^{-2},\ \lambda_{\text{curv}_{\text{12}}}={10}^{-4},\ \text{and}\ \lambda_{\text{proximity}}=8\times{10}^{-4}$, serve to control the relative importance of each corresponding component in the overall loss function. As described in \secref{subsec:pso}, the input NURBS surface is defined by a bi-cubic $3\times6$ grid of control points. The process employs an $8\times32$ grid of evaluation points and utilizes the Adam optimizer with an initial learning rate of $10^{-2}$ which decays to half the previous learning rate every $50$ iterations over $500$ total iterations.

\section{Results and Discussion}
\label{sec:results}

\begin{figure}[t!]
    \centering
    \begin{subfigure}[b]{0.21\linewidth}
        \centering
        \includegraphics[trim=1cm 7cm 1cm 2.2cm, clip, width=\linewidth]{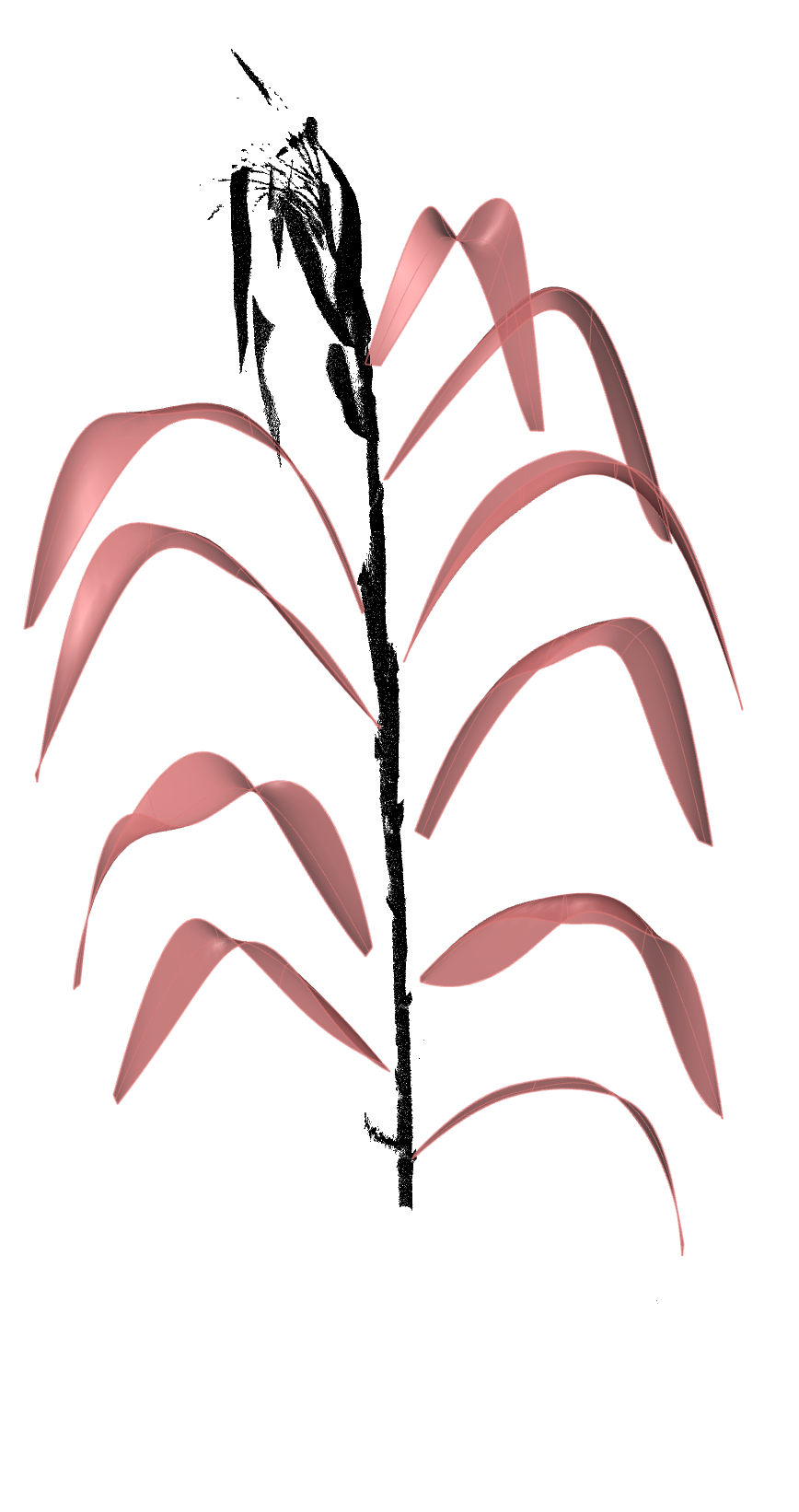}
        \caption{CML238}
    \end{subfigure}
    \begin{subfigure}[b]{0.20\linewidth}
        \centering
        \includegraphics[trim=4cm 9cm 3cm 2.0cm, clip, width=\linewidth]{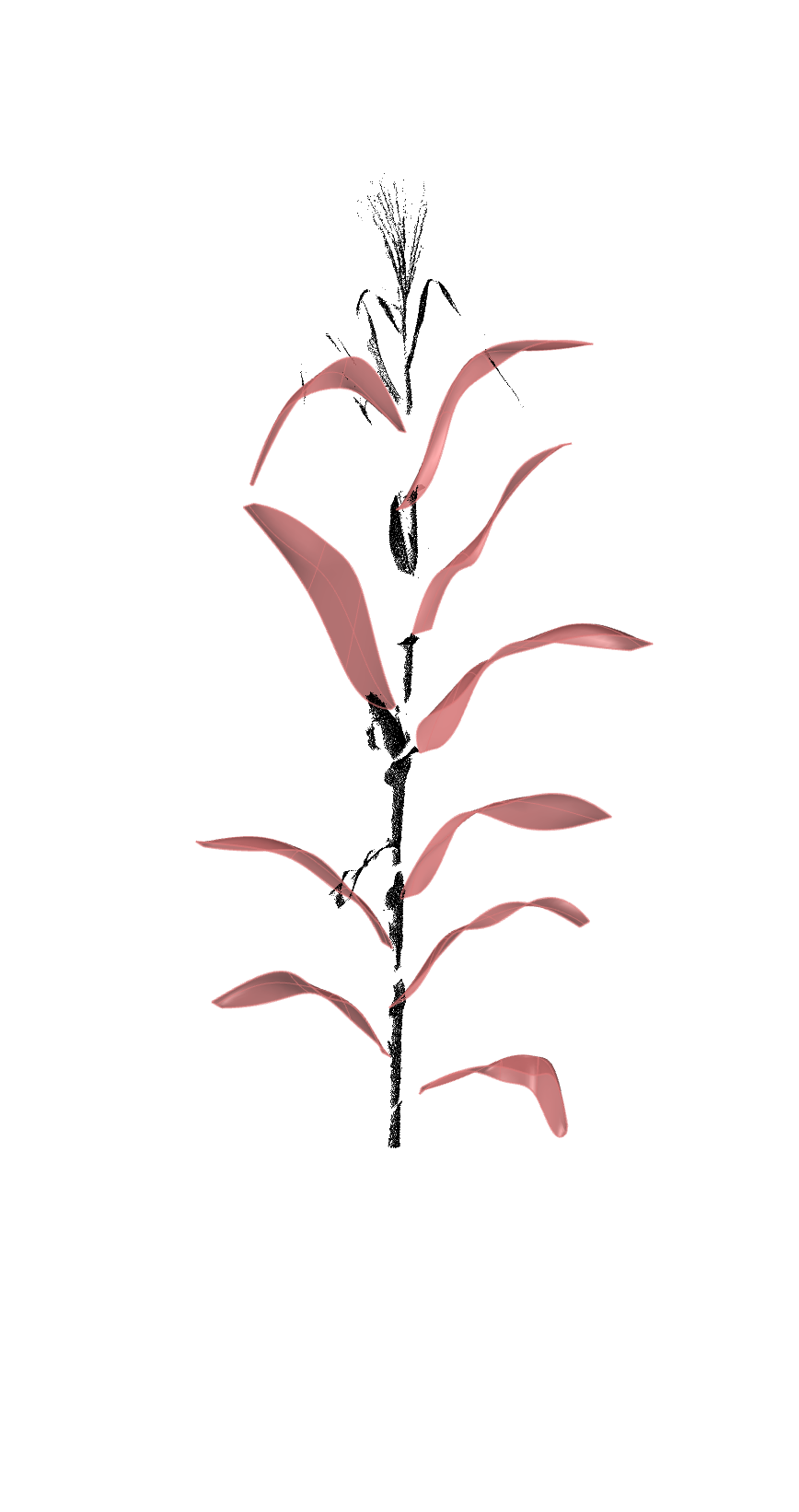}
        \caption{T8}
    \end{subfigure}
    \begin{subfigure}[b]{0.21\linewidth}
        \centering
        \includegraphics[trim=2cm 6cm 2cm 6cm, clip, width=\linewidth]{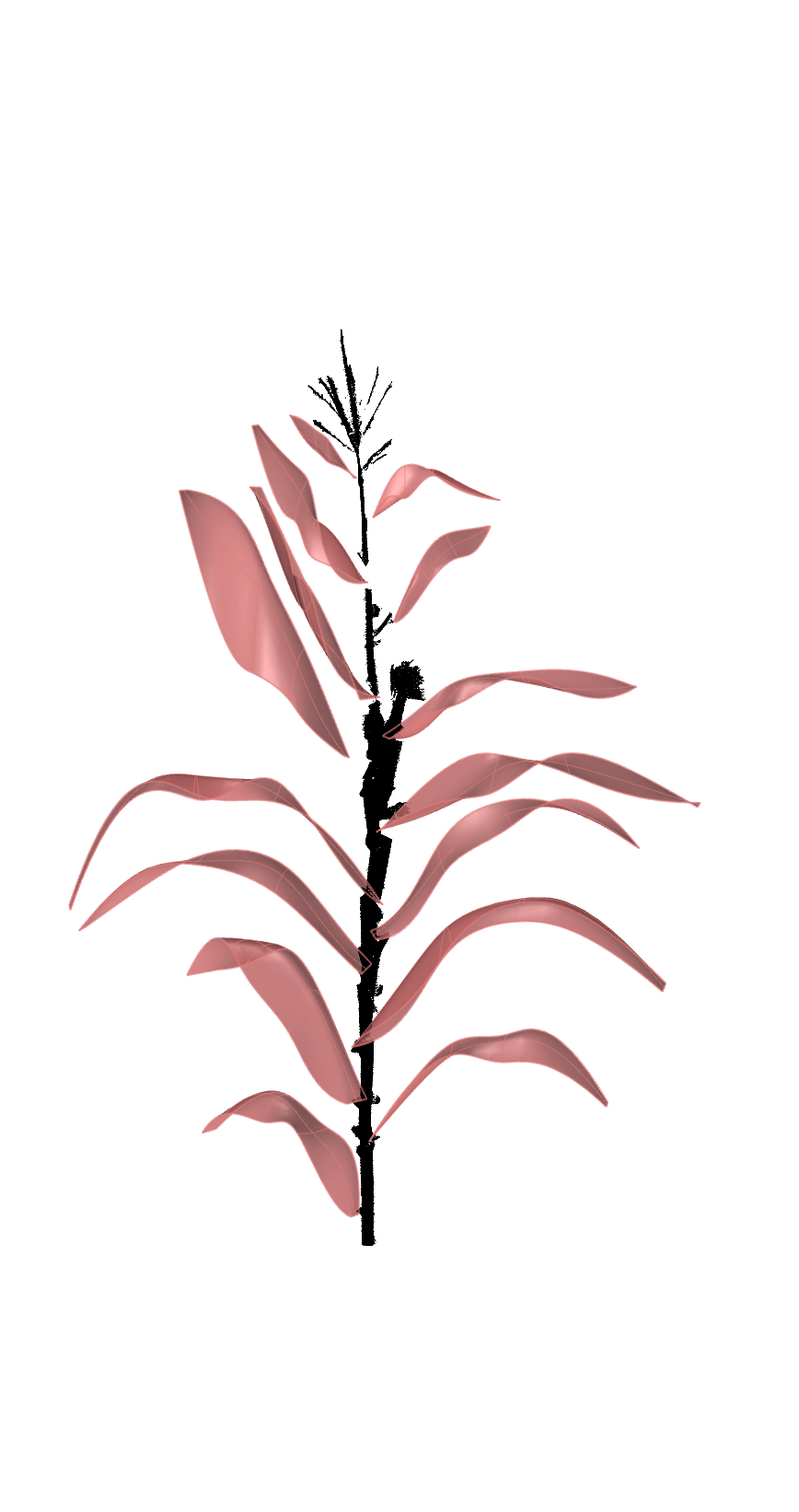}
        \caption{M162W}
    \end{subfigure}
    \begin{subfigure}[b]{0.21\linewidth}
        \centering
        \includegraphics[trim=2cm 2cm -2cm 6cm, clip, width=\linewidth]{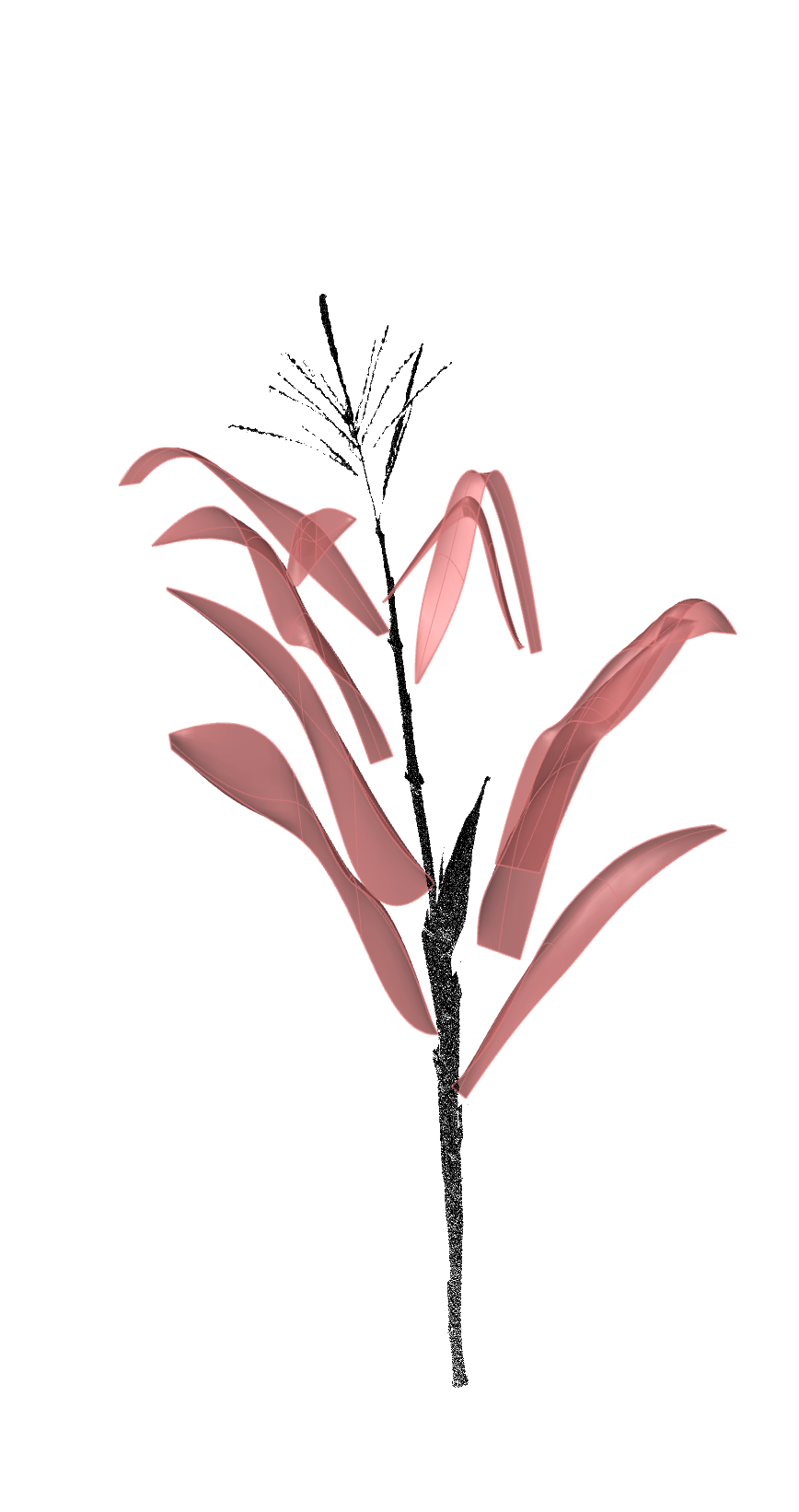}
        \caption{CI90C}
    \end{subfigure}
    \caption{Output of the PSO optimization step for four genotypes. The PSO algorithm provides an initial approximation of the NURBS surface, which captures the overall leaf structure but requires further refinement to resolve fine details, especially along edges and tips.}
    \label{fig:pso_output}
\end{figure}

This section presents the optimization results obtained using PSO and NURBS-Diff methodologies for leaf surface reconstruction. The segmented LIDAR point cloud data for four different genotypes are shown in \figref{fig:segmented_GT}. We begin by examining the performance of PSO in approximating individual leaf surfaces, followed by a detailed analysis of the NURBS-Diff approach and its comparative advantages. Our preliminary approach approximates the surface of each leaf using PSO. \figref{fig:pso_output} shows how well this approach can approximate the segmented point cloud data.

\begin{figure}[t!]
    \centering
    \begin{subfigure}[b]{0.21\linewidth}
        \centering
        \includegraphics[trim=1cm 8cm 1cm 2.2cm, clip, width=\linewidth]{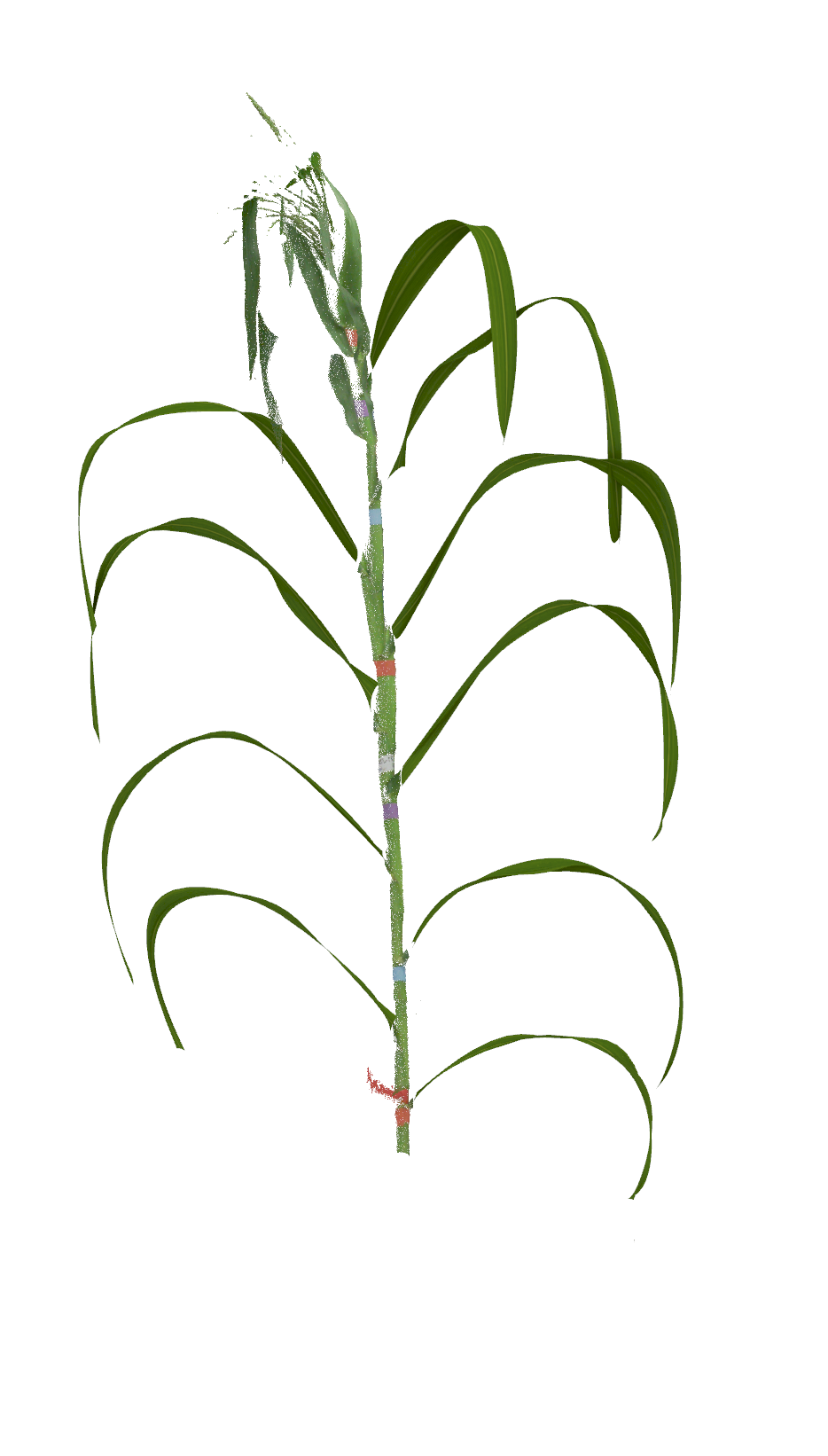}
         \caption{CML238}
    \end{subfigure}
    \begin{subfigure}[b]{0.21\linewidth}
        \centering
        \includegraphics[trim=2cm 3.5cm 1cm 2.0cm, clip, width=\linewidth]{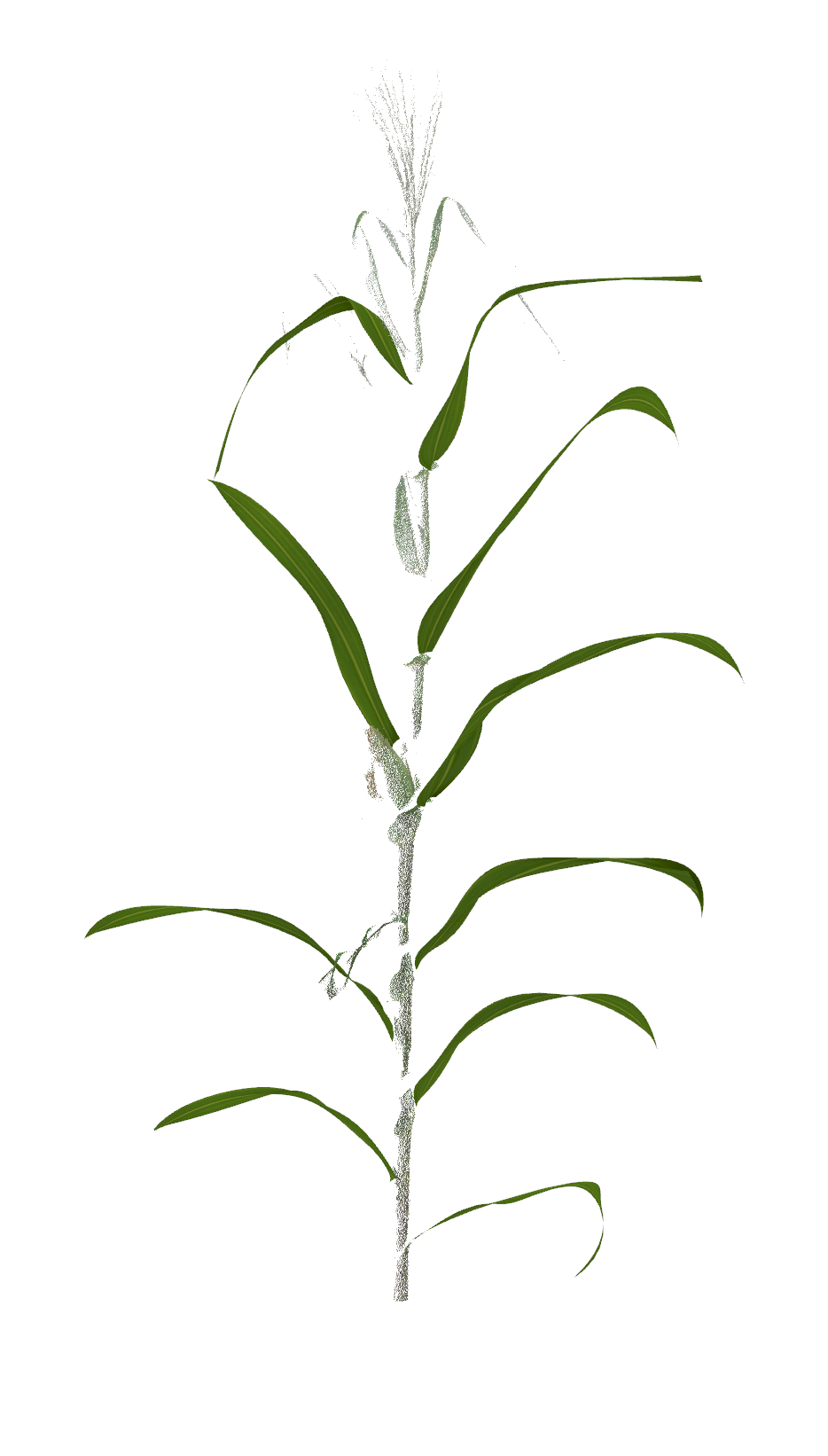}
        \caption{T8}
    \end{subfigure}
    \begin{subfigure}[b]{0.21\linewidth}
        \centering
        \includegraphics[trim=1cm 4cm 1cm 6.1cm, clip, width=\linewidth]{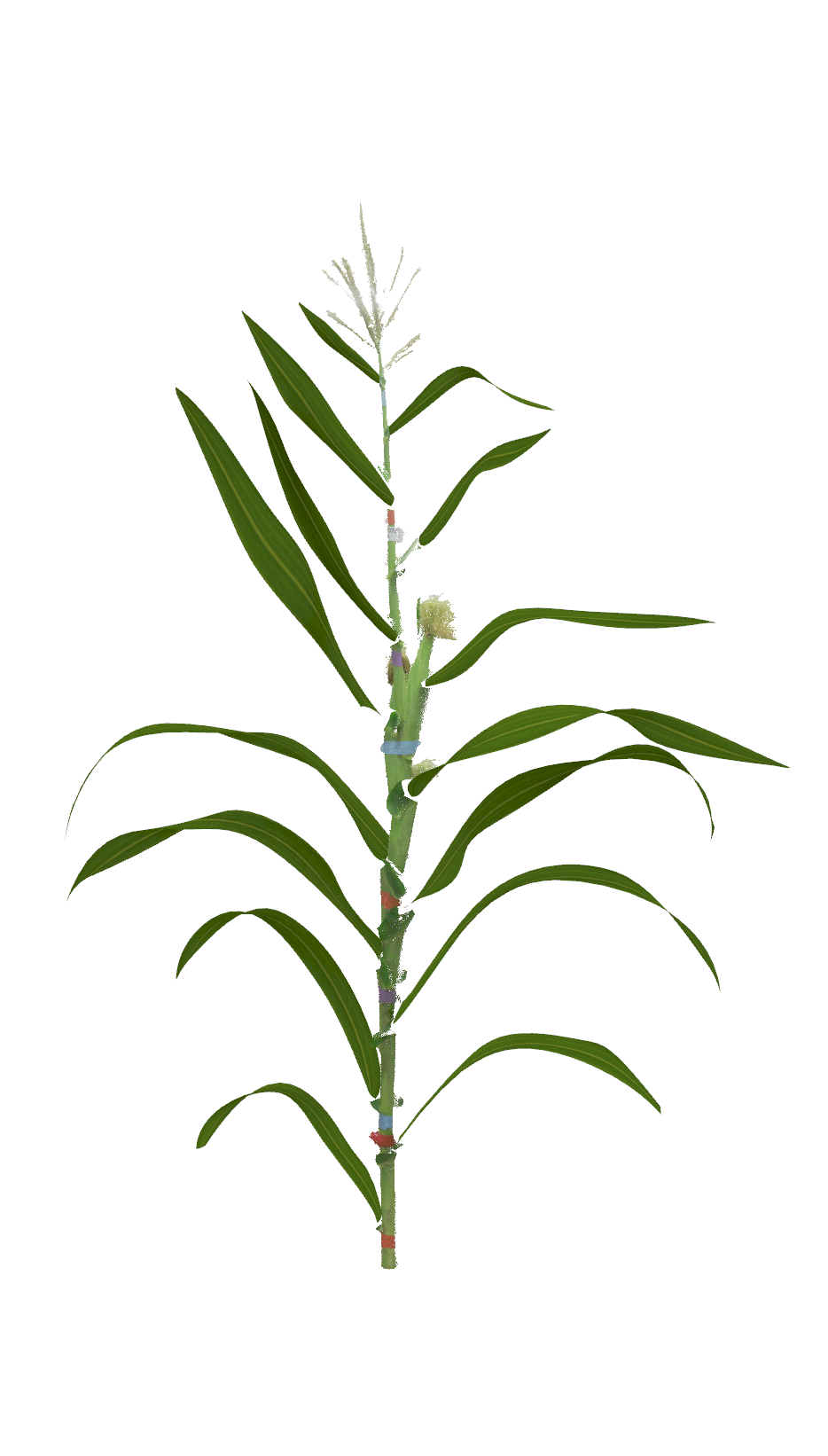}
        \caption{M162W}
    \end{subfigure}
    \begin{subfigure}[b]{0.21\linewidth}
        \centering
        \includegraphics[trim=1cm 2cm -2cm 6.1cm, clip, width=\linewidth]{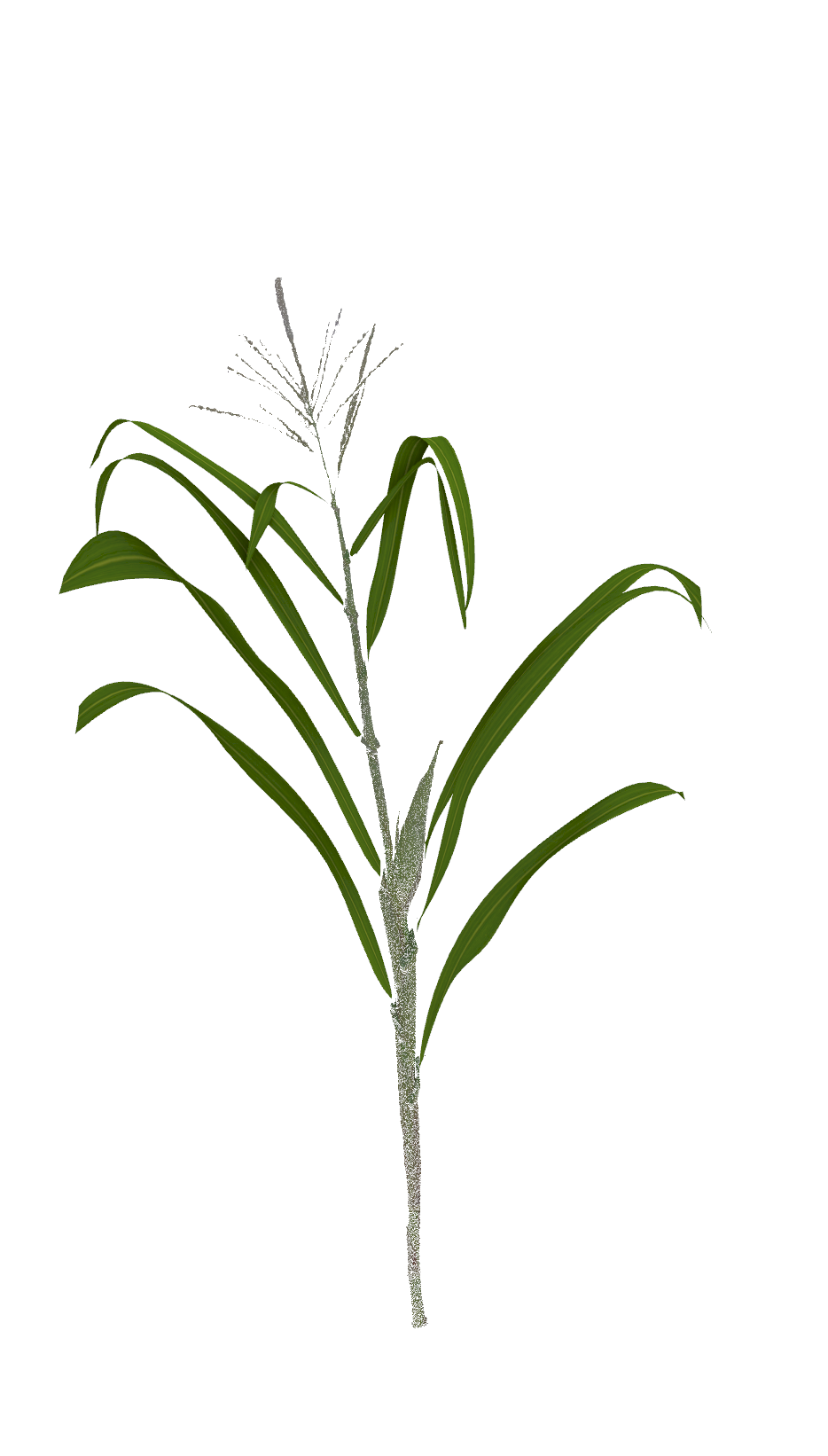}
        \caption{CI90C}
    \end{subfigure}
    \caption{Final NURBS surface reconstruction using the NURBS-Diff optimization method for four genotypes. This step enhances the initial PSO results by achieving high-fidelity alignment with the point cloud data, accurately capturing leaf edges, tips, and curvatures.}
    \label{fig:nurbs_output}
\end{figure}

Even though the PSO surface and the input point clouds have a similar overall form and structure, there are differences at granular scales, especially along the edges and tips of the leaves, which need further refinement. We implement this step to quickly obtain an initial NURBS surface to feed the NURBS-Diff. The leaf surface reconstruction accuracy is improved by applying the NURBS-Diff optimization after the PSO. As shown in \figref{fig:nurbs_output}, the NURBS-Diff output closely matches the segmented point clouds for CML238, T8, M162W, and CI90C genotypes, and more detailed information about the leaf structure is captured by the improved surfaces, especially near the edges and tips, where the PSO results showed differences.

\begin{figure}[t!]
    \centering
    \includegraphics[width=0.48\textwidth]{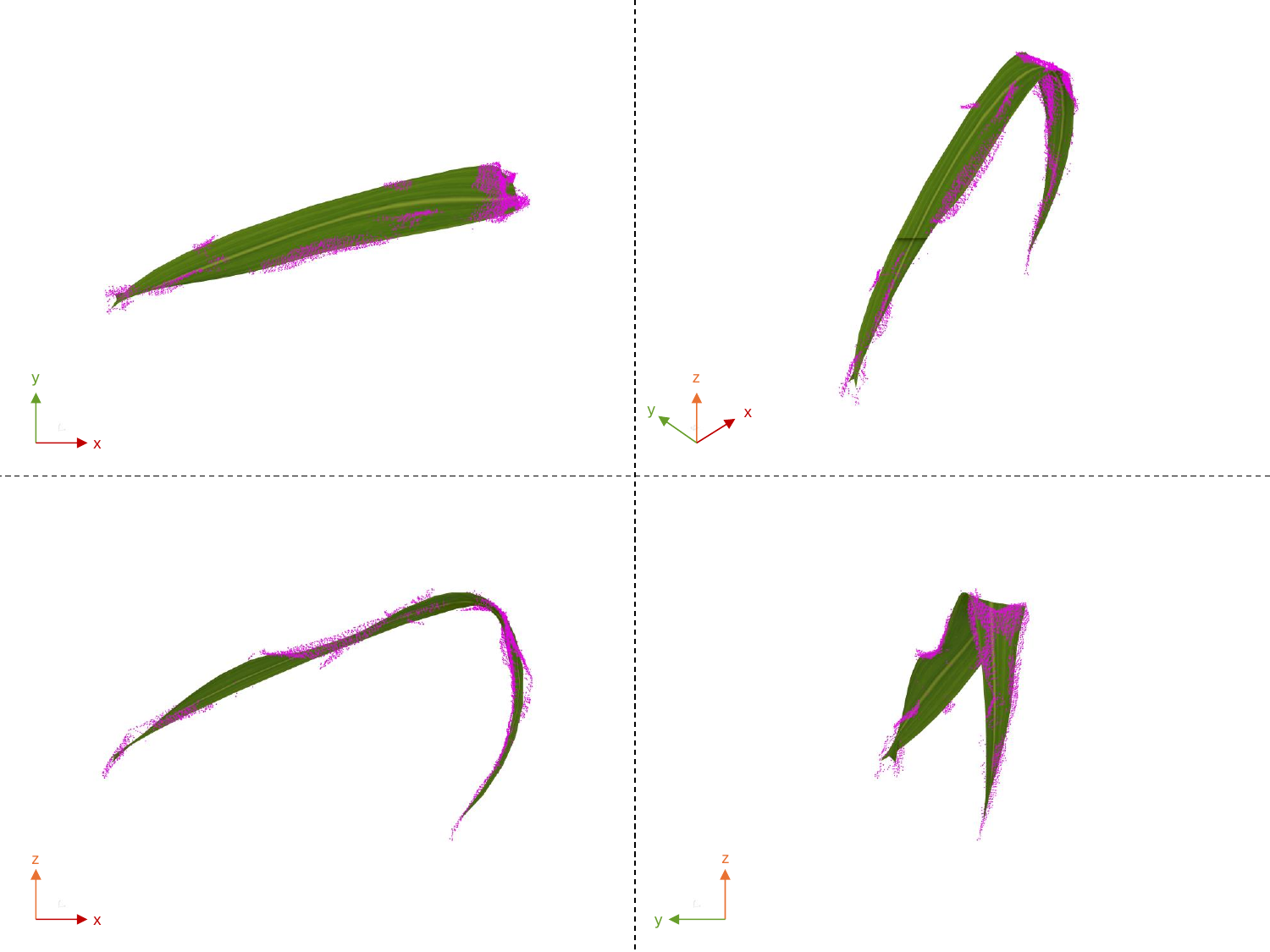}
    \includegraphics[width=0.48\textwidth]{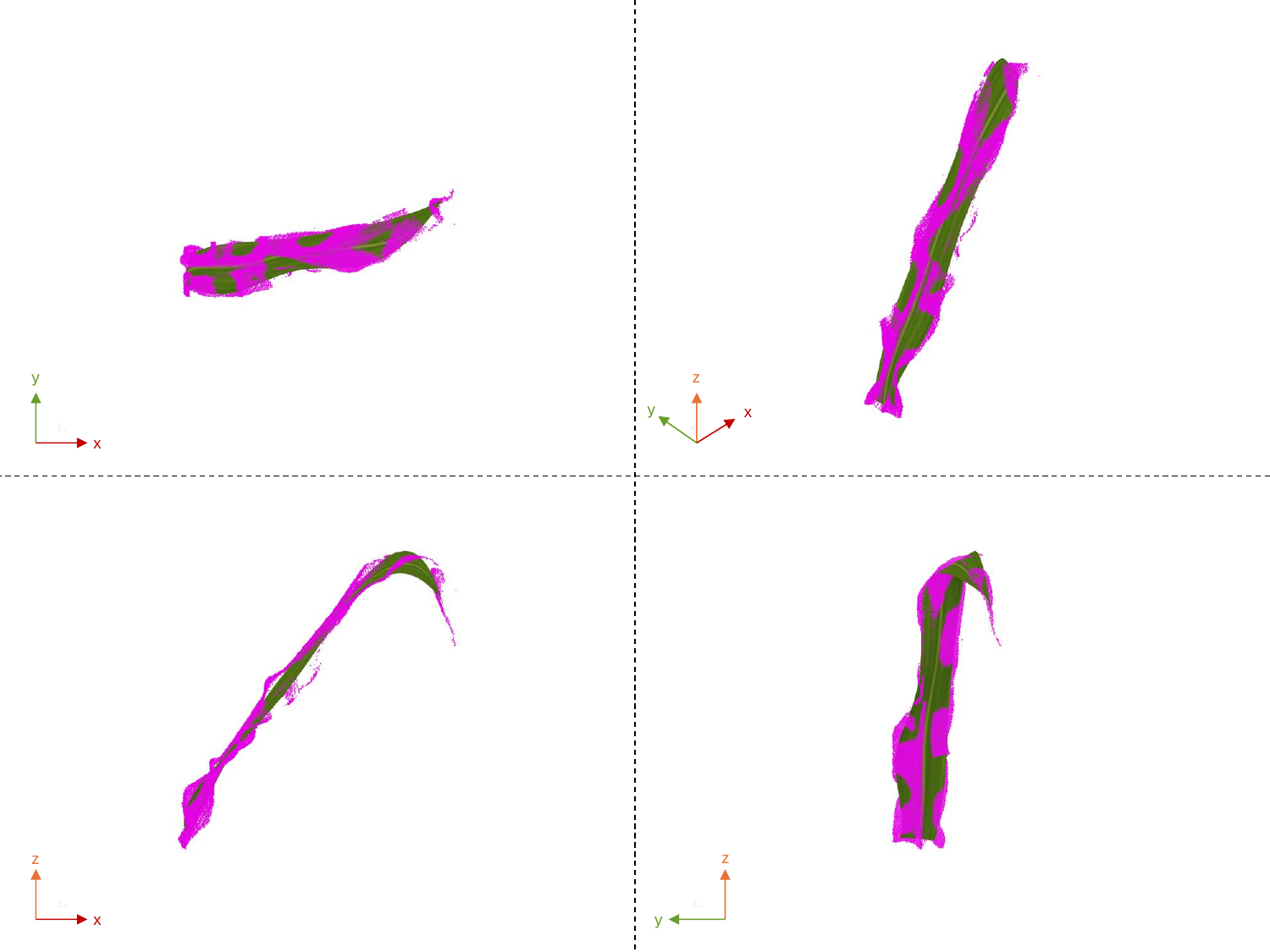}
    \caption{Comparison of input LiDAR point clouds (magenta color) and final NURBS-Diff surfaces (green color) for two leaves from the T8 genotype: Leaf 1 (left) and Leaf 8 (right). These examples illustrate the method's effectiveness in handling variations in data quality, capturing complex curvatures, and reconstructing realistic leaf geometries across different views.}
    \label{fig:T8_Leaves}
\end{figure}

\begin{figure}[t!]
    \centering
    \setlength{\unitlength}{1cm} 
    \begin{picture}(0,0)
        \put(0.9,0){\footnotesize (a) M162W} 
        \put(4.0,0){\footnotesize (b) CML238} 
        \put(7.1,0){\footnotesize( c) T8} 
        \put(10.0,0){\footnotesize (d) CML69} 
        \put(12.9,0){\footnotesize (e) CI90C} 
    \end{picture}
    \includegraphics[trim=9cm 18cm 9cm 24cm, clip, width=0.9\linewidth]{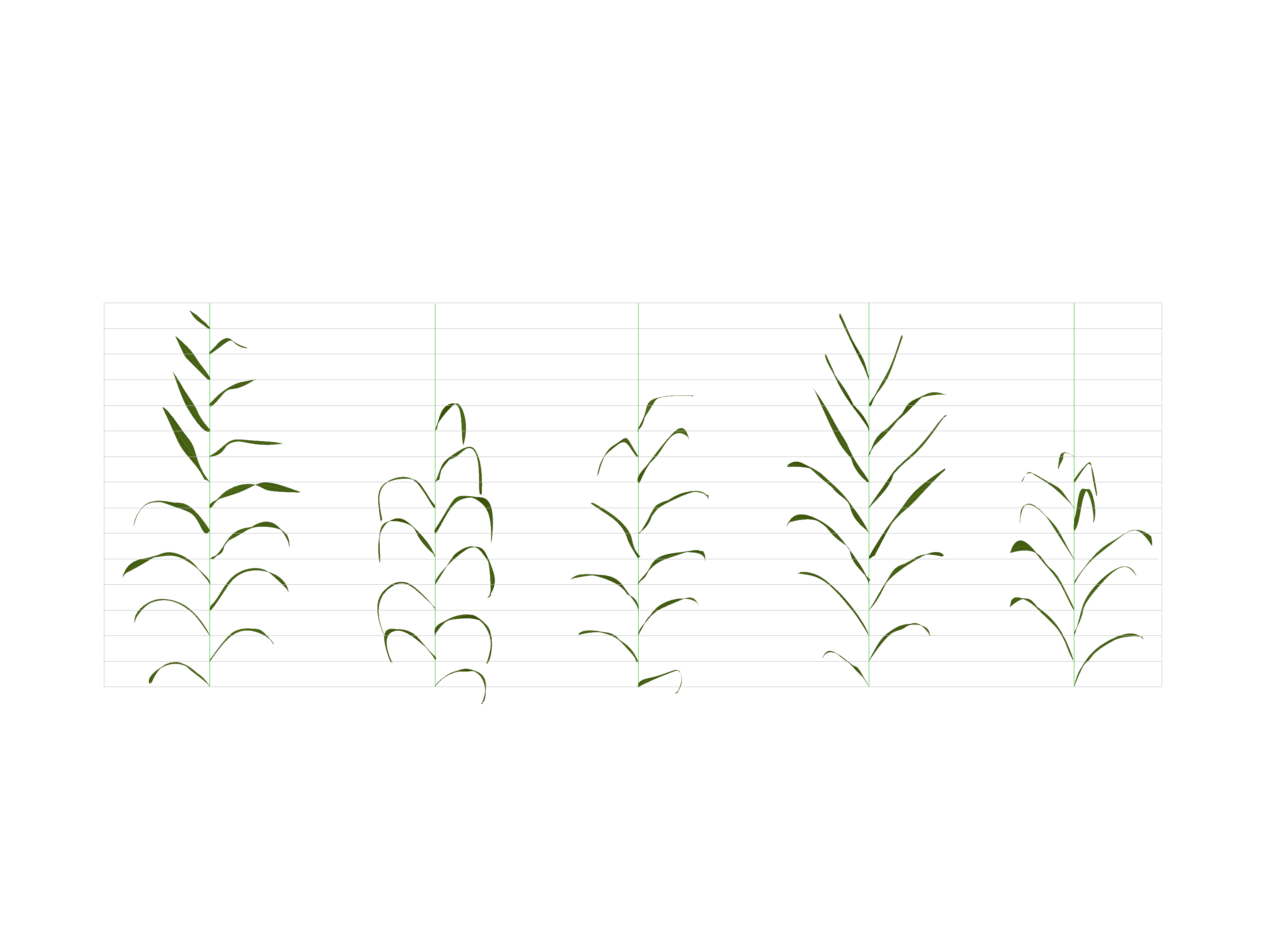}
    \caption{Final reconstructed 3D leaf surfaces for five maize genotypes. The leaves are aligned in a standardized format by adjusting the z-coordinates of them to visually separate them and positioning all leaves at 0 and 180 degrees. This adjustment highlights the variability in leaf size, shape, and curvature across genotypes, providing a clear visualization of structural differences. The figure demonstrates the method's ability to handle diverse leaf morphologies, maintaining robustness and adaptability for precise reconstruction across genetic populations.}
    \label{fig:Diverse_genotype}
\end{figure}

The combination of PSO for initialization and NURBS-Diff for fine-tuning ensures a balance between global smoothness and local detail, enabling accurate reconstructions across diverse leaf structures. This adaptability is crucial for applications in high-throughput phenotyping, where datasets may vary widely in quality and completeness. The results in \figref{fig:T8_Leaves} emphasize the versatility of the proposed method in reconstructing maize leaves with varying geometric complexities and data quality. Leaf 1 presents a unique challenge due to significant gaps and missing points in the point-cloud data. Despite these irregularities, the NURBS-Diff approach effectively bridges the gaps, producing a smooth and continuous surface that aligns well with the available data. This outcome demonstrates the robustness of the curvature and proximity penalties in the loss function, which guide the surface generation process even when faced with incomplete or sparse point cloud data. The method also accurately captures the distinct shapes of leaves, including pronounced curvature and narrow tip, for example, the sharp curvature in leaf eight. Furthermore, the method's performance across leaves with both sparse and dense point clouds suggests its potential scalability to other plant species and broader phenotypic studies, making it a reliable tool for generating realistic virtual plant models (also see \tabref{tab:AllGenotypesTable} that shows the procedural model side by side with the 2D photos of these plants illustrating the applicability of the framework to fit diverse maize genotypes.).

Our method shows strong performance across phenotypically diverse maize genotypes, despite differences in leaf size, shape, and curvature, demonstrating its adaptability in precisely reconstructing 3D leaf surfaces. The final reconstructed surfaces for the several different genotypes are shown in \figref{fig:Diverse_genotype}. Although each genotype has distinct leaf traits, our approach can handle these differences successfully. This robustness demonstrates how our method can be adapted to diverse genotypes without requiring special modifications to leaf size, shape, or curvature. Reconstructing these genotypes shows how our approach can be modified to different morphological characteristics, making it a valuable tool for studies on plant phenotype.  

\pagebreak

The effectiveness of using a two-step optimization approach is evident in the Chamfer distance reduction for both CML238 and T8 genotypes. As illustrated in \tabref{tab:chamfer_distance_reduction}, there is a significant decrease in Chamfer distance across all leaves after applying NURBS-Diff. For CML238, initial distances after PSO are as high as 1.02 mm (Leaf4), which are reduced to 0.06 mm post-optimization. Similarly, for T8, distances decrease from 0.36 mm (Leaf5) to 0.04 mm. This significant reduction highlights how well the NURBS-Diff optimization can finetune leaf surfaces to get an improved fit to the input point cloud data.

The average runtime for both PSO and NURBS-Diff optimization processes for all plants is shown in \tabref{tab:runtime}. The PSO stage, which provides the initial approximation of the leaf surfaces, takes the majority of the time. The subsequent NURBS-Diff optimization, which refines these initial approximations, requires a significantly smaller computation time. A maize plant with ten leaves takes~$\approx$ 1 hour to generate the complete procedural model using our framework. We note that this can be easily reduced by reconstructing each leaf in parallel with multiple processors.

\begin{table}[t!]
    \centering
    \caption{Chamfer distance reduction for the CML238 and the T8 genotype. The table compares distances for ten leaves before and after NURBS-Diff optimization.}
    \label{tab:chamfer_distance_reduction}
    \setlength{\extrarowheight}{3pt}
    \footnotesize
    \begin{tabular}{|c|r|r|r|r|}
        \hline
        & \multicolumn{2}{c|}{\textbf{CML238}} & \multicolumn{2}{c|}{\textbf{T8}} \\ \hline
        \textbf{leaf No.} & \textbf{After PSO (mm)} & \textbf{After NURBS-Diff (mm)} & \textbf{After PSO (mm)} & \textbf{After NURBS-Diff (mm)} \\
        \hline
        1  & 0.13 & 0.03 & 0.24 & 0.03 \\
        2  & 0.31 & 0.04 & 0.12 & 0.03 \\
        3  & 0.28 & 0.05 & 0.20 & 0.03 \\
        4  & 1.02 & 0.06 & 0.29 & 0.03 \\
        5  & 0.44 & 0.06 & 0.36 & 0.04 \\
        6  & 0.36 & 0.05 & 0.16 & 0.06 \\
        7  & 0.43 & 0.07 & 0.12 & 0.05 \\
        8  & 0.31 & 0.06 & 0.34 & 0.06 \\
        9  & 0.32 & 0.06 & 0.09 & 0.03 \\
        10 & 0.26 & 0.05 & 0.14 & 0.02 \\
        \hline
    \end{tabular}
\end{table}

\begin{table}[t!]
    \centering
    \footnotesize
    \caption{Average computational time for reconstructing a maize plant using the combined PSO and NURBS-Diff approach.}
    \label{tab:runtime}
    \setlength{\extrarowheight}{3pt}
    \begin{tabular}{| l | r |}
        \hline
        \textbf{Method} & \textbf{Runtime (Mean ± Standard Deviation) (s)} \\ \hline
        PSO & 2740 ± 180 \\
        NURBS-Diff & 690 ± 310 \\ \hline
        Total & 3430 ± 490 \\ \hline
    \end{tabular}
\end{table}

\begin{figure}[t!]
    \centering
    \includegraphics[width=0.38\linewidth]{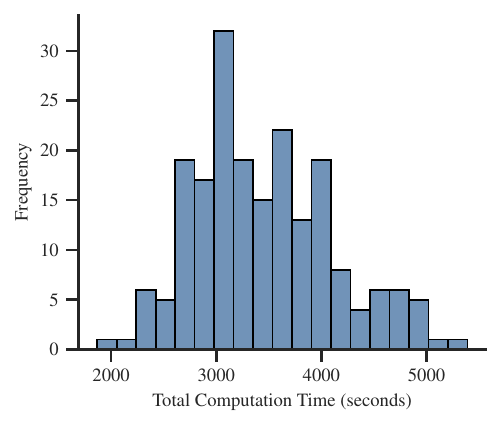}
    \caption{Histogram of the total computation time (in seconds) for generating the 3D maize plant procedural models for 200 plants representing various genotypes from the SAM diversity panel.}
    \label{fig:total_time}
\end{figure}

\pagebreak

The histogram in \figref{fig:total_time} highlights the variation of the total computational time required for creating the procedural models for 200 diverse maize plants. The computations were performed on a high-performance 96-core AMD CPU, which significantly speeds up the optimization process. The variability in computation time across plants arises primarily due to the differing number of leaves and varying numbers of points in each point cloud. Plants with more leaves or denser point clouds naturally require more processing time for both the PSO and NURBS-Diff phases. The histogram illustrates that the procedural model for the vast majority of plant was  computed within the range of 2500–4000 seconds, with some outliers extending up to approximately 5000 seconds. These outliers correspond to plants with either high leaf counts or particularly dense point clouds.

\begin{figure}[t!]
    \centering
    \begin{subfigure}[b]{0.21\linewidth}
        \centering
        \includegraphics[trim=1cm 2cm 1cm 2cm, clip, width=\linewidth]{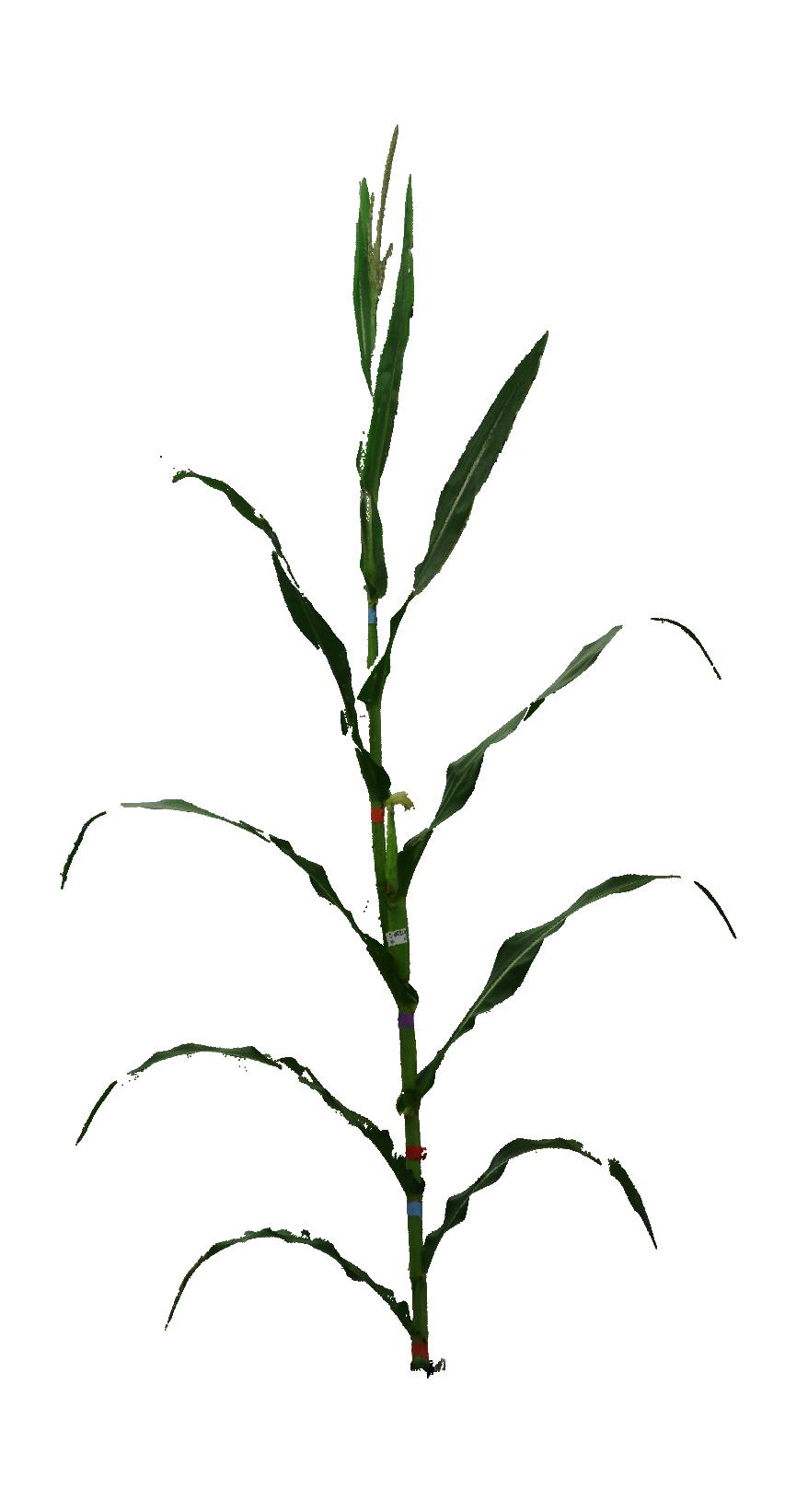}
        \caption{}
    \end{subfigure}
    \begin{subfigure}[b]{0.21\linewidth}
        \centering
        \includegraphics[trim=1cm 2cm 1cm 2cm, clip, width=\linewidth]{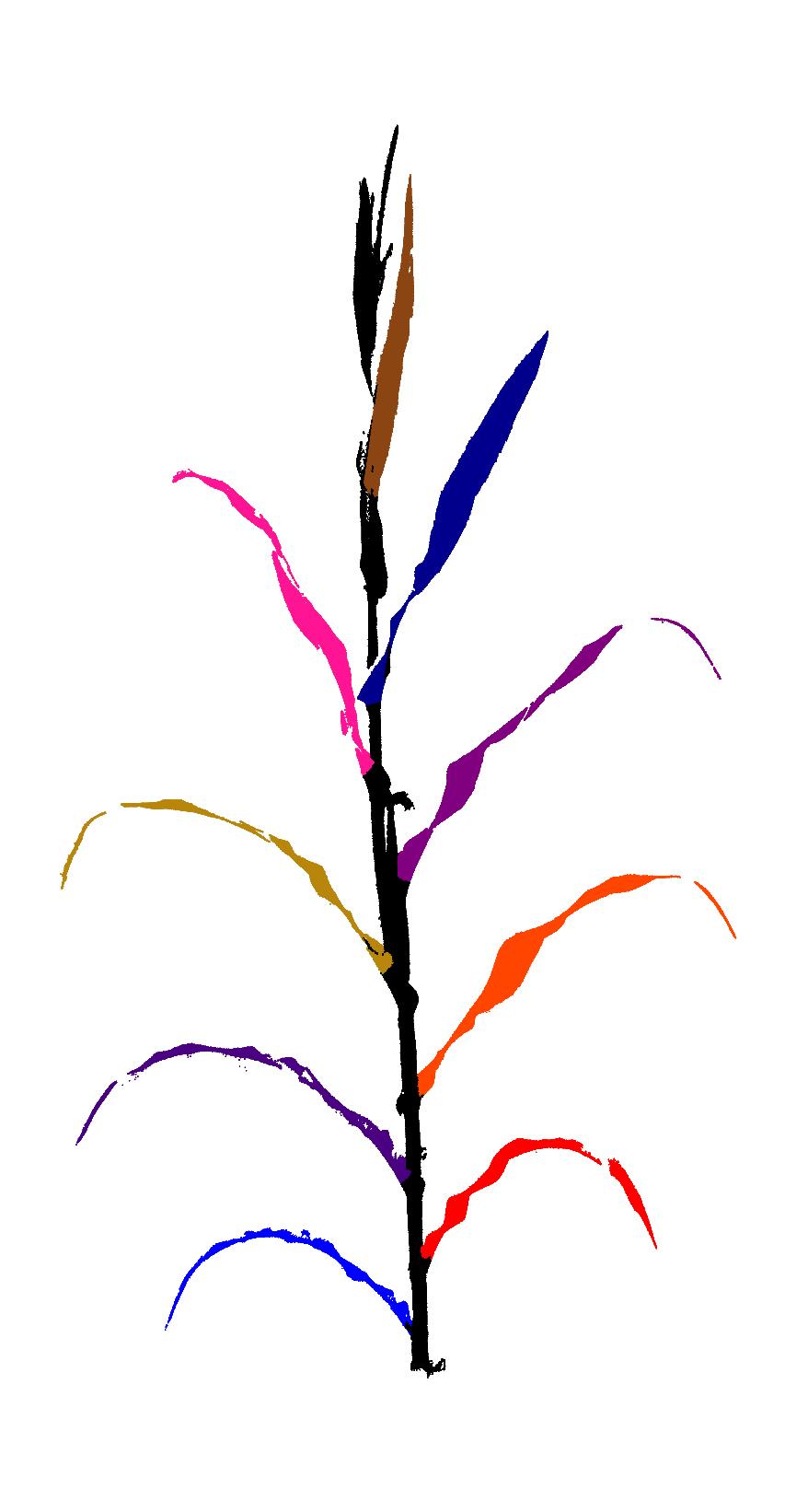}
        \caption{}
    \end{subfigure}
    \begin{subfigure}[b]{0.21\linewidth}
        \centering
        \includegraphics[trim=1cm 2cm 1cm 2cm, clip, width=\linewidth]{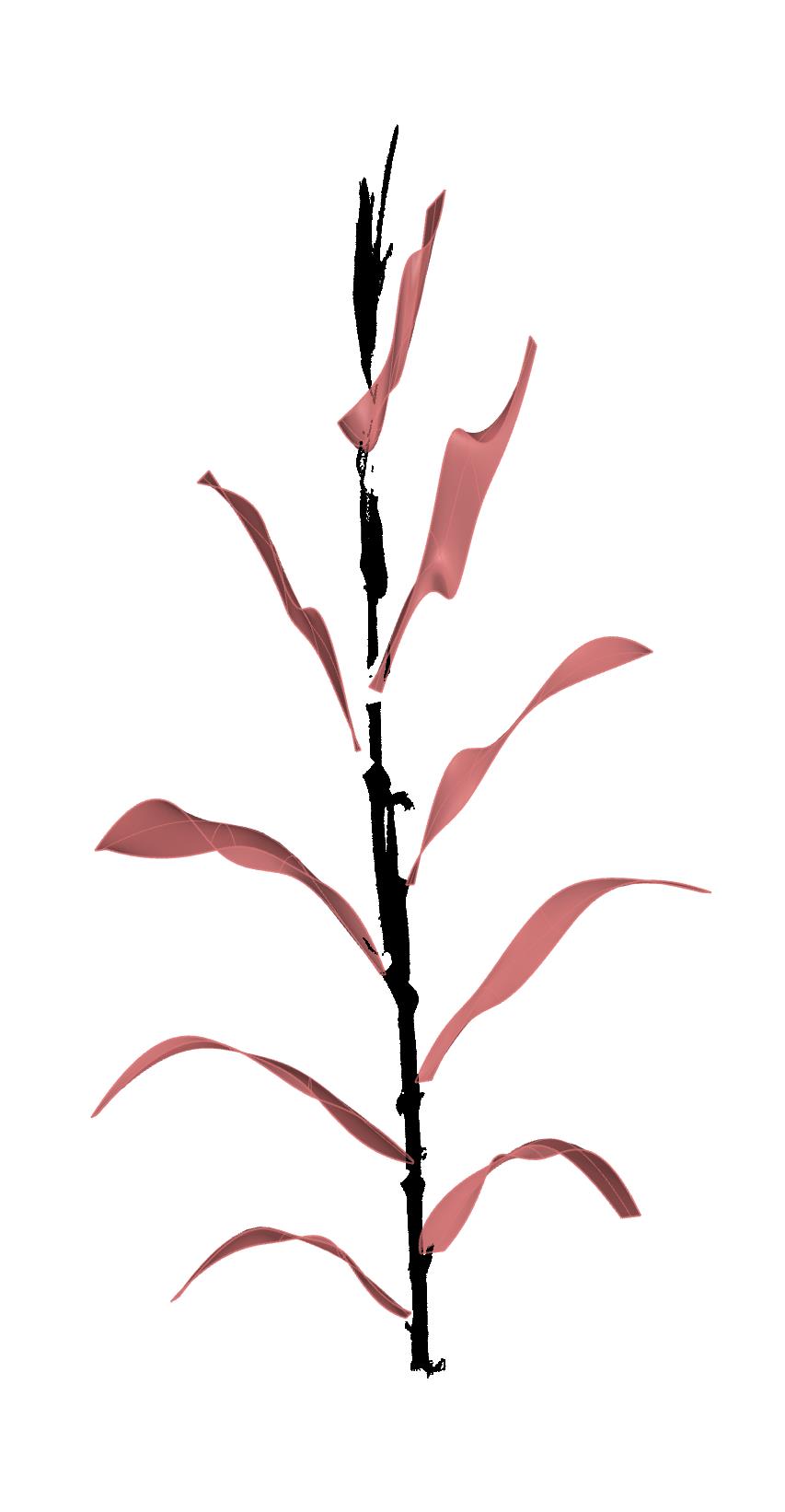}
        \caption{}
    \end{subfigure}
    \begin{subfigure}[b]{0.21\linewidth}
        \centering
        \includegraphics[trim=1cm 2cm 1cm 2cm, clip, width=\linewidth]{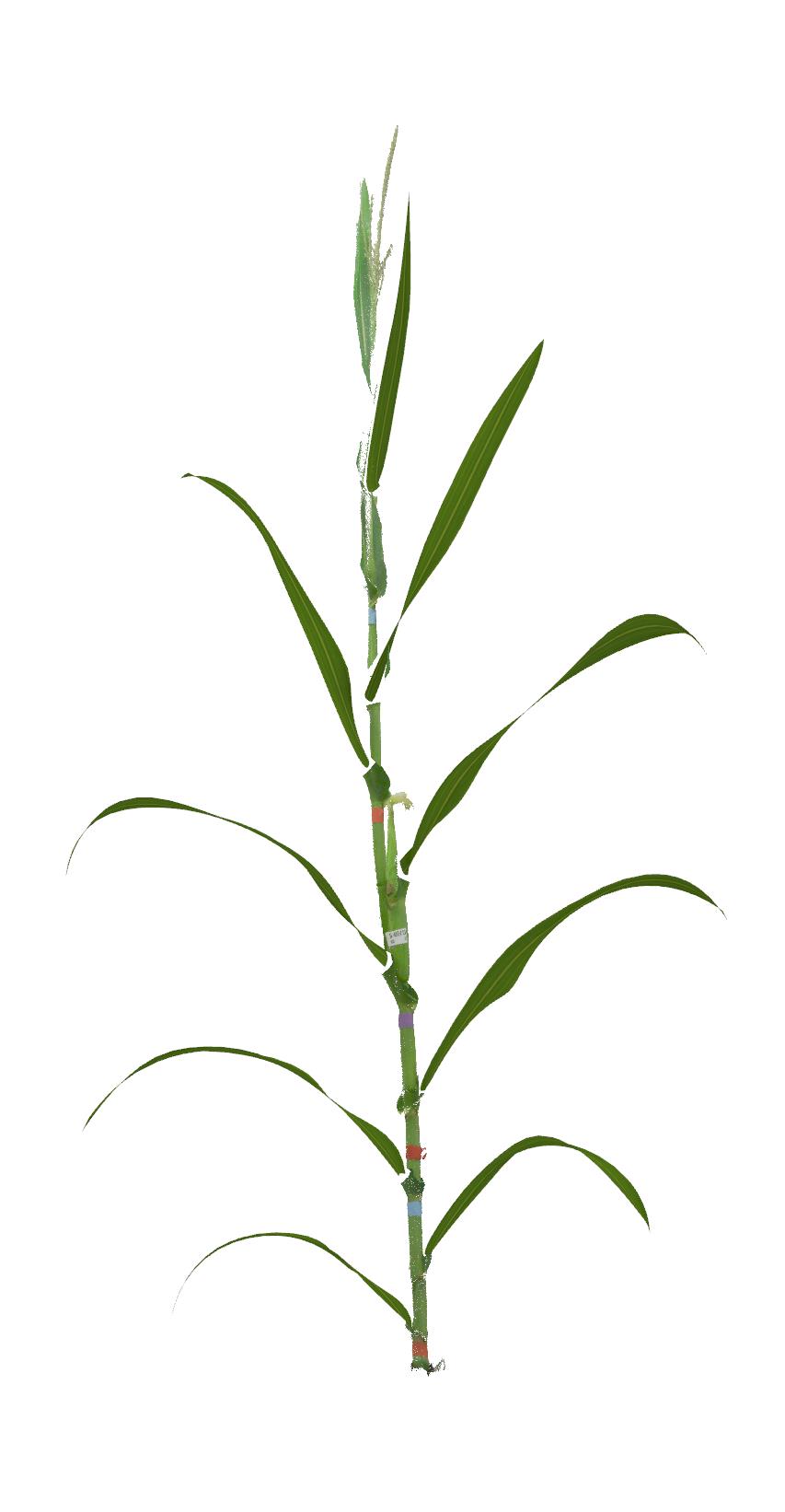}
        \caption{}
    \end{subfigure}
    \caption{Step-by-step processing of B73 maize plant. (a) Raw data, (b) segmented data, (c) PSO output, and (d) final NURBS surface.}
    \label{fig:B73}
\end{figure}

\begin{figure}[t!]
    \centering
    \begin{subfigure}[b]{0.21\linewidth}
        \centering
        \includegraphics[trim=1cm 4cm 1cm 2cm, clip, width=\linewidth]{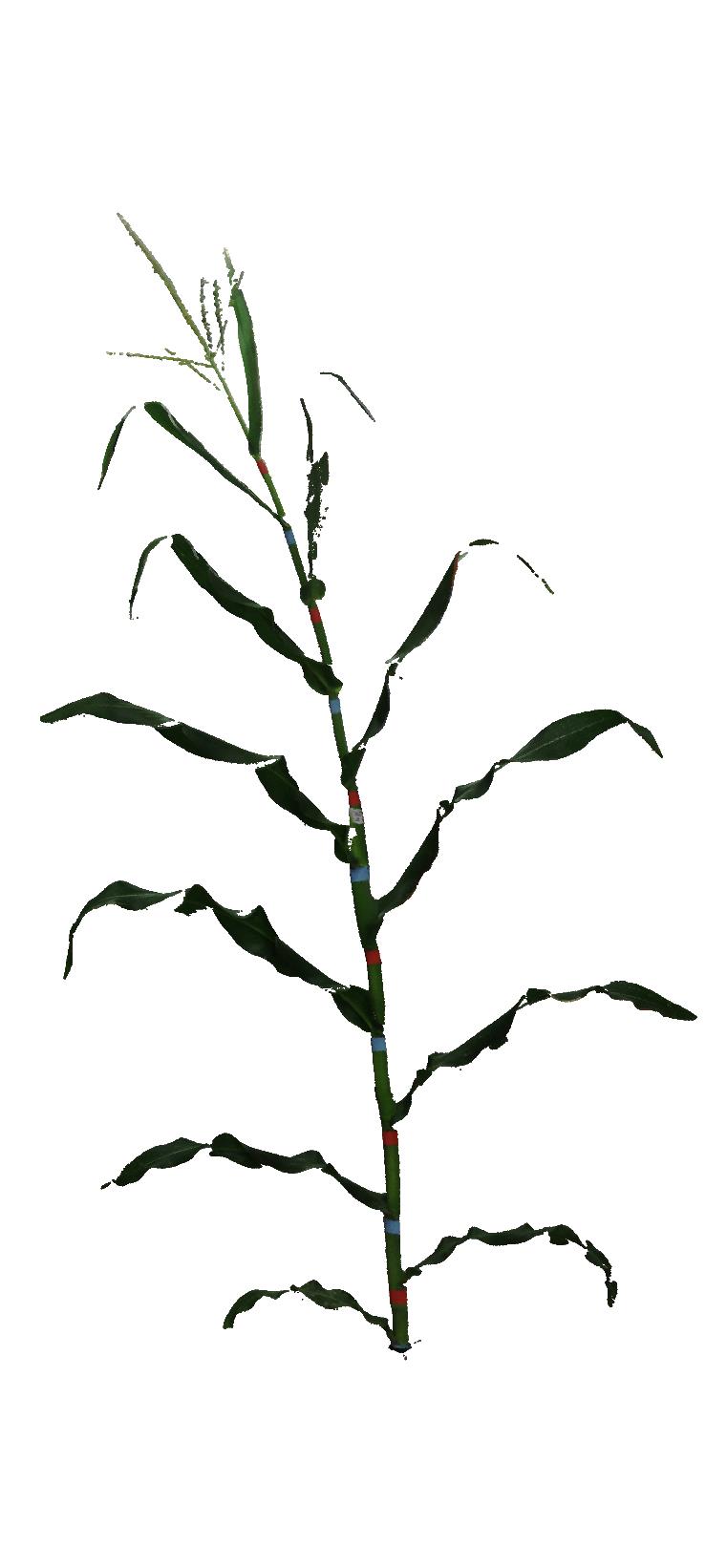}
        \caption{}
    \end{subfigure}
    \begin{subfigure}[b]{0.21\linewidth}
        \centering
        \includegraphics[trim=1cm 4cm 1cm 2cm, clip, width=\linewidth]{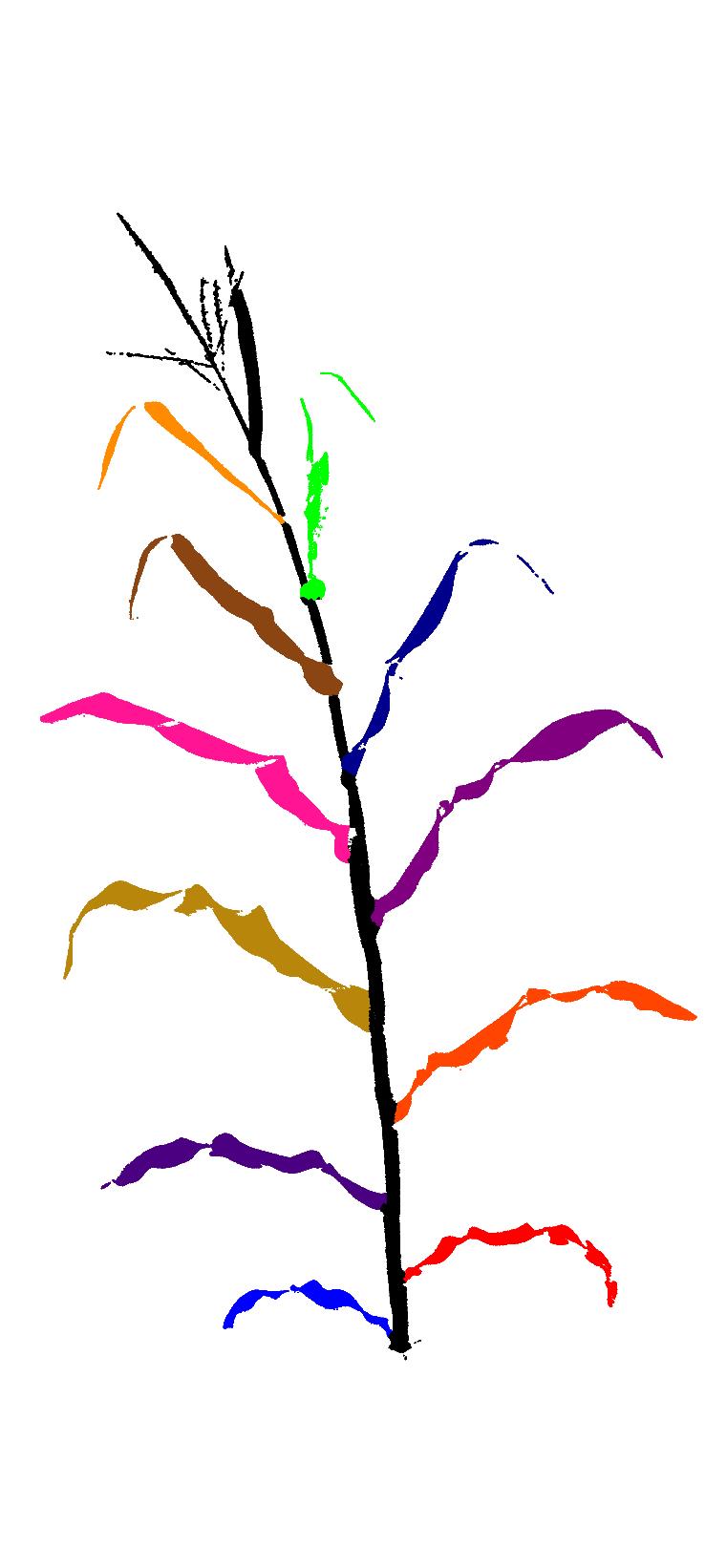}
        \caption{}
    \end{subfigure}
    \begin{subfigure}[b]{0.21\linewidth}
        \centering
        \includegraphics[trim=1cm 4cm 1cm 2cm, clip, width=\linewidth]{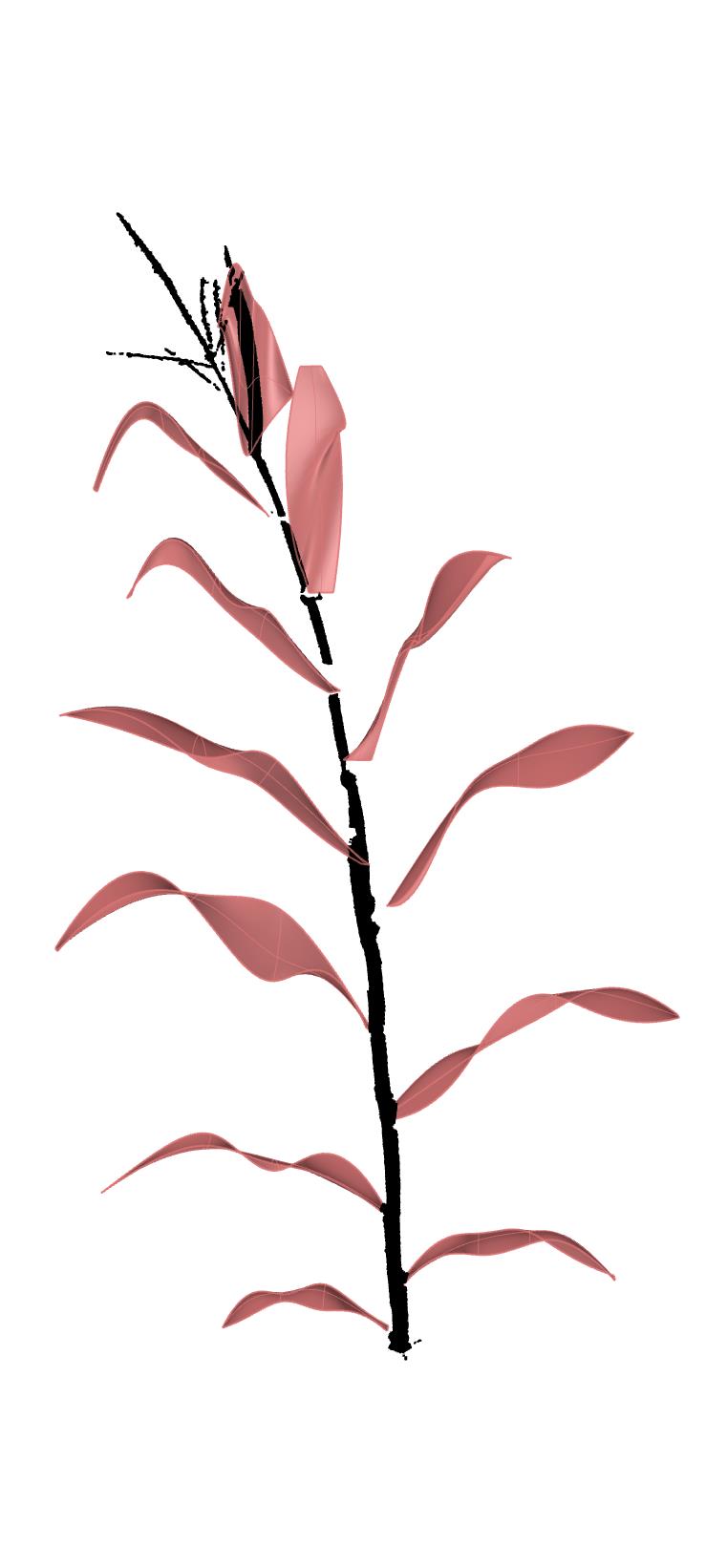}
        \caption{}
    \end{subfigure}
    \begin{subfigure}[b]{0.21\linewidth}
        \centering
        \includegraphics[trim=1cm 4cm 1cm 2cm, clip, width=\linewidth]{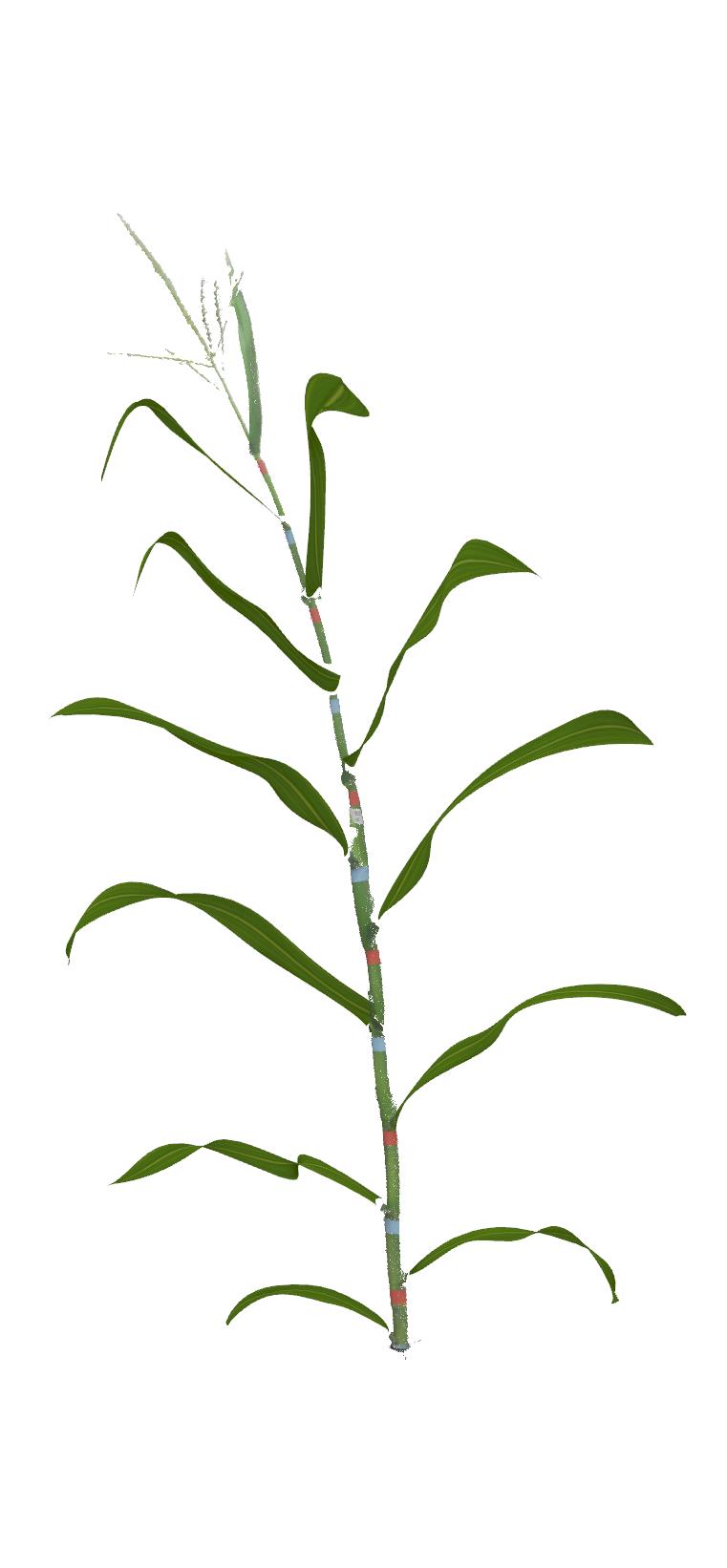}
        \caption{}
    \end{subfigure}
    \caption{Step-by-step processing of Mo17 maize plant. (a) Raw data, (b) segmented data, (c) PSO output, and (d) final NURBS surface.}
    \label{fig:Mo17}
\end{figure}

\figref{fig:B73} and \figref{fig:Mo17} shows the reconstructions for B73 and Mo17 genotypes of maize. We selected these specifically since they are among the most extensively studied maize inbred lines in genetics and plant biology. These two lines have been foundational in the development of maize hybrid breeding programs and are frequently used as representative genotypes in research~\citep{springer2007allelic, yu2008genetic}. B73 (\figref{fig:B73}) is a maize-inbred line developed at Iowa State University in the 1970s as part of the Corn Belt Dent germplasm. It is a temperate maize line with remarkably high yield potential, uniformity, and adaptability to diverse growing conditions. It served as the parent line for many first-generation commercial hybrids and has been the subject of numerous genomic studies, including the sequencing of the maize reference genome \citep{schnable2009b73}. Mo17 (\figref{fig:Mo17}), on the other hand, is another historically significant maize inbred line developed from the Lancaster Surecrop germplasm. It has complementary genetic traits to B73, including robust stress tolerance, resistance to diseases, and other agronomic traits. Mo17 is widely used in research due to its genetic diversity and its role in creating hybrids with B73. \figref{fig:B73} and \figref{fig:Mo17} demonstrate the robustness and adaptability of the proposed modeling framework, highlighting its ability to accurately capture and represent the structural diversity of different maize genotypes, which is crucial for advancing phenotypic studies and precision breeding applications.

\section{Conclusions}
\label{sec:conclusion}

The integration of PSO and NURBS-Diff in our study offers a novel approach to reconstructing 3D surfaces from point cloud data. This combination leverages the strengths of both methods: PSO efficiently provides an initial approximation, while NURBS-Diff refines this approximation to achieve high accuracy. By achieving a significant reduction in the Chamfer distance, it is evident that this two-step optimization technique greatly improves the quality of fit of NURBS surfaces to the LIDAR point cloud data. One of the main challenges encountered during this study was dealing with missing points in the point cloud data. These gaps in the data may cause surface reconstruction to be inaccurate, and our approach makes use of optimization strategies that reduce the influence of such data gaps to mitigate these problems and ensure a more accurate reconstruction. Furthermore, our approach offers the benefit of accurate trait evaluations in addition to the potential scalability to other plant species, which will enable researchers to explore the phenotypic diversity of agronomically relevant traits and conduct \textit{in silico} analyses to broaden the scope of agricultural research applications and speed their reduction to practice. 

Future research can focus on various promising directions. While this study focuses on maize plants, adapting and validating this methodology for other crop species could significantly extend its impact. Expanding the method to handle point clouds with even greater levels of noise and occlusion would enhance its robustness, especially for field-acquired datasets. Additionally, integrating this workflow with advanced machine learning models could automate the segmentation process and optimize parameter selection for broader applications. Finally, the potential for coupling 3D plant models with simulation environments opens new opportunities for studying plant-environment interactions under various stress conditions, providing valuable insights for precision agriculture.

\section*{Data Availability}
The code created for this work will be publicly available upon acceptance of the paper. 

\section*{Acknowledgements}
This work was supported by the AI Institute for Resilient Agriculture (USDA-NIFA 2021-67021-35329), NSF BTT-EAGER IOS-1842097, and Iowa State University's Plant Science Institute.  Ms. Lisa Coffey designed and managed the field experiment; Ms. Sasha Fetty and undergraduate, Hasin Khan, provided logistical support for the collection of plant materials imaged in this study; and Cheng-Ting "Eddy" Yeh assisted with data management.

\bibliographystyle{plainnat}
\bibliography{Refs}

\clearpage
\appendix

\renewcommand{\thefigure}{A.\arabic{figure}}
\setcounter{figure}{0}
\setcounter{table}{0}

\section{Additional Results}
\label{sec:sample:plants}

\begin{longtable}{| c | c | c | c | c | c | }
    \caption{Procedural models of different maize plants.}\label{tab:AllGenotypesTable}\\ 
    \hline
    \textbf{Genotype} & \textbf{Point Cloud} & \textbf{Segmented Leaves} & \textbf{PSO Output} & \textbf{Procedural Model} & \textbf{Image} \\ \hline
    \endfirsthead
    \hline
    \textbf{Genotype} & \textbf{Point Cloud} & \textbf{Segmented Leaves} & \textbf{PSO Output} & \textbf{Procedural Model} & \textbf{Image} \\ \hline
    \endhead
    \hline
    \endfoot
    \hline
    \endlastfoot
    M162W &
    \includegraphics[trim=2cm 6cm 2cm 8cm, clip, width=0.12\linewidth]{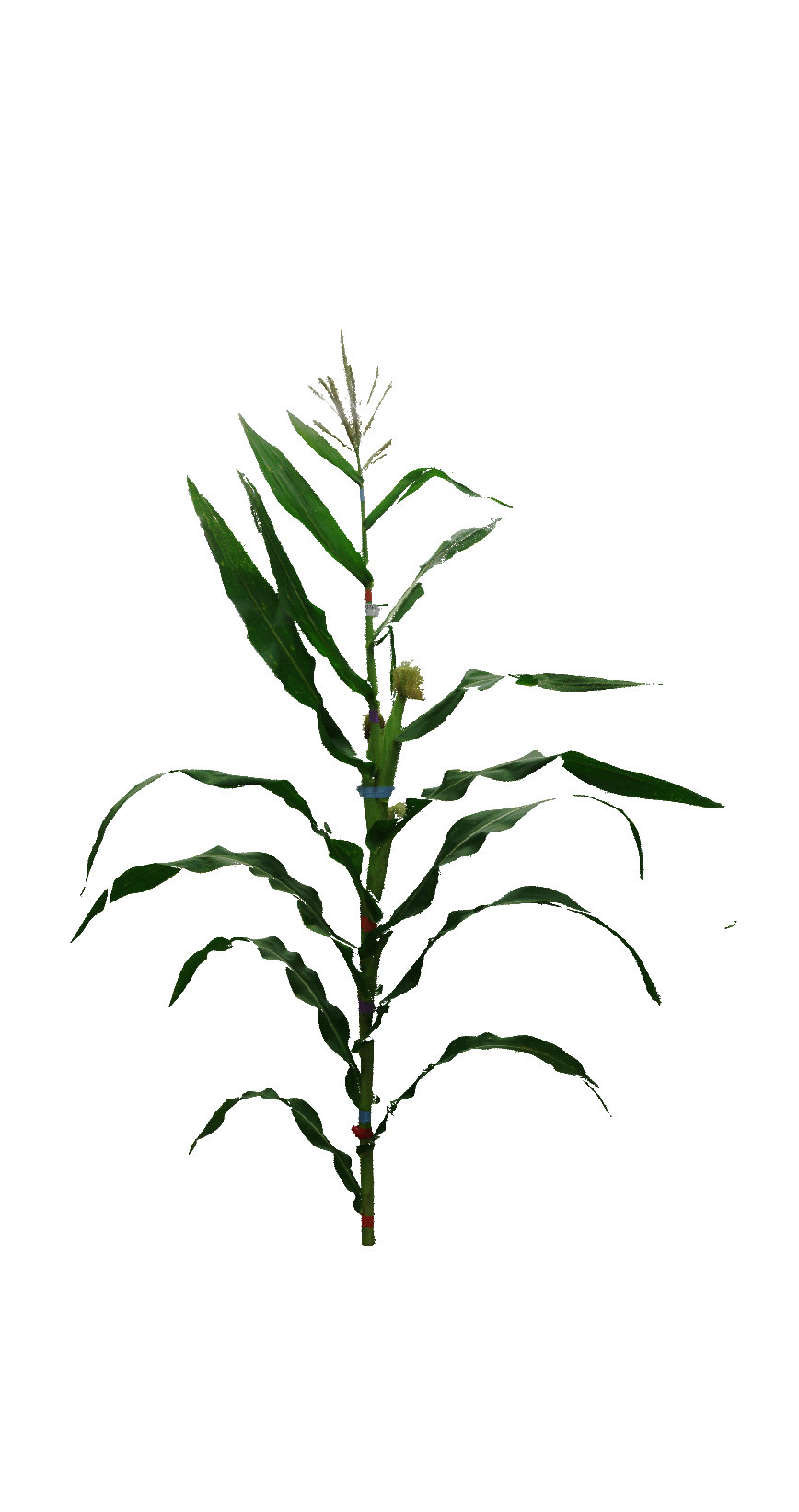} &
    \includegraphics[trim=2cm 6cm 2cm 8cm, clip, width=0.12\linewidth]{Figures_M/Plants_steps/21-JN3761-1_color.png} &
    \includegraphics[trim=2cm 6cm 2cm 8cm, clip, width=0.12\linewidth]{Figures_M/Plants_steps/21-JN3761-1_pso.png} &
    \includegraphics[trim=1cm 4cm 1cm 6cm, clip, width=0.12\linewidth]{Figures_M/Plants_steps/21-JN3761-1_nurbs1.png} &
    \includegraphics[width=0.14\linewidth]{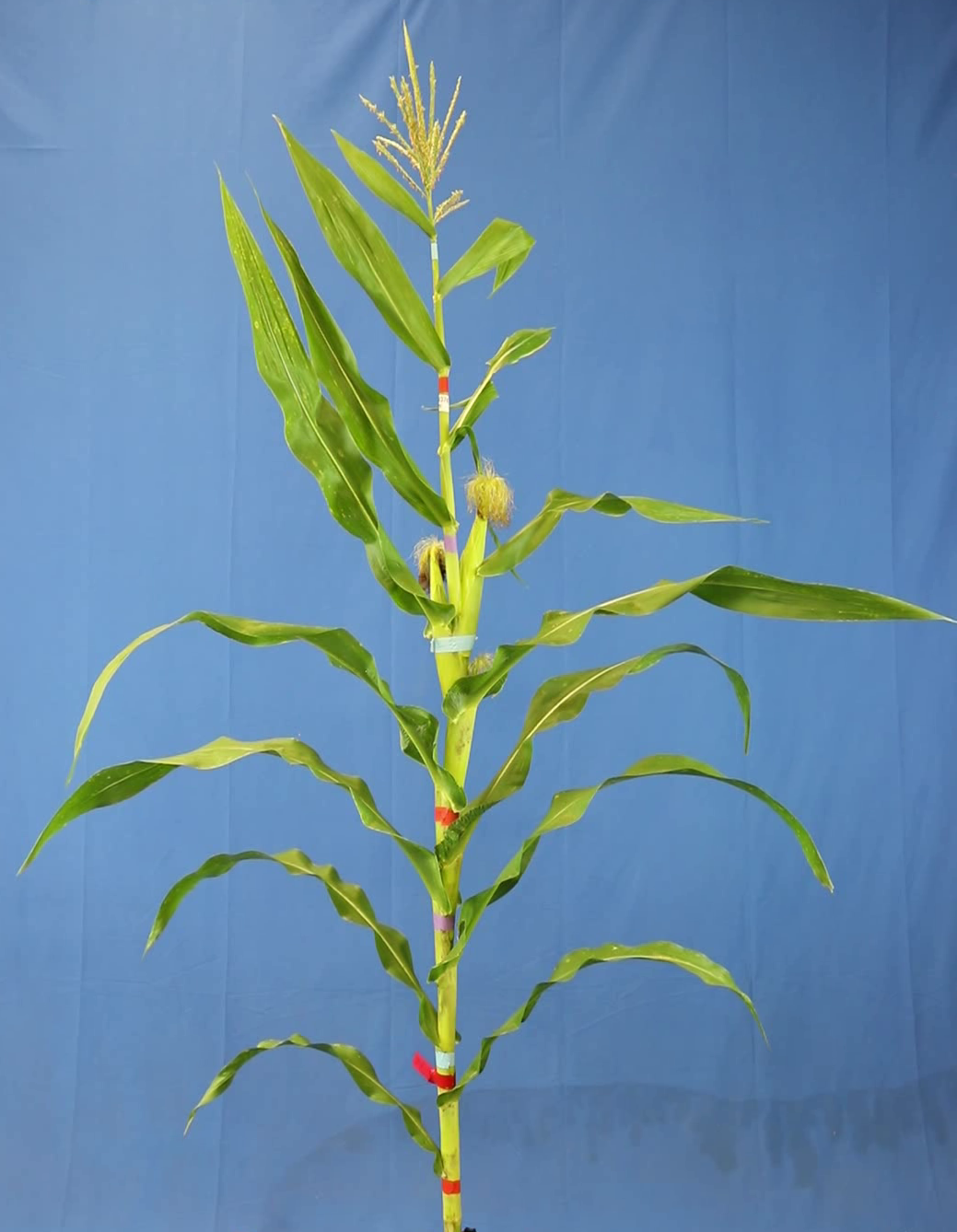} \\\hline
    CML69 &  
    \includegraphics[trim=2cm 6cm 2cm 5cm, clip, width=0.12\linewidth]{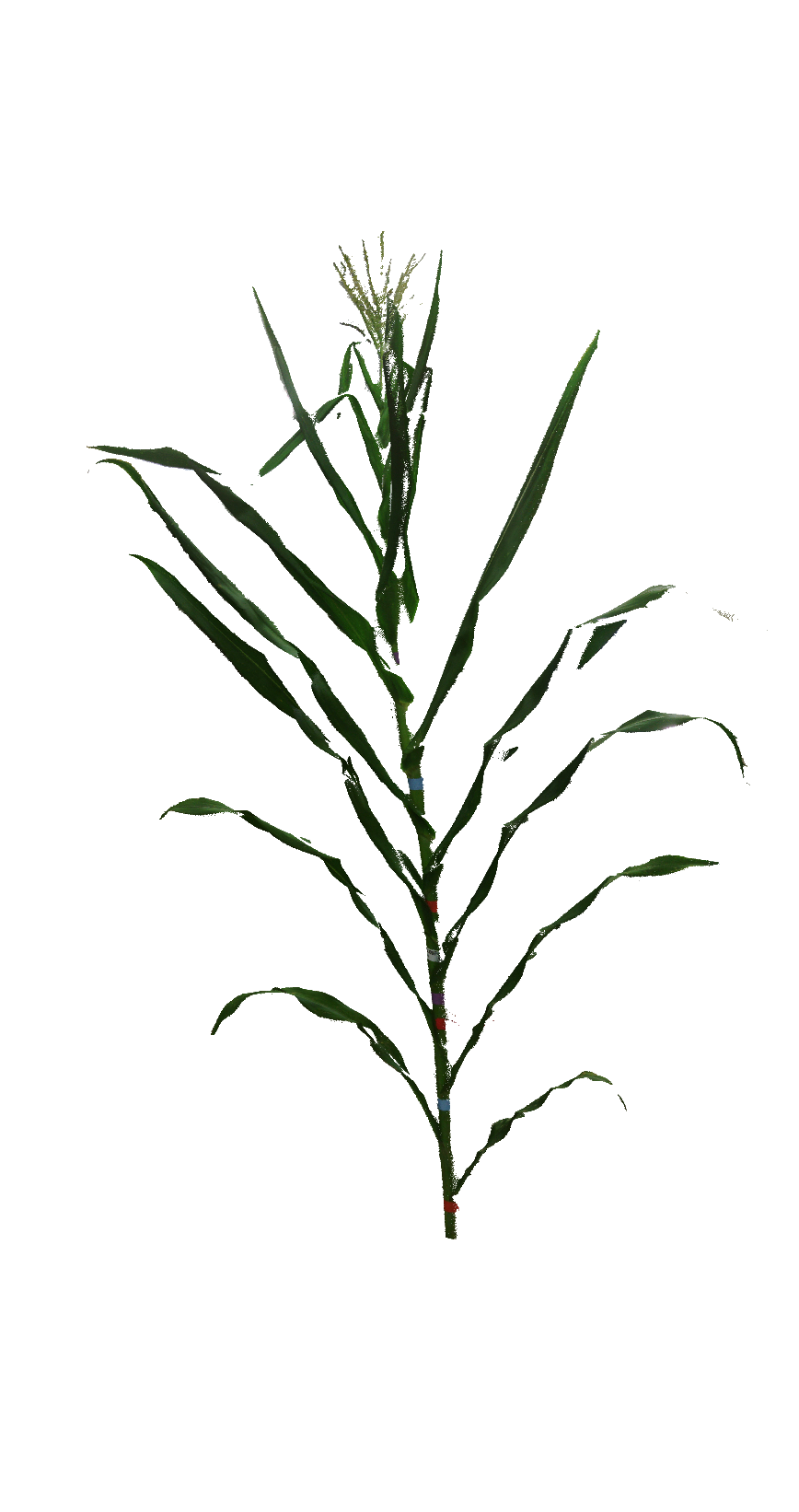} &
    \includegraphics[trim=2cm 6cm 2cm 5cm, clip, width=0.12\linewidth]{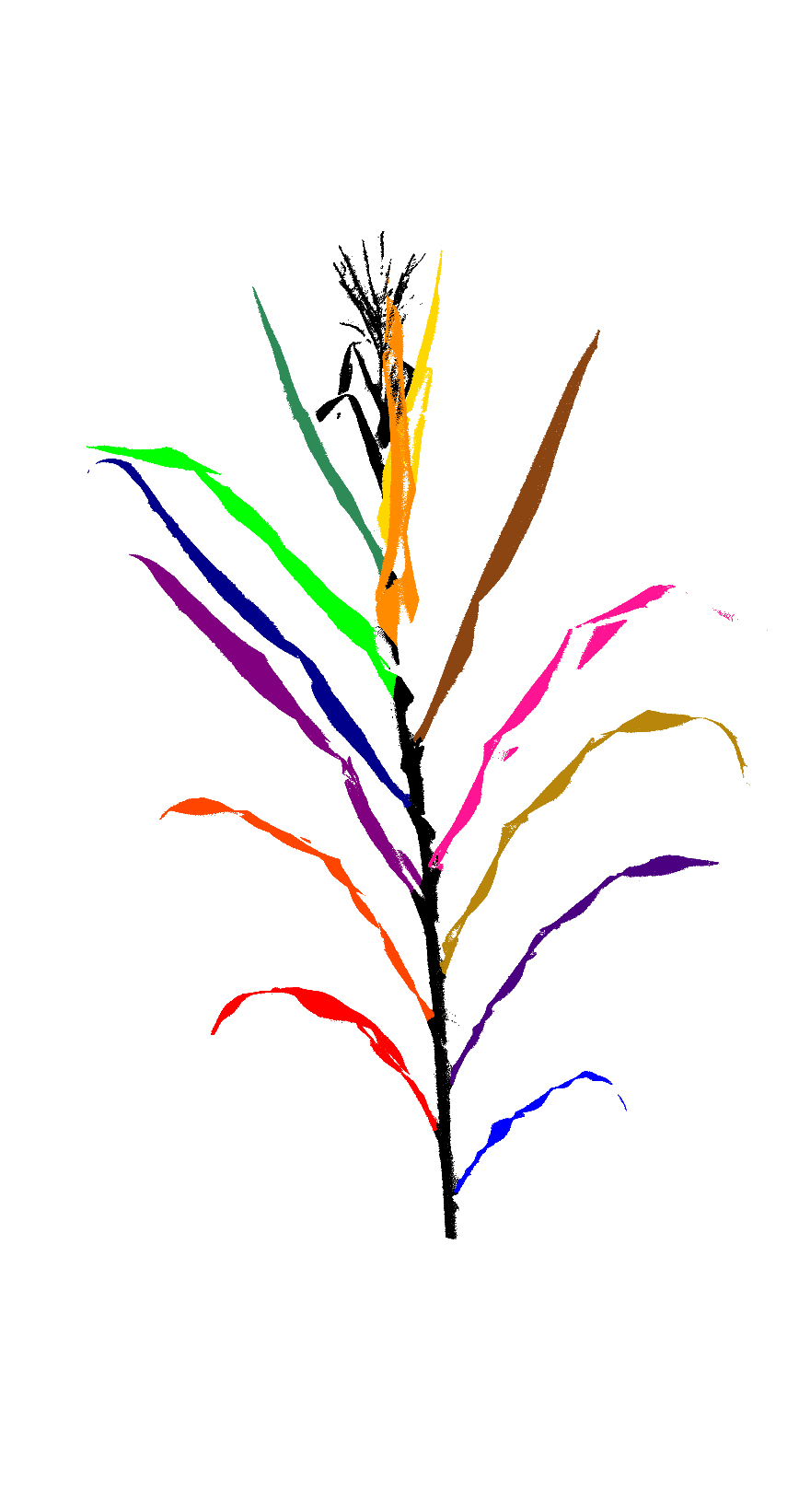} &
    \includegraphics[trim=2cm 6cm 2cm 5cm, clip, width=0.12\linewidth]{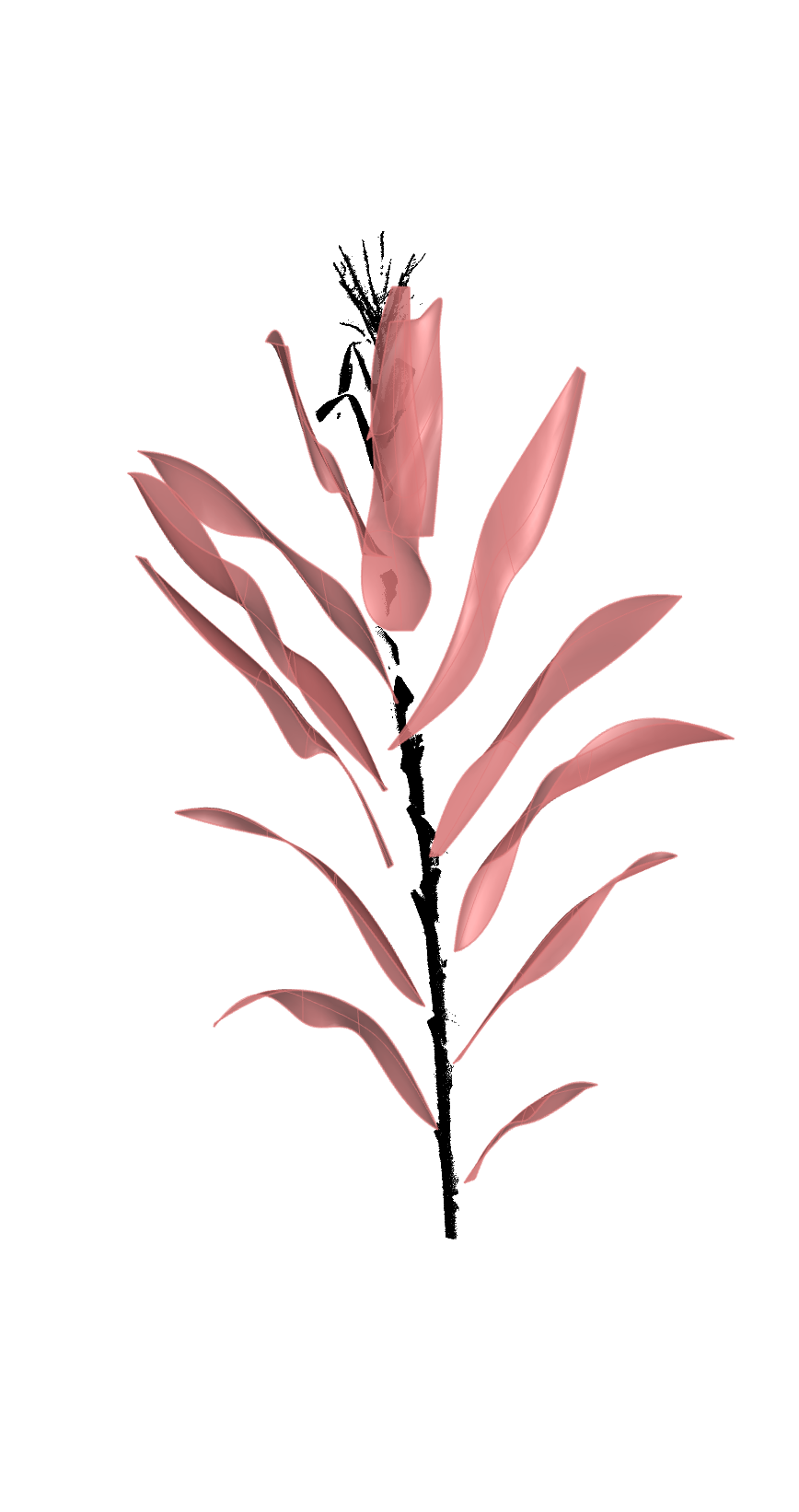} &
    \includegraphics[trim=1cm 4cm 2cm 4cm, clip, width=0.12\linewidth]{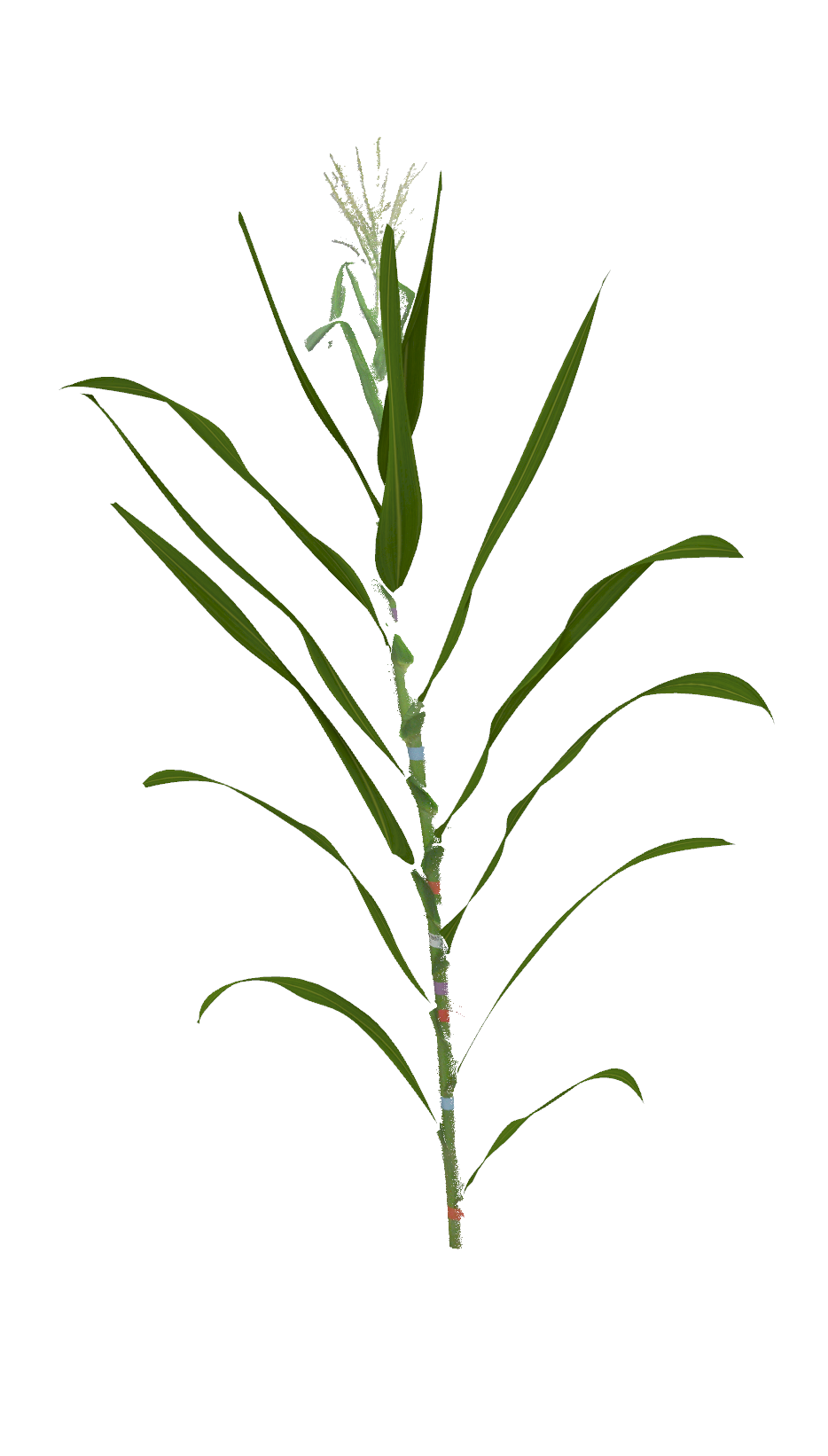} &
    \includegraphics[width=0.14\linewidth]{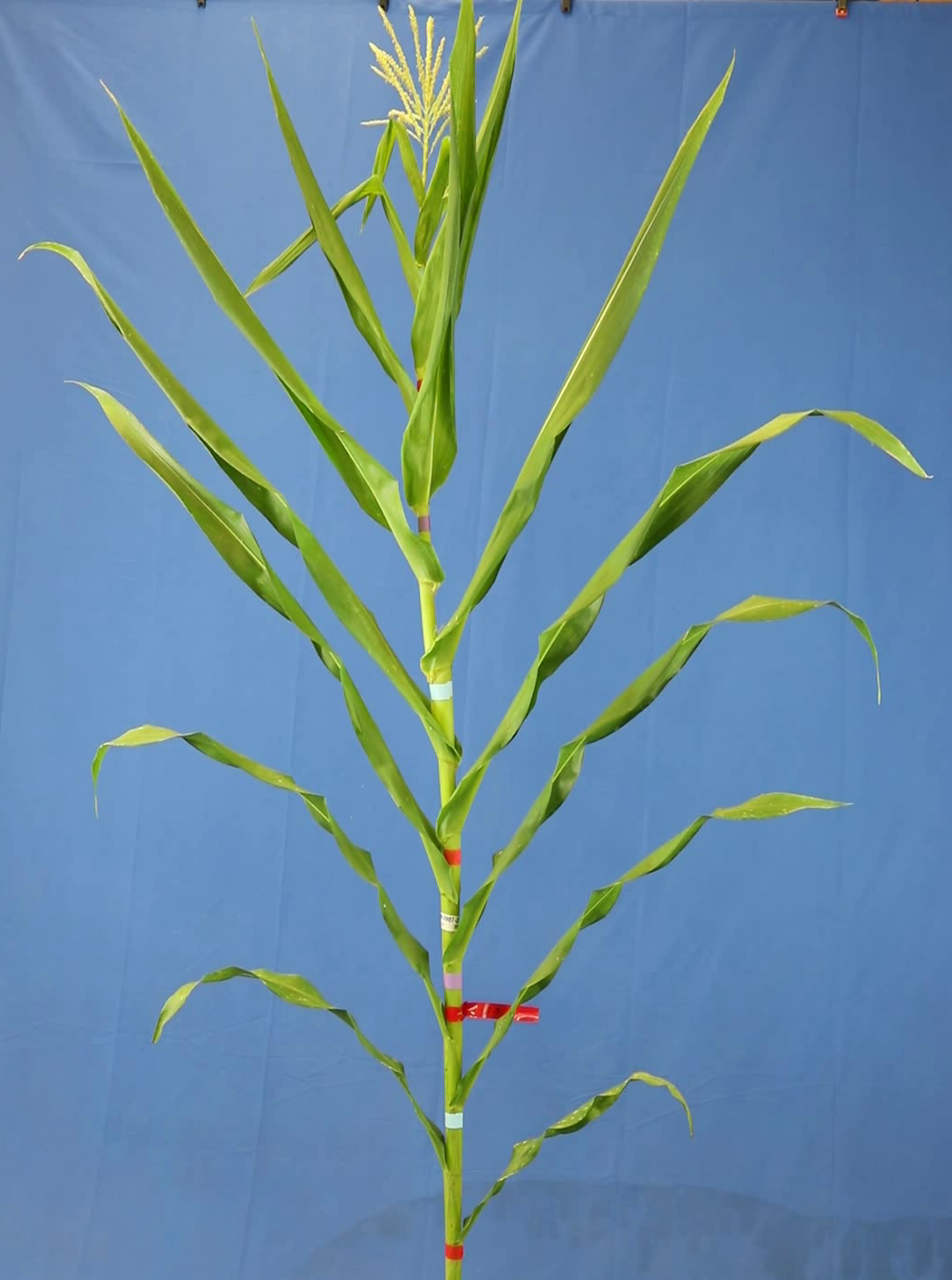} \\\hline
    Kui2021 &  
    \includegraphics[trim=2cm 6cm 2cm 5cm, clip, width=0.12\linewidth]{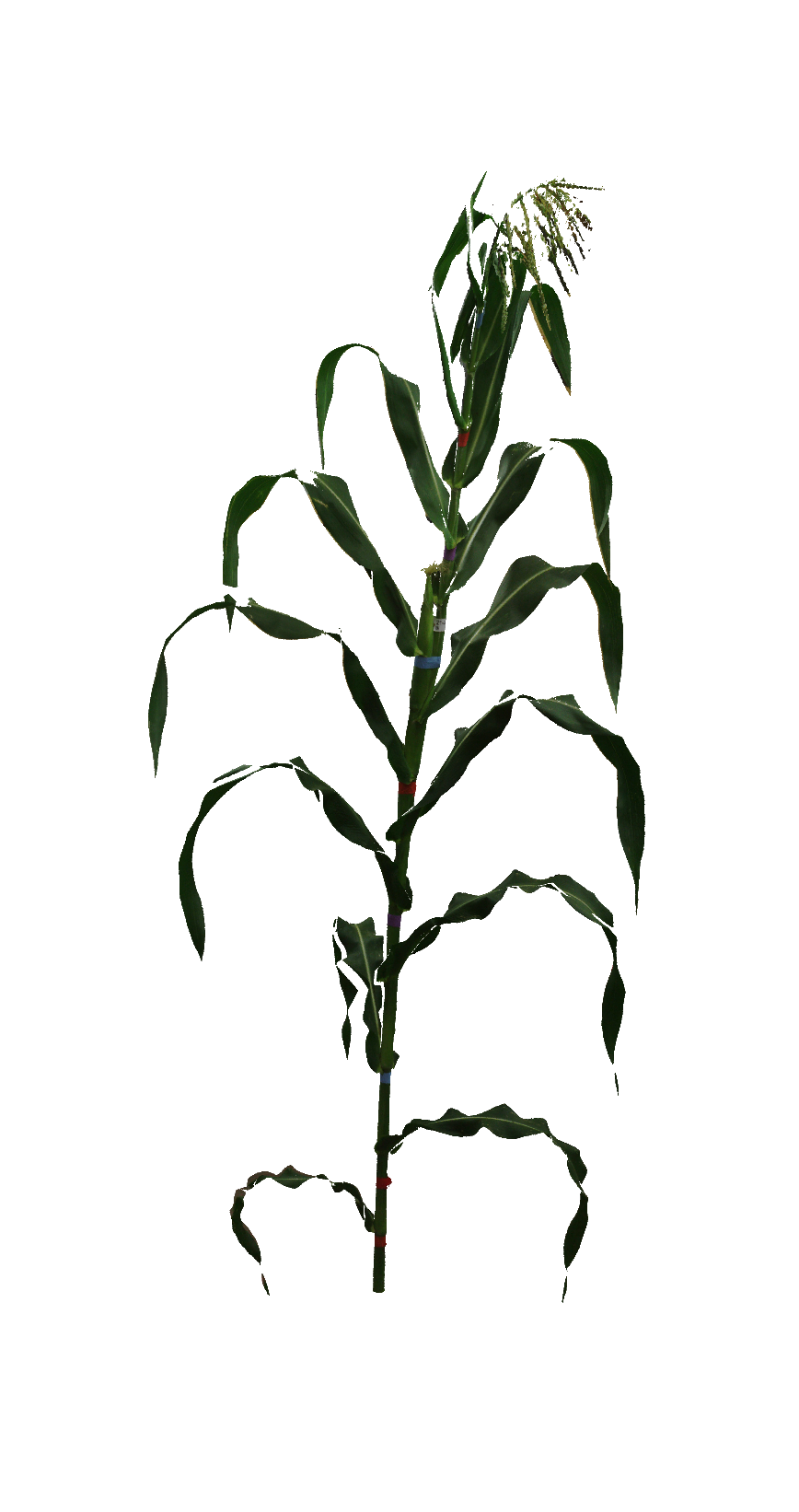} &
    \includegraphics[trim=2cm 6cm 2cm 5cm, clip, width=0.12\linewidth]{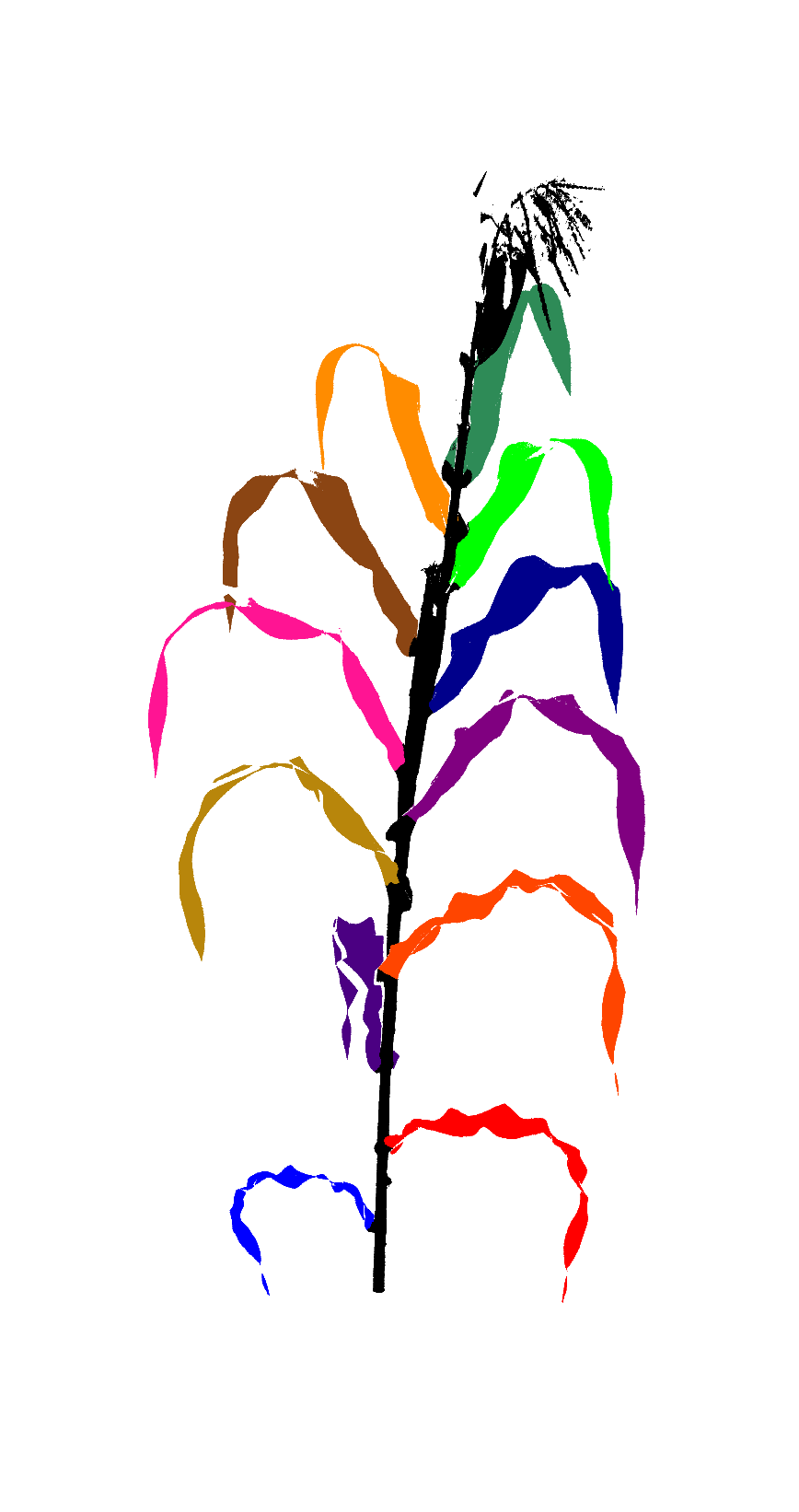} &
    \includegraphics[trim=2cm 6cm 2cm 5cm, clip, width=0.12\linewidth]{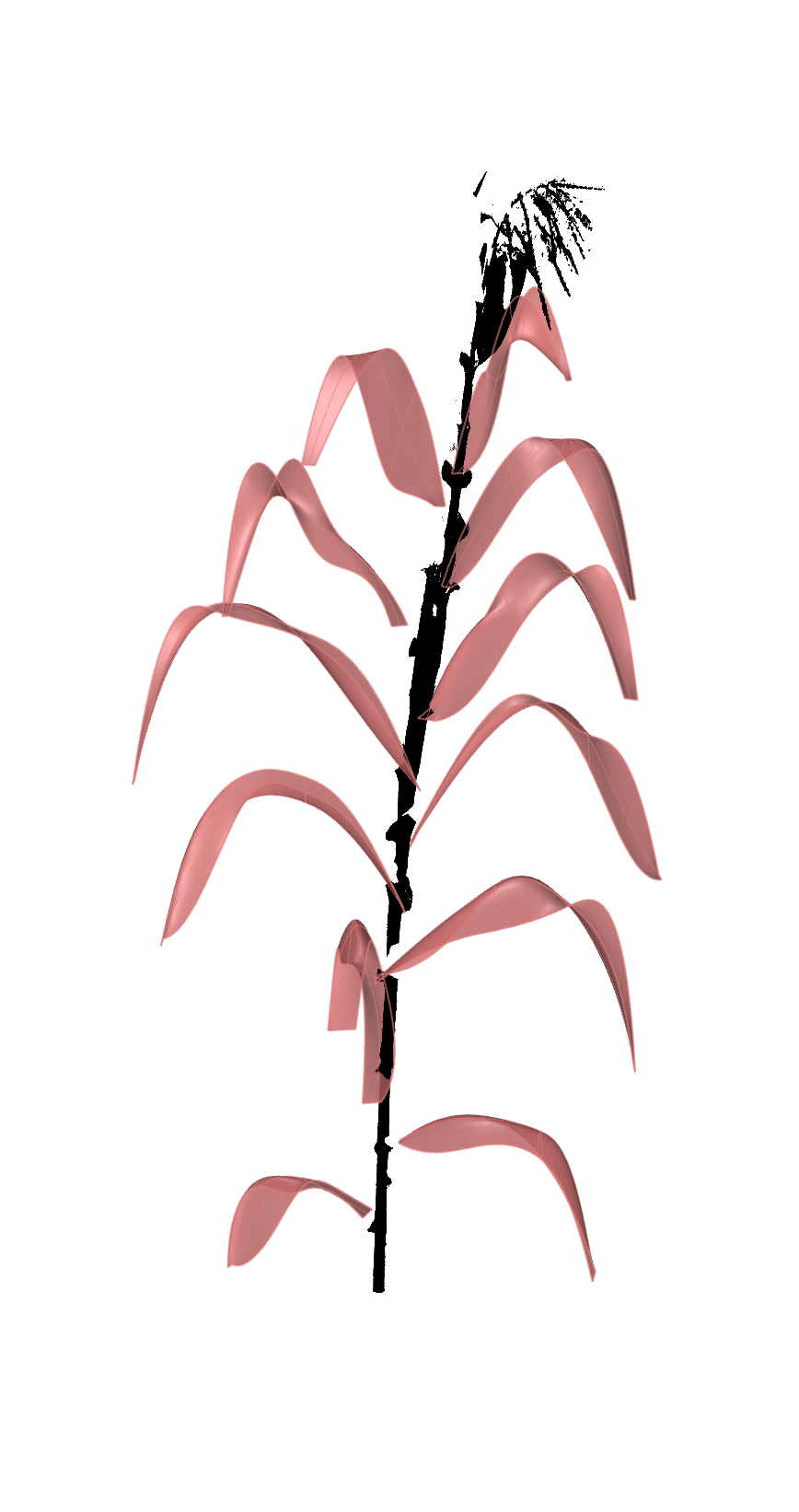} &
    \includegraphics[trim=3.5cm 6cm 3.5cm 6cm, clip, width=0.12\linewidth]{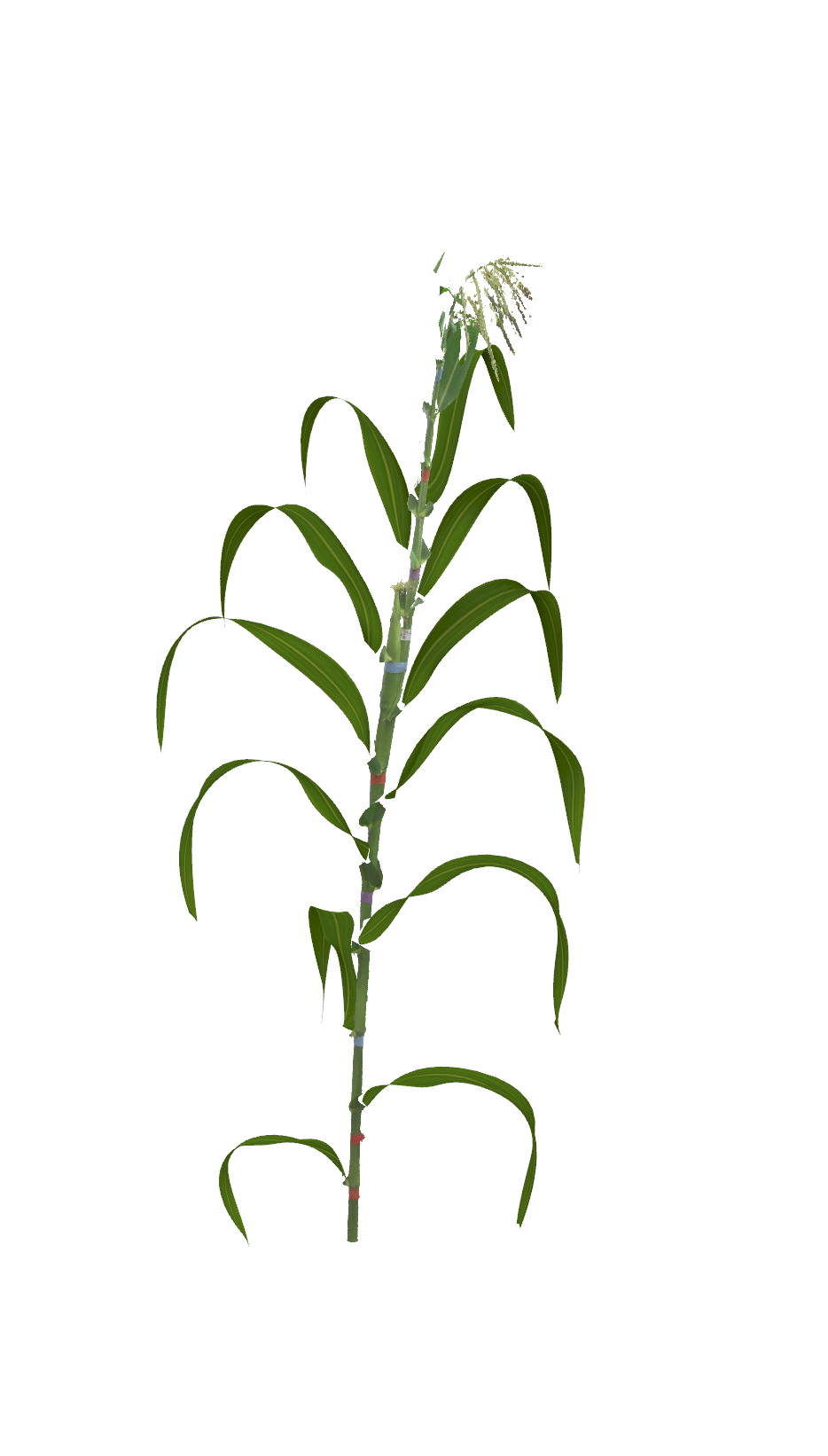} &
    \includegraphics[width=0.12\linewidth]{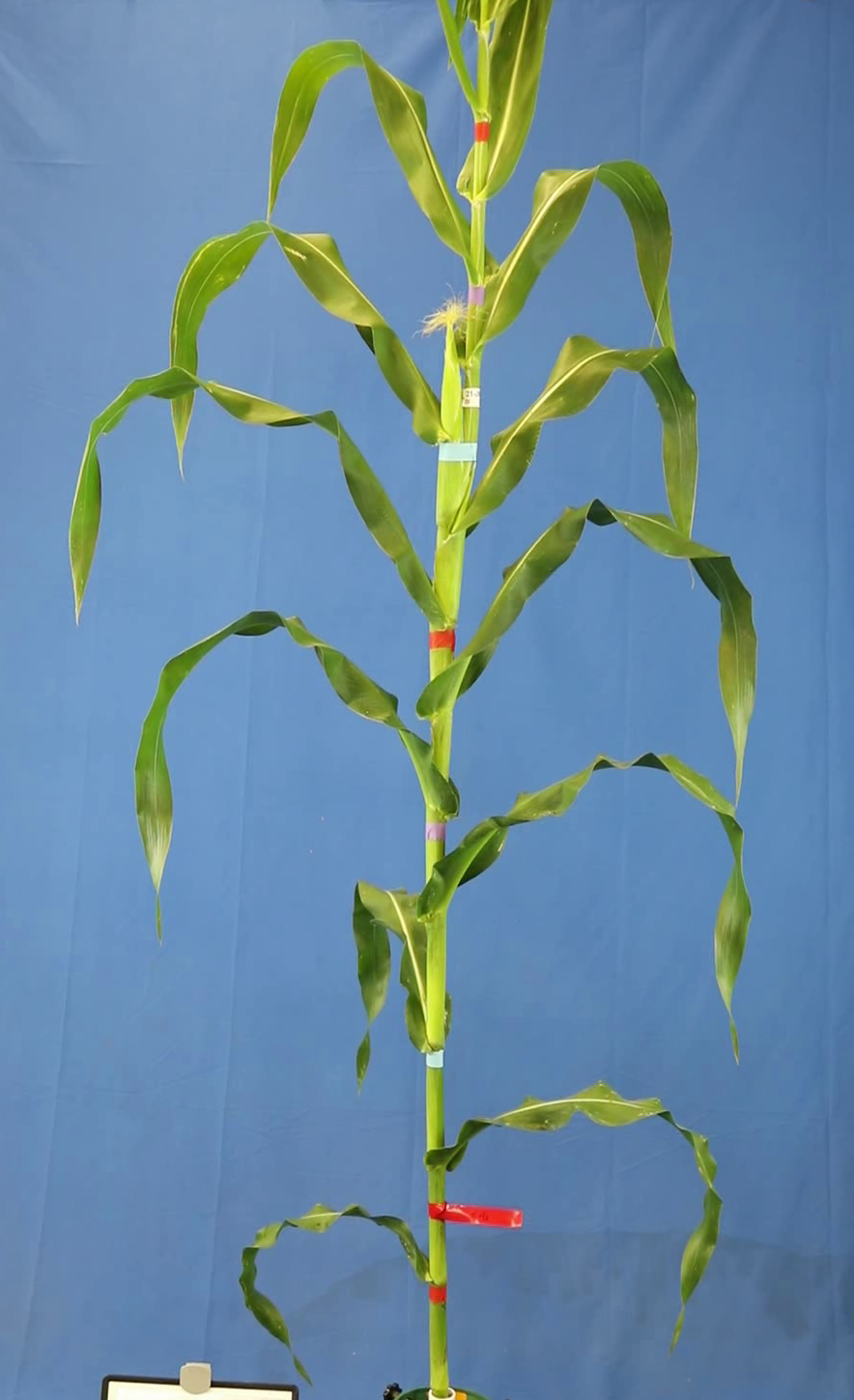} \\\hline
    CML238 &  
    \includegraphics[trim=0cm 6cm 0cm 2cm, clip, width=0.12\linewidth]{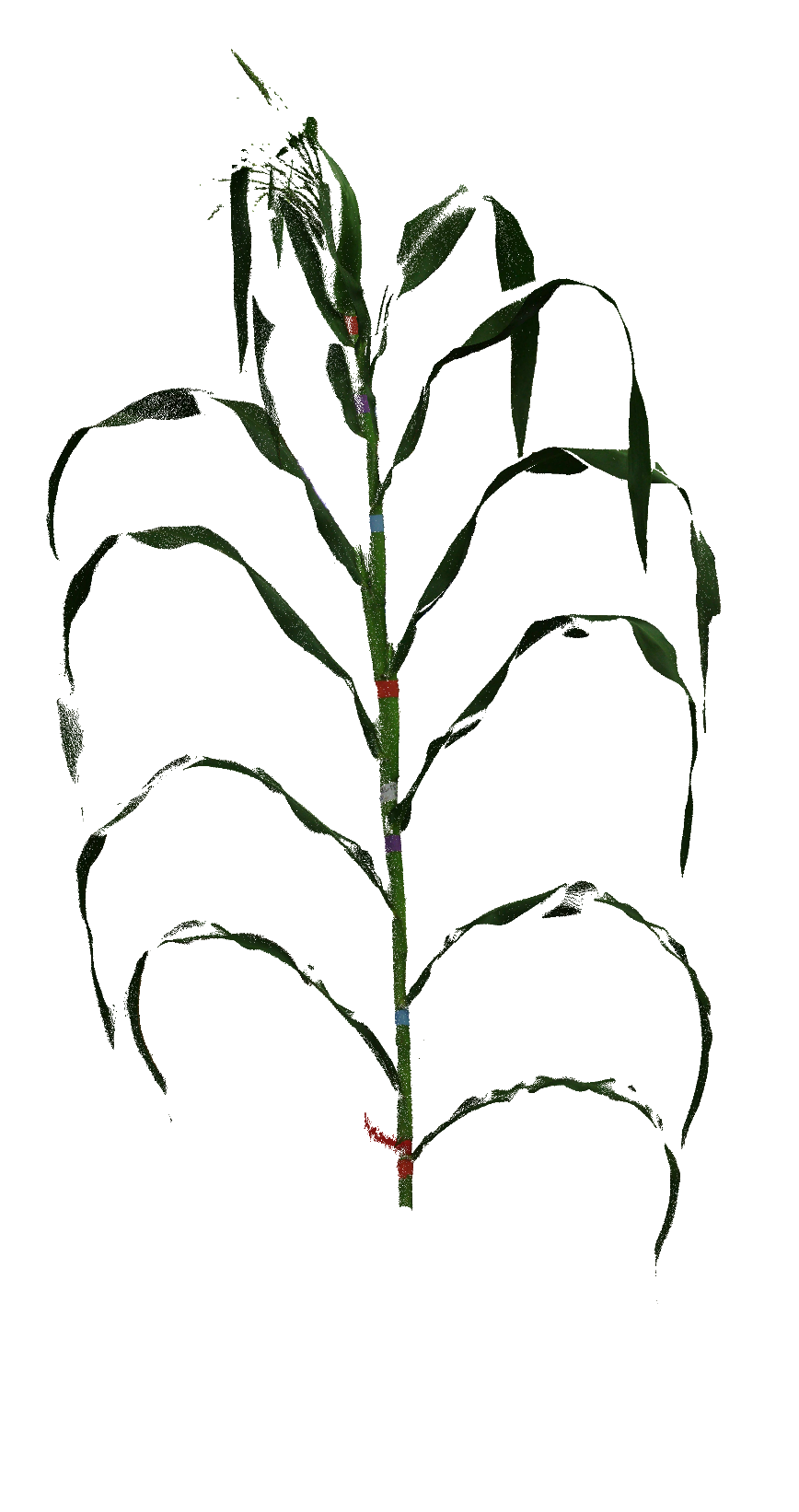} &
    \includegraphics[trim=0cm 6cm 0cm 2cm, clip, width=0.12\linewidth]{Figures_M/Plants_steps/CML238_color.png} &
    \includegraphics[trim=0cm 6cm 0cm 2cm, clip, width=0.12\linewidth]{Figures_M/Plants_steps/CML238_pso.png} &
    \includegraphics[trim=1cm 6cm 1cm 2cm, clip, width=0.12\linewidth]{Figures_M/Plants_steps/CML238_nurbs1.png} &
    \includegraphics[width=0.13\linewidth]{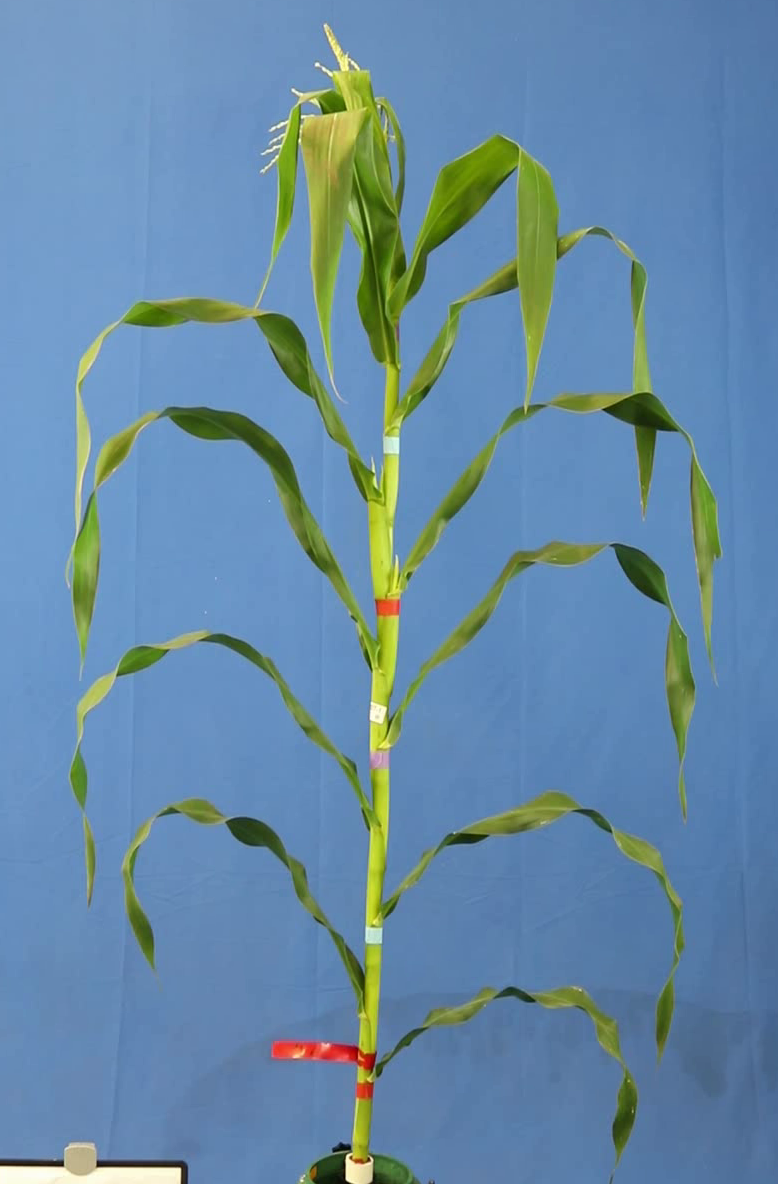} \\\hline
    CML331 &  
    \includegraphics[trim=2cm 6cm 2cm 6cm, clip, width=0.12\linewidth]{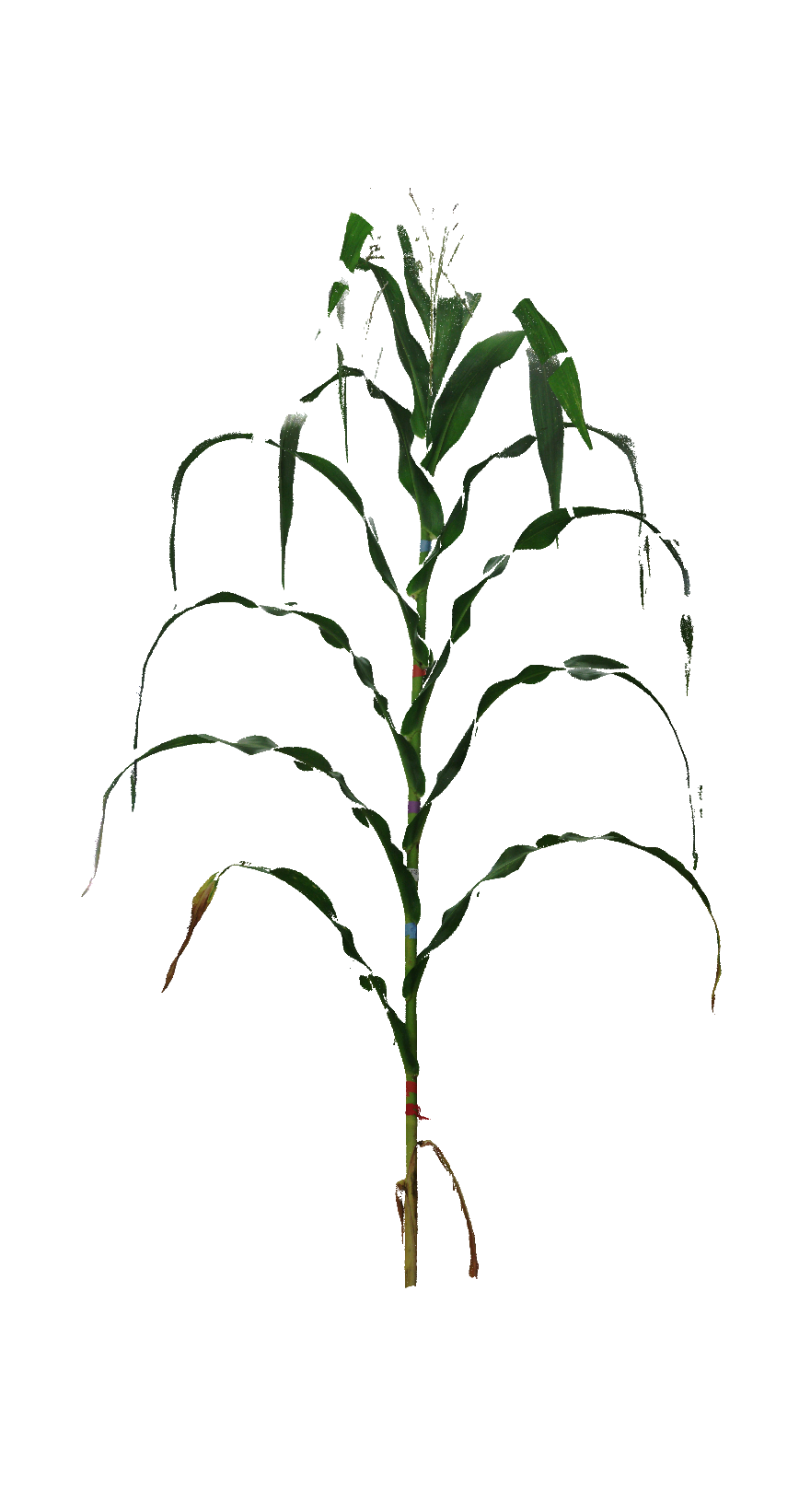} &
    \includegraphics[trim=2cm 3cm 2cm 6cm, clip, width=0.12\linewidth]{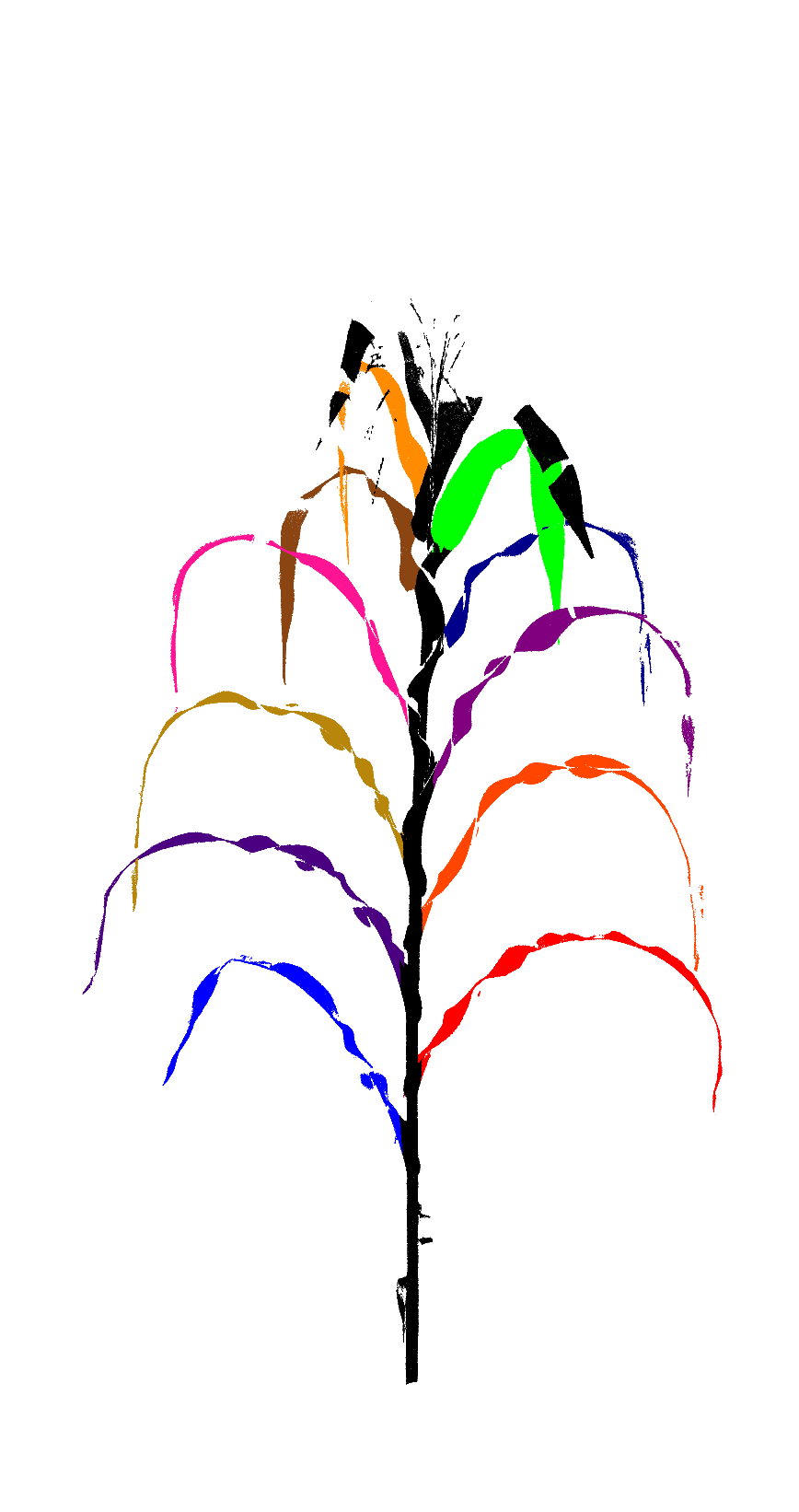} &
    \includegraphics[trim=2cm 3cm 2cm 6cm, clip, width=0.12\linewidth]{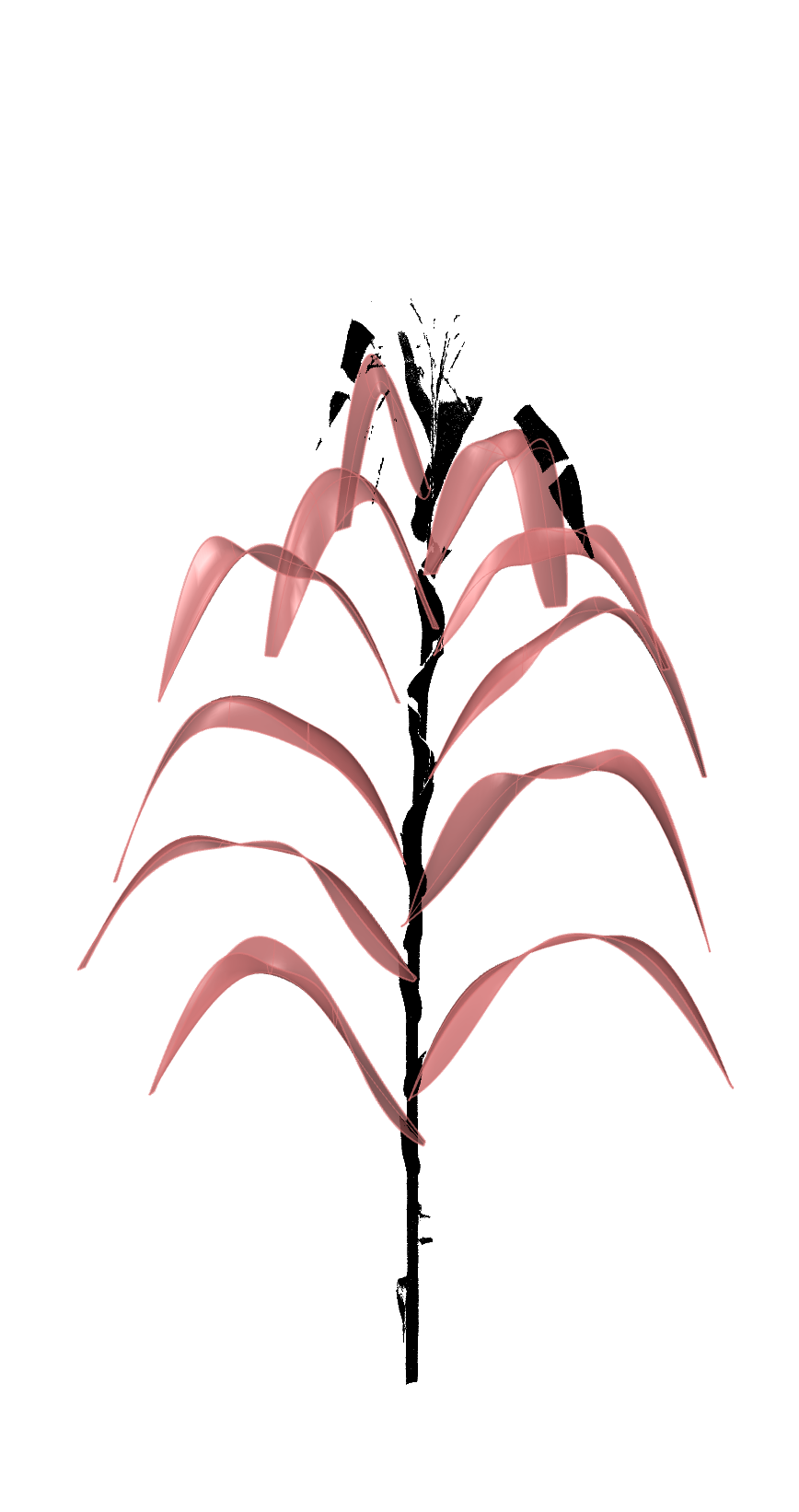} &
    \includegraphics[trim=2cm 2cm 2.5cm 5cm, clip, width=0.12\linewidth]{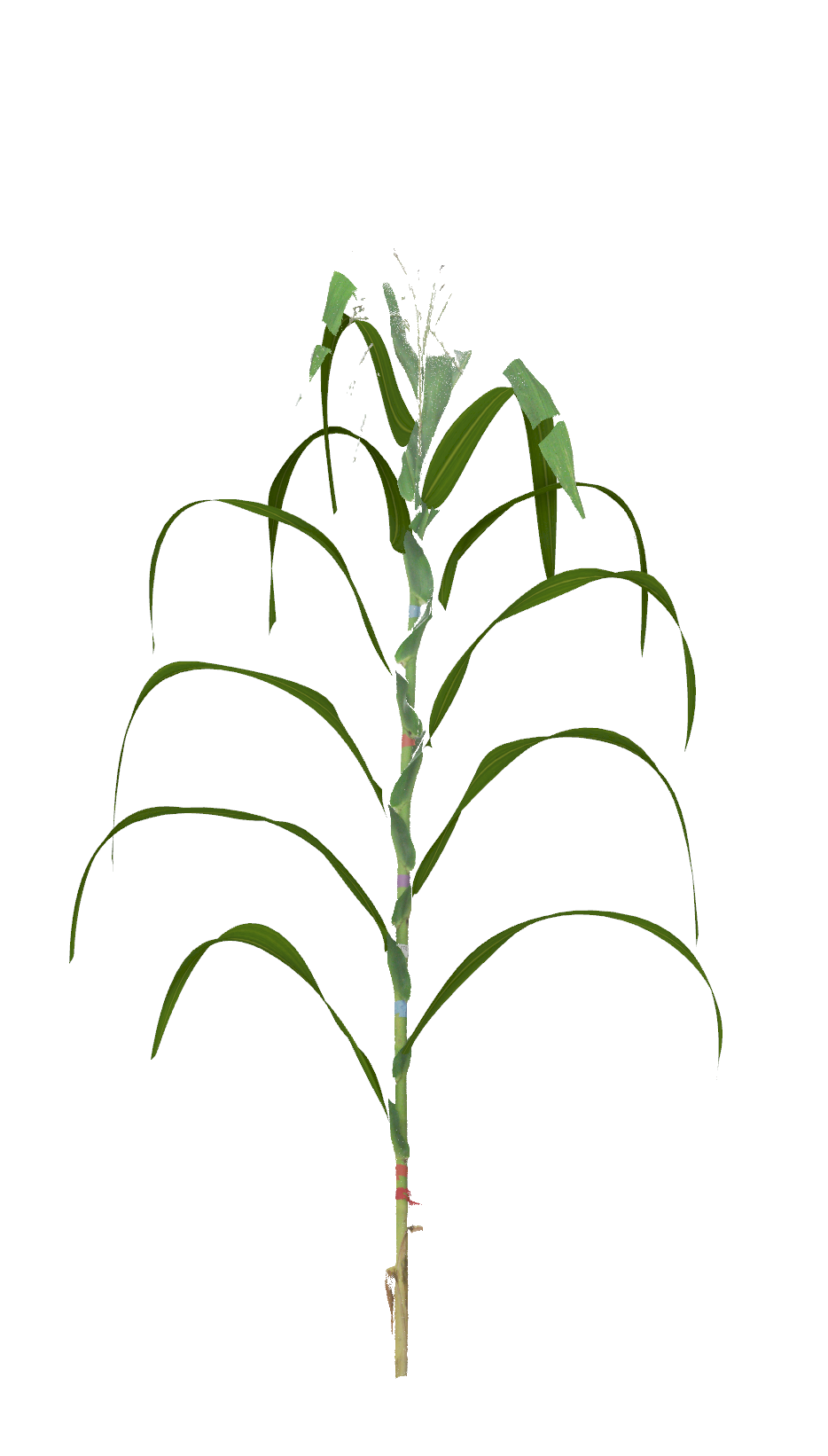} &
    \includegraphics[width=0.12\linewidth]{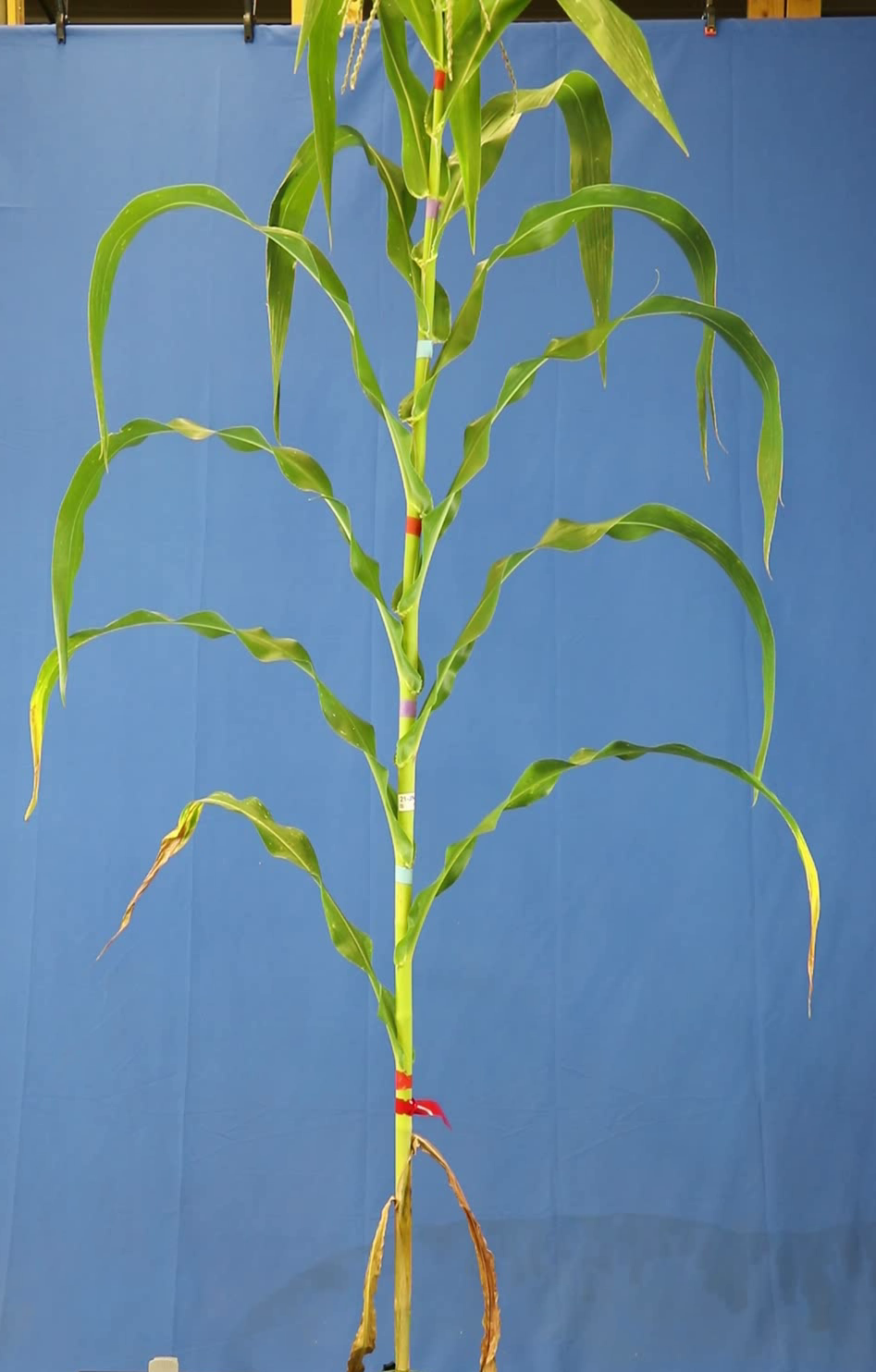} \\\hline
\end{longtable}

\begin{longtable}{| c | c | c | c | c | }
    \caption{Procedural models of different maize plants.}\label{tab:AllGenotypesTable2}\\ 
    \hline
    \textbf{Genotype} & \textbf{Point Cloud} & \textbf{Segmented Leaves} & \textbf{PSO Output} & \textbf{Procedural Model} \\ \hline
    \endfirsthead
    \hline
    \textbf{Genotype} & \textbf{Point Cloud} & \textbf{Segmented Leaves} & \textbf{PSO Output} & \textbf{Procedural Model} \\ \hline
    \endhead
    \hline
    \endfoot
    \hline
    \endlastfoot    
    4226 &  
    \includegraphics[trim=2cm 5cm 2cm 6cm, clip, width=0.12\linewidth]{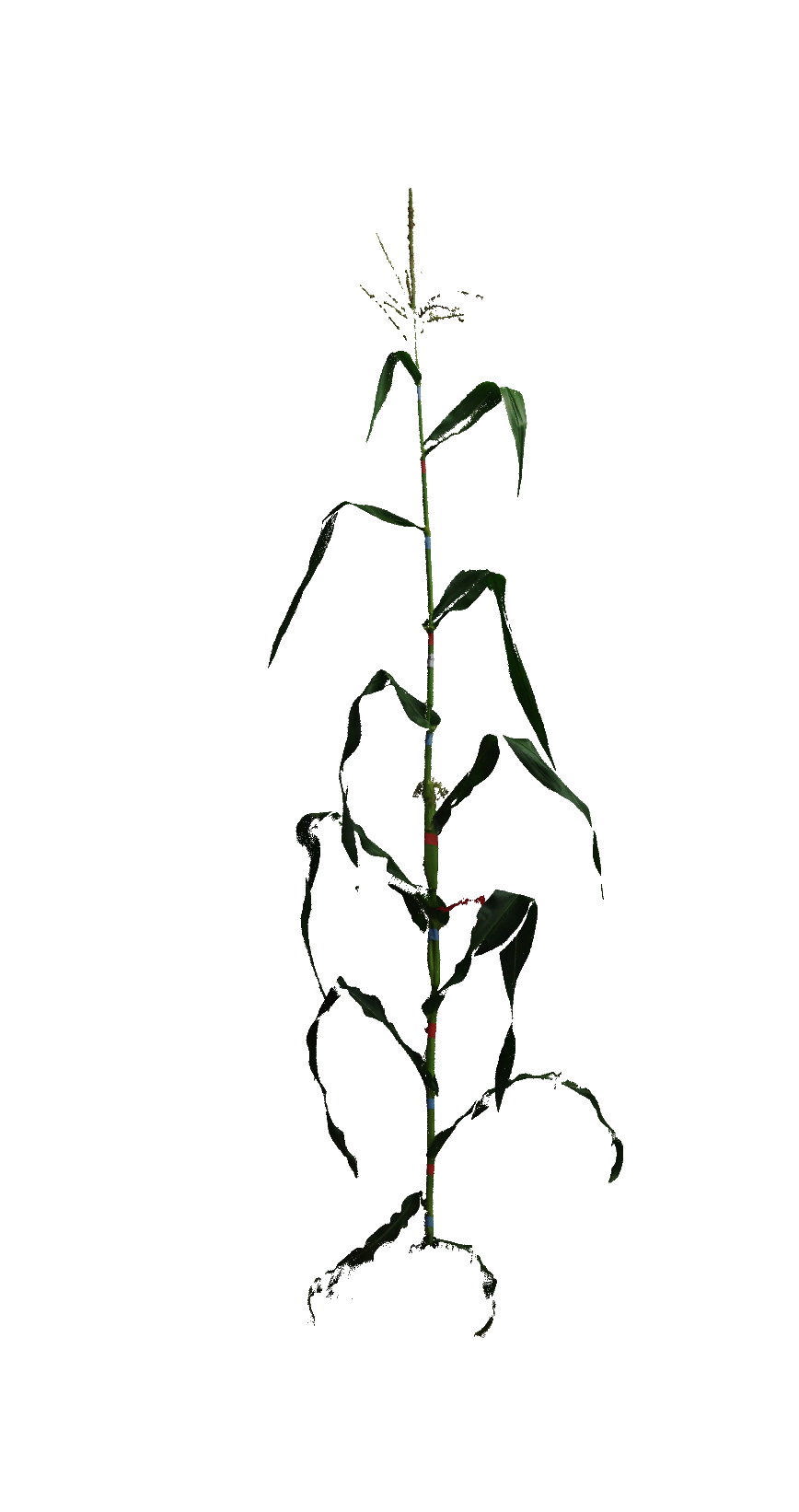} &
    \includegraphics[trim=2cm 5cm 2cm 6cm, clip, width=0.12\linewidth]{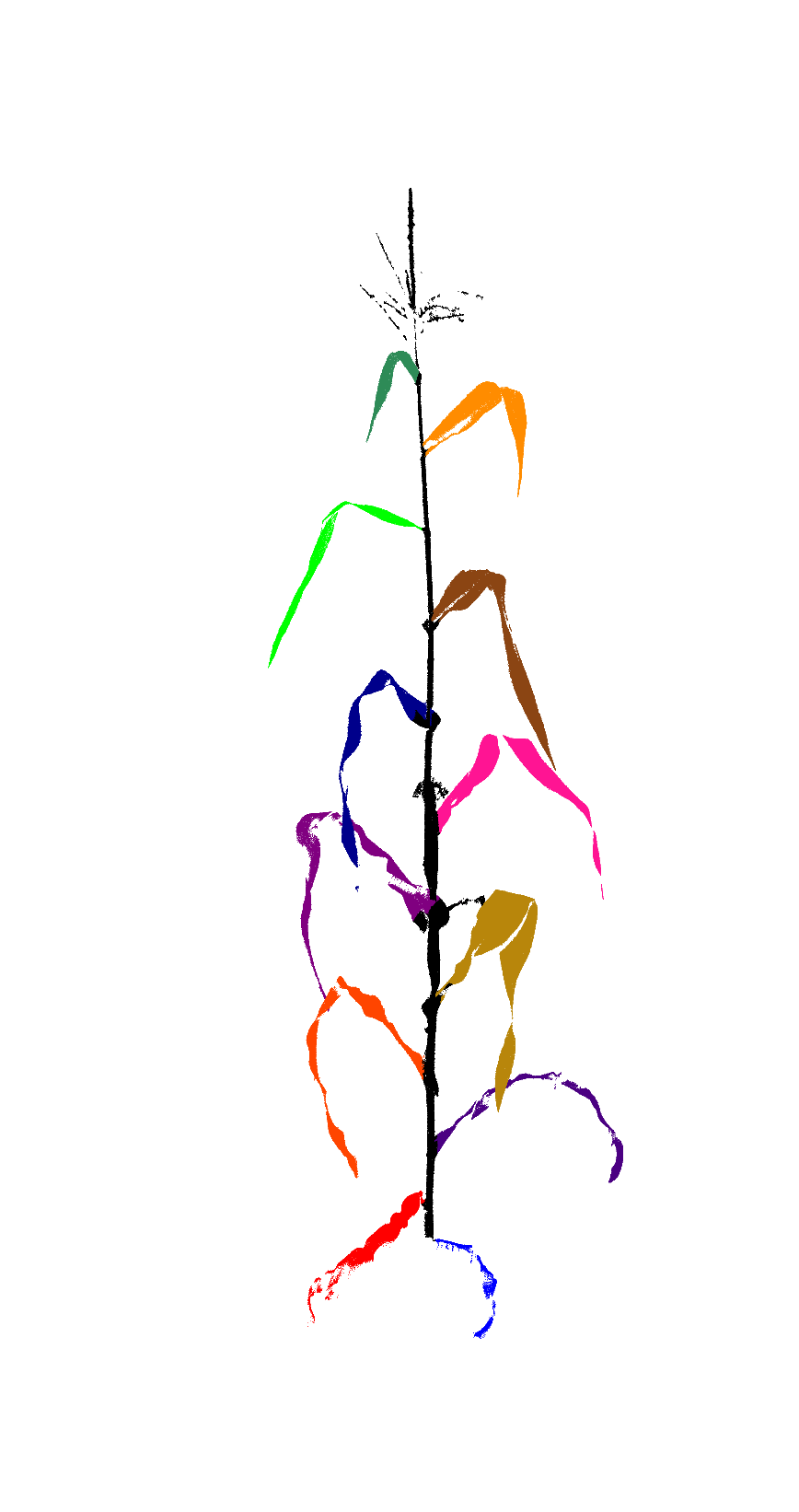} &
    \includegraphics[trim=2cm 5cm 2cm 6cm, clip, width=0.12\linewidth]{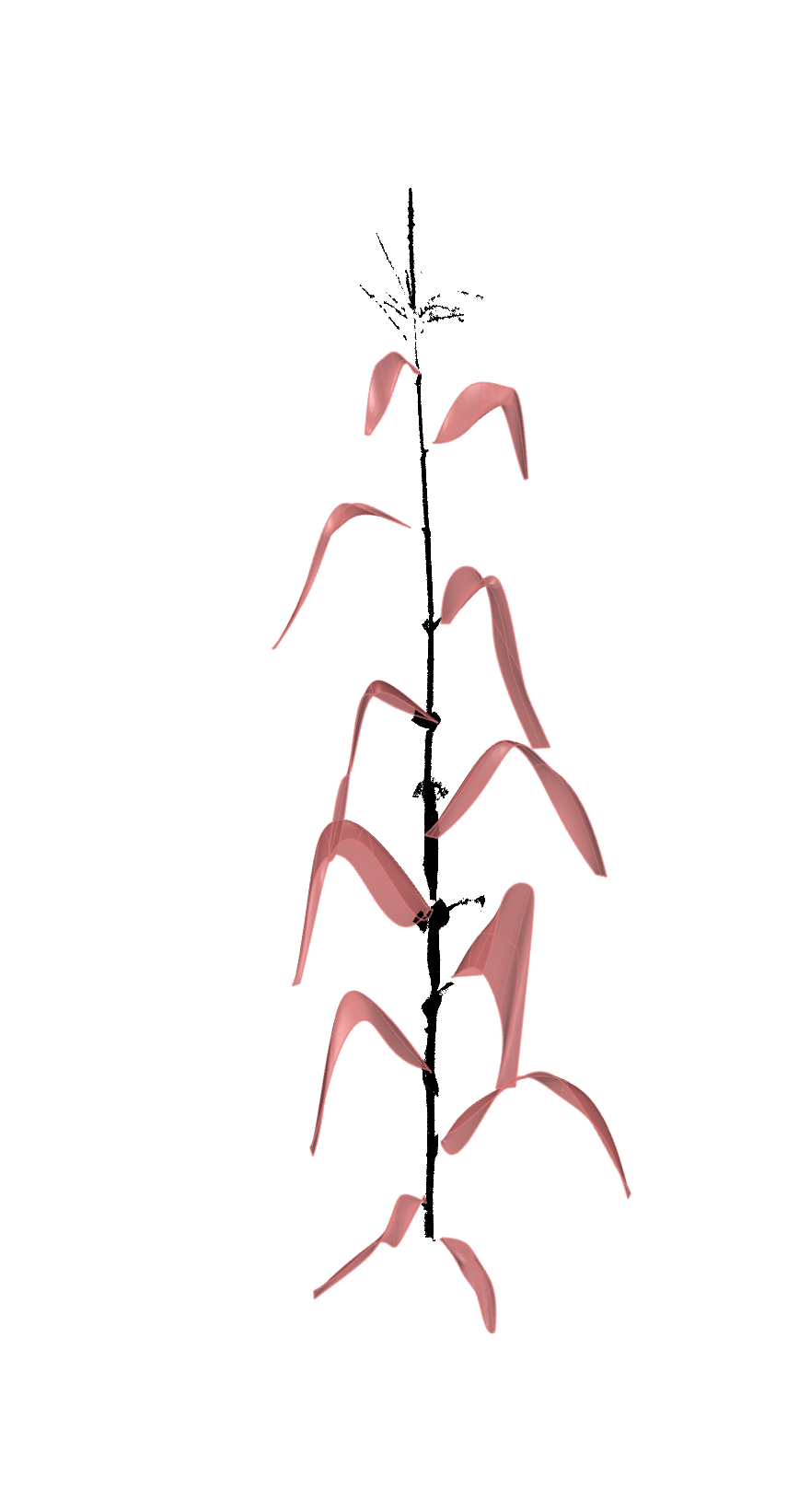} &
    \includegraphics[trim=3cm 5cm 2cm 4cm, clip, width=0.12\linewidth]{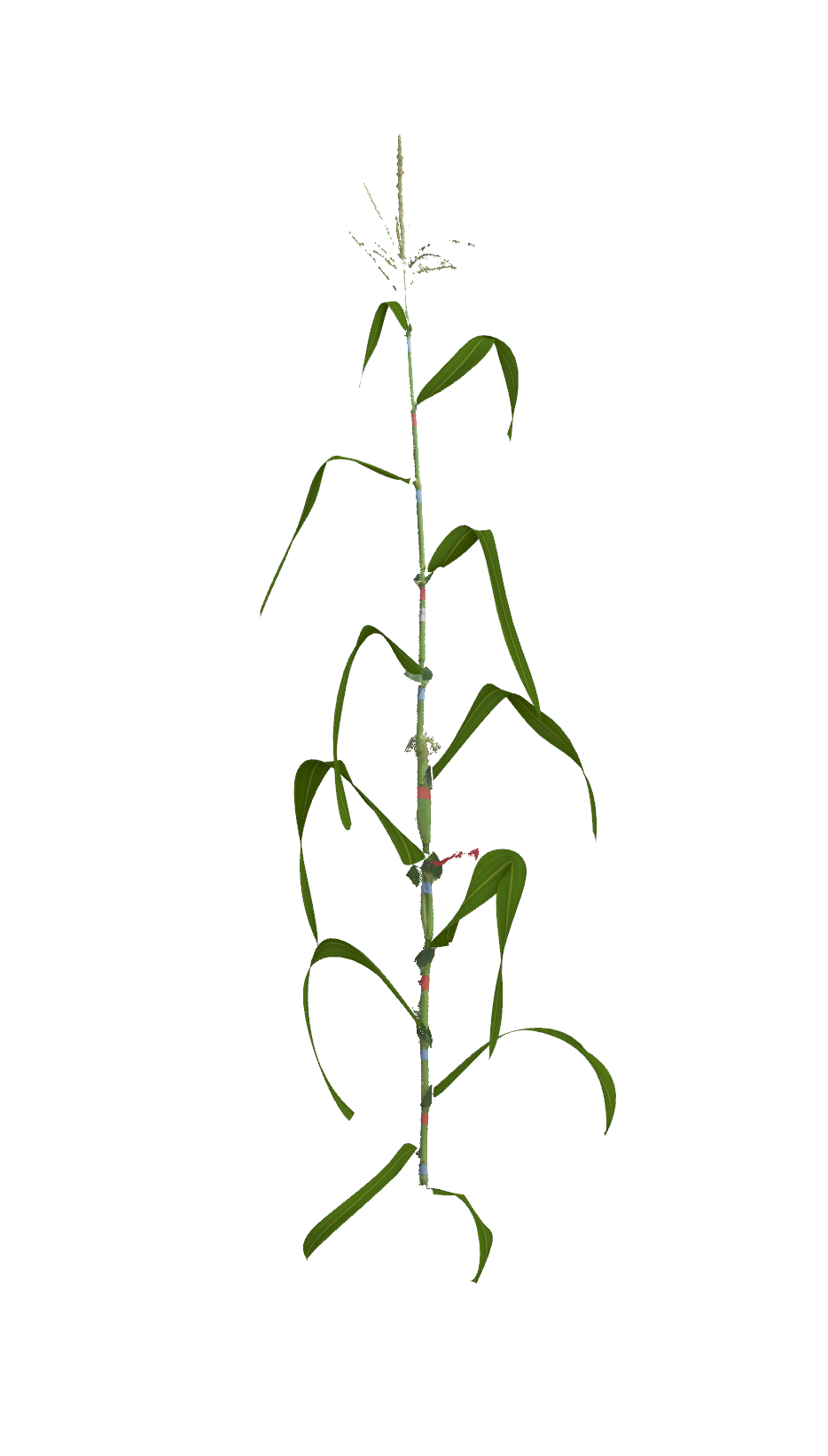} \\ 
    \hline
    78551S &  
    \includegraphics[trim=2cm 6cm 2cm 5cm, clip, width=0.12\linewidth]{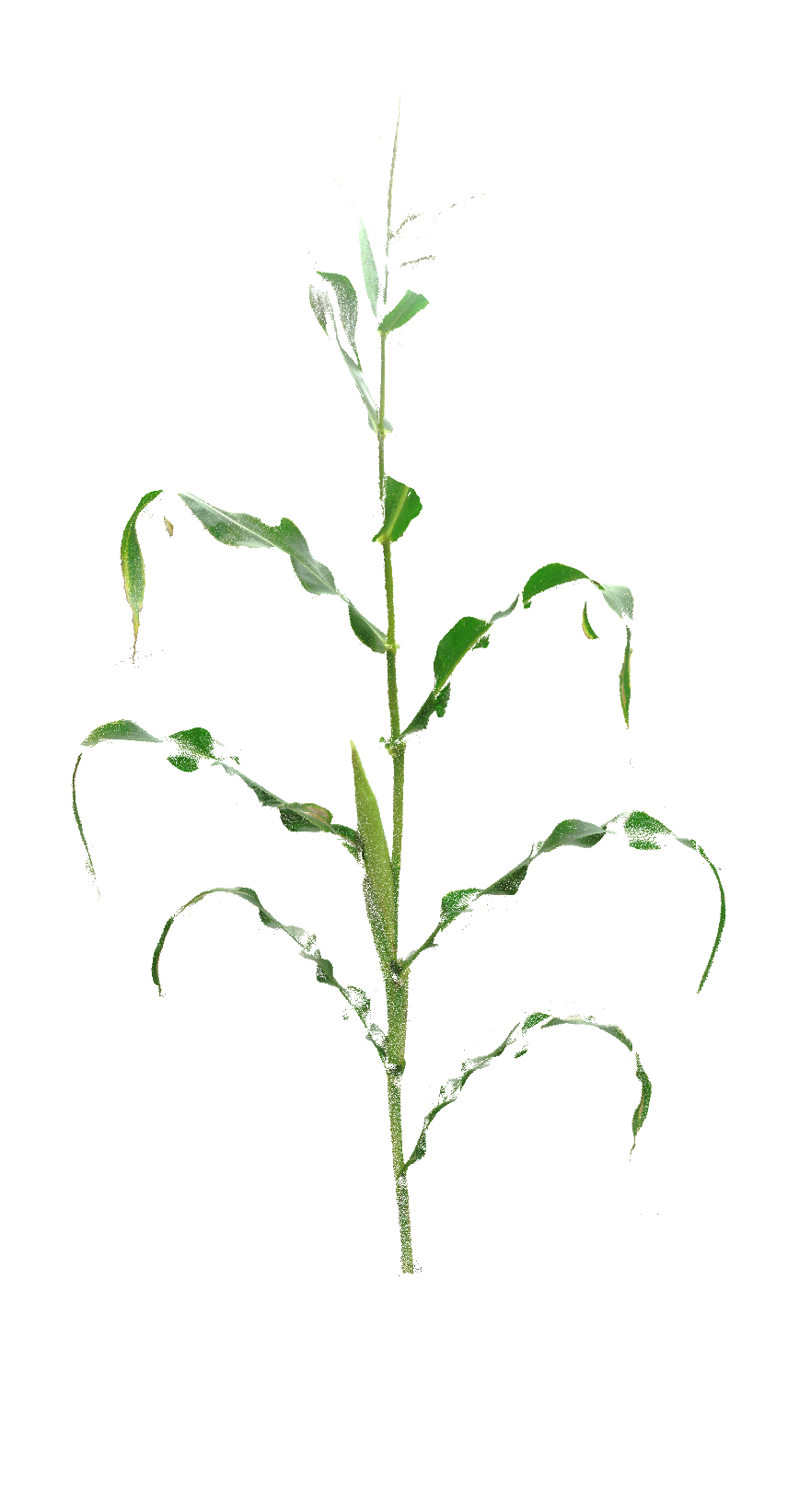} &
    \includegraphics[trim=2cm 6cm 2cm 5cm, clip, width=0.12\linewidth]{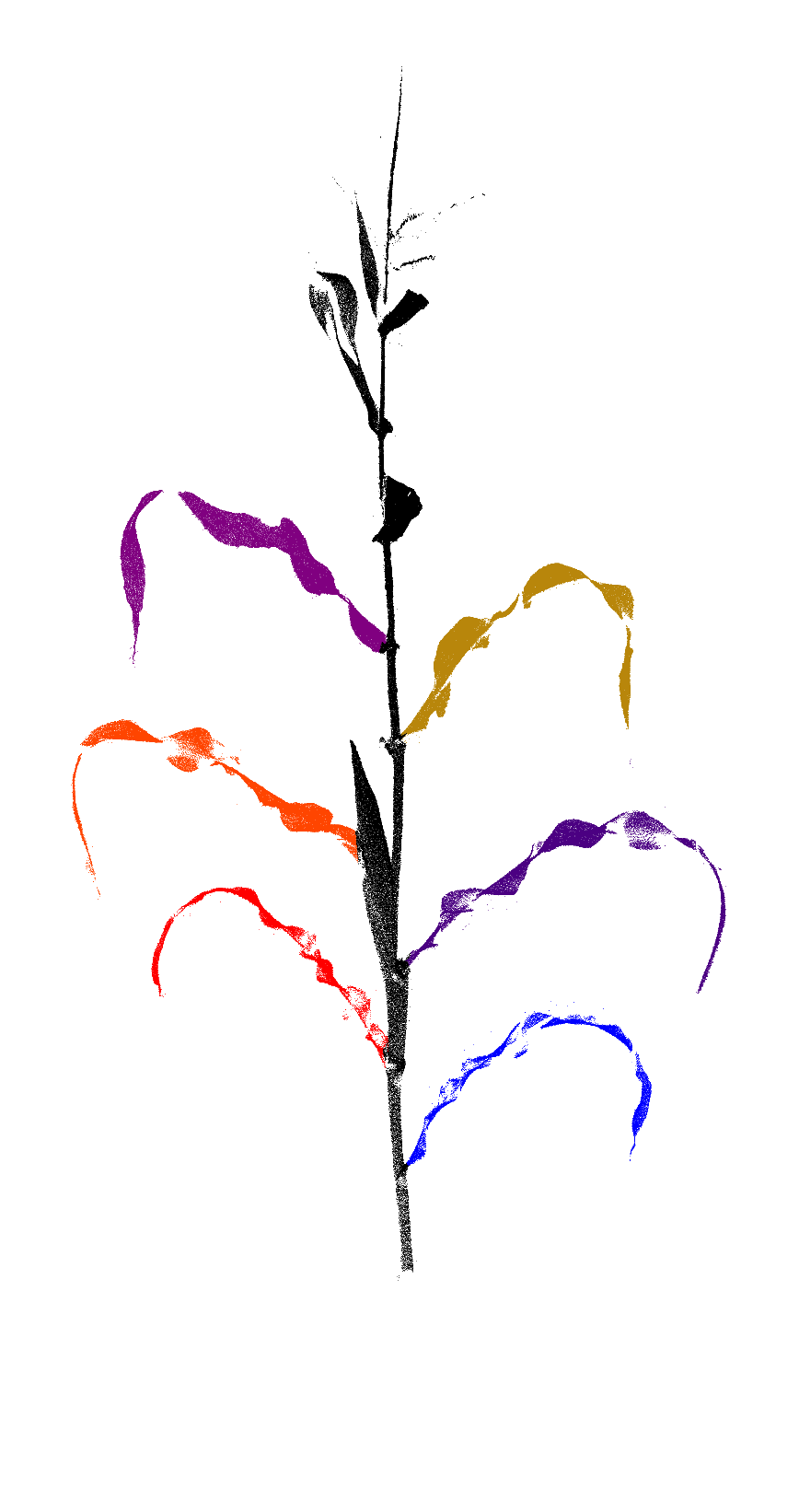} &
    \includegraphics[trim=2cm 6cm 2cm 5cm, clip, width=0.12\linewidth]{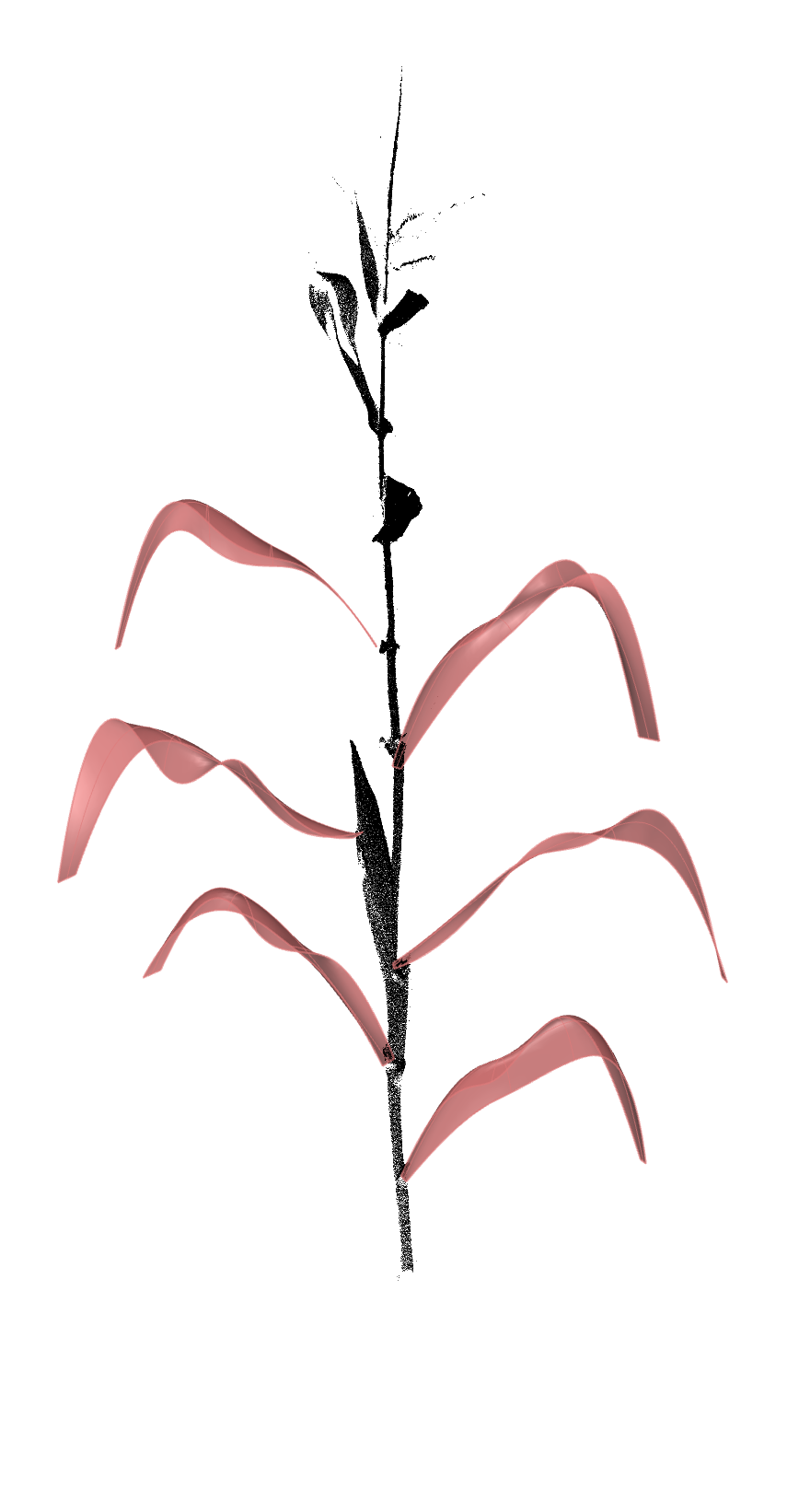} &
    \includegraphics[trim=3cm 4.5cm 2.5cm 3cm, clip, width=0.12\linewidth]{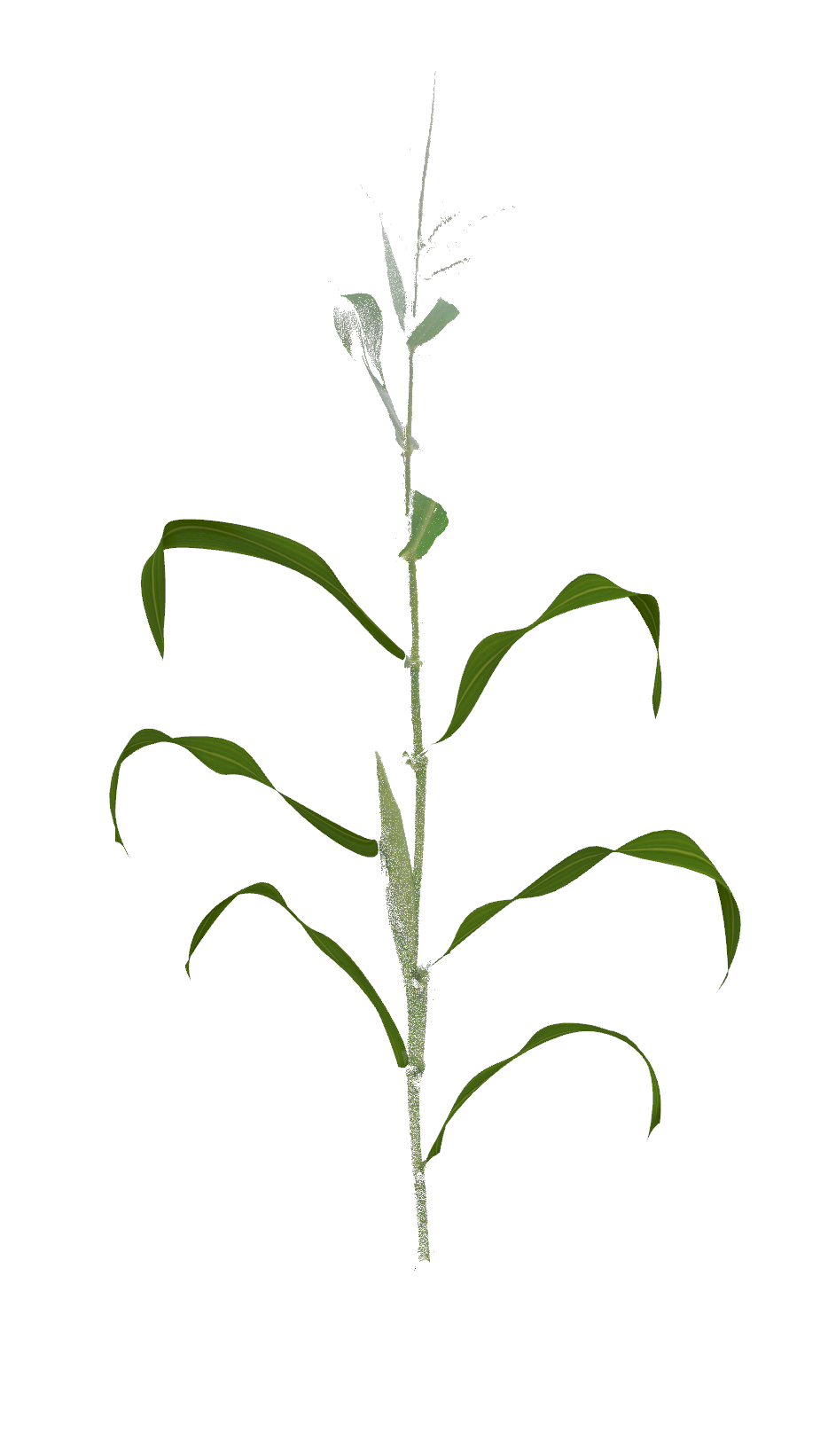} \\
    \hline
    A679 &  
    \includegraphics[trim=0cm 6cm 0cm 6cm, clip, width=0.12\linewidth]{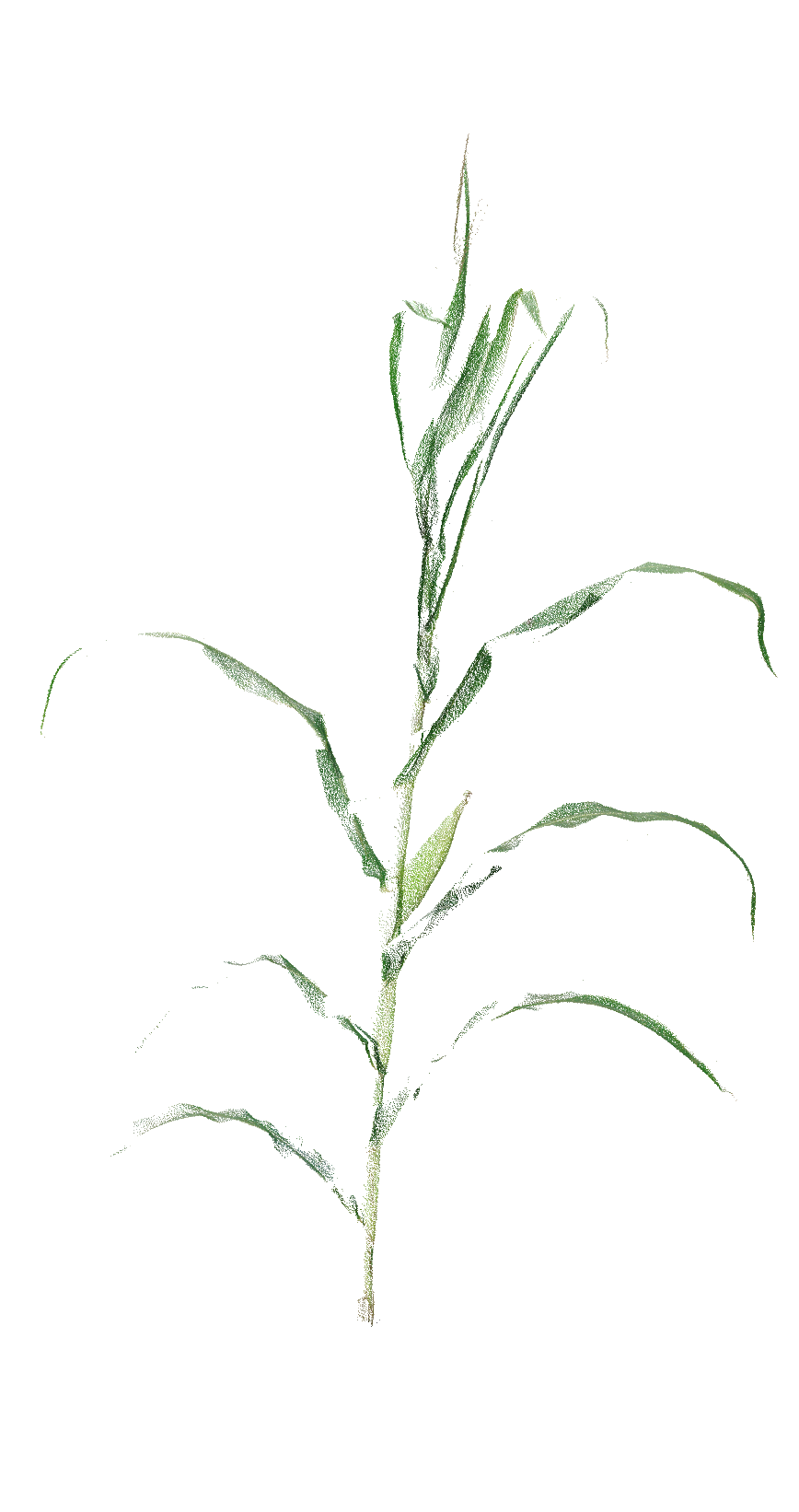} &
    \includegraphics[trim=0cm 6cm 0cm 6cm, clip, width=0.12\linewidth]{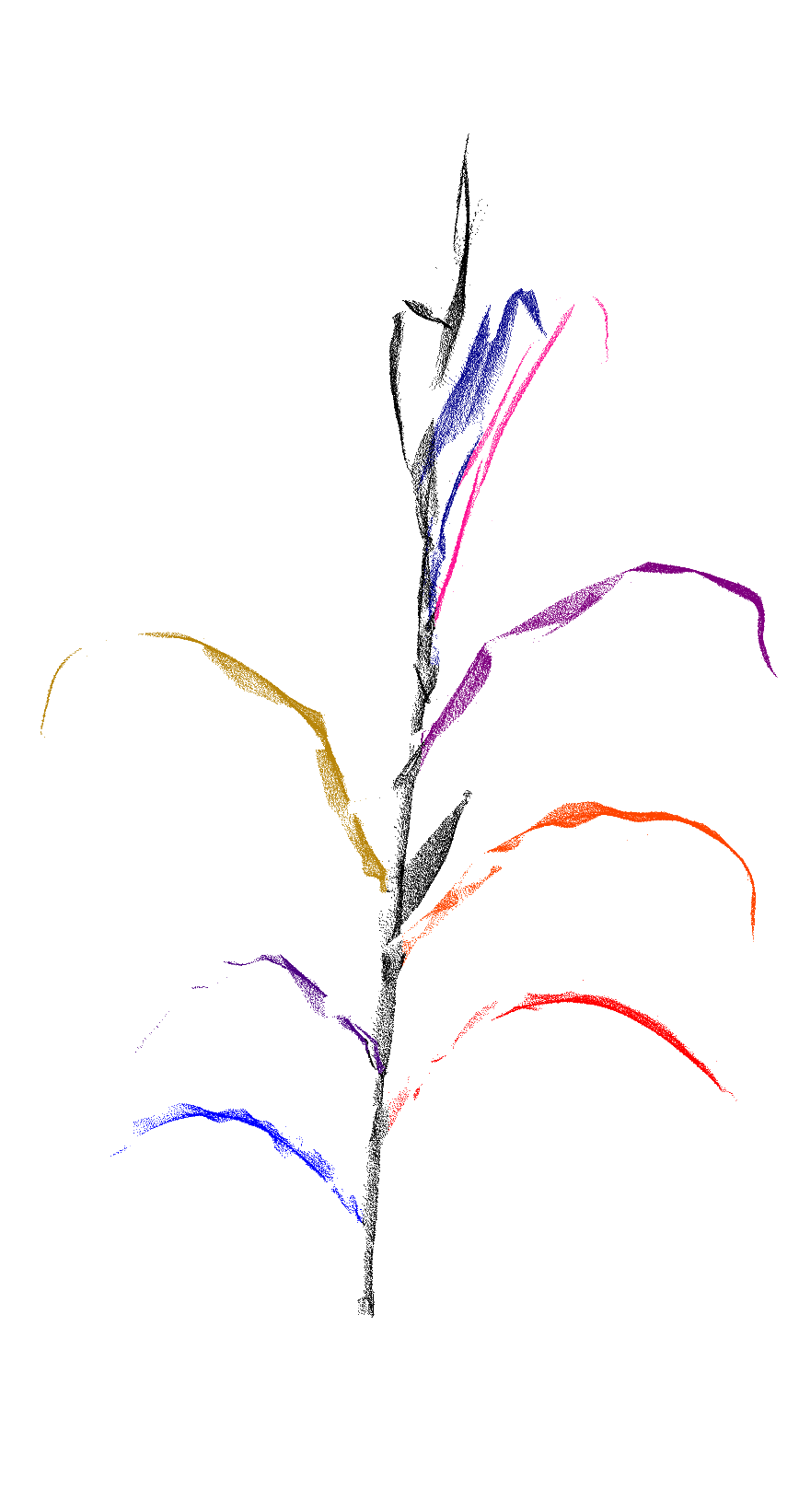} &
    \includegraphics[trim=0cm 6cm 0cm 6cm, clip, width=0.12\linewidth]{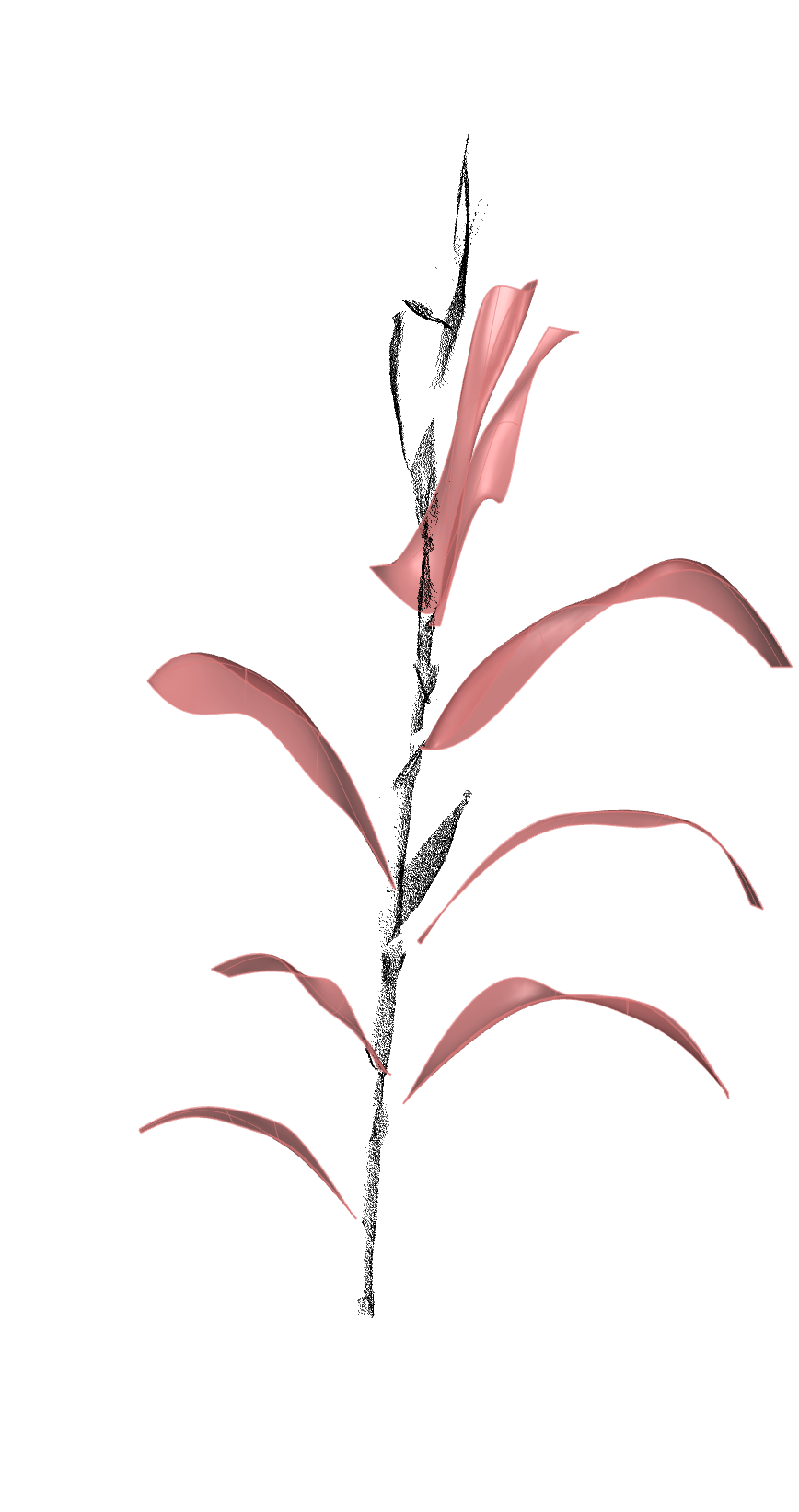} &
    \includegraphics[trim=1.5cm 7cm 1.5cm 5cm, clip, width=0.12\linewidth]{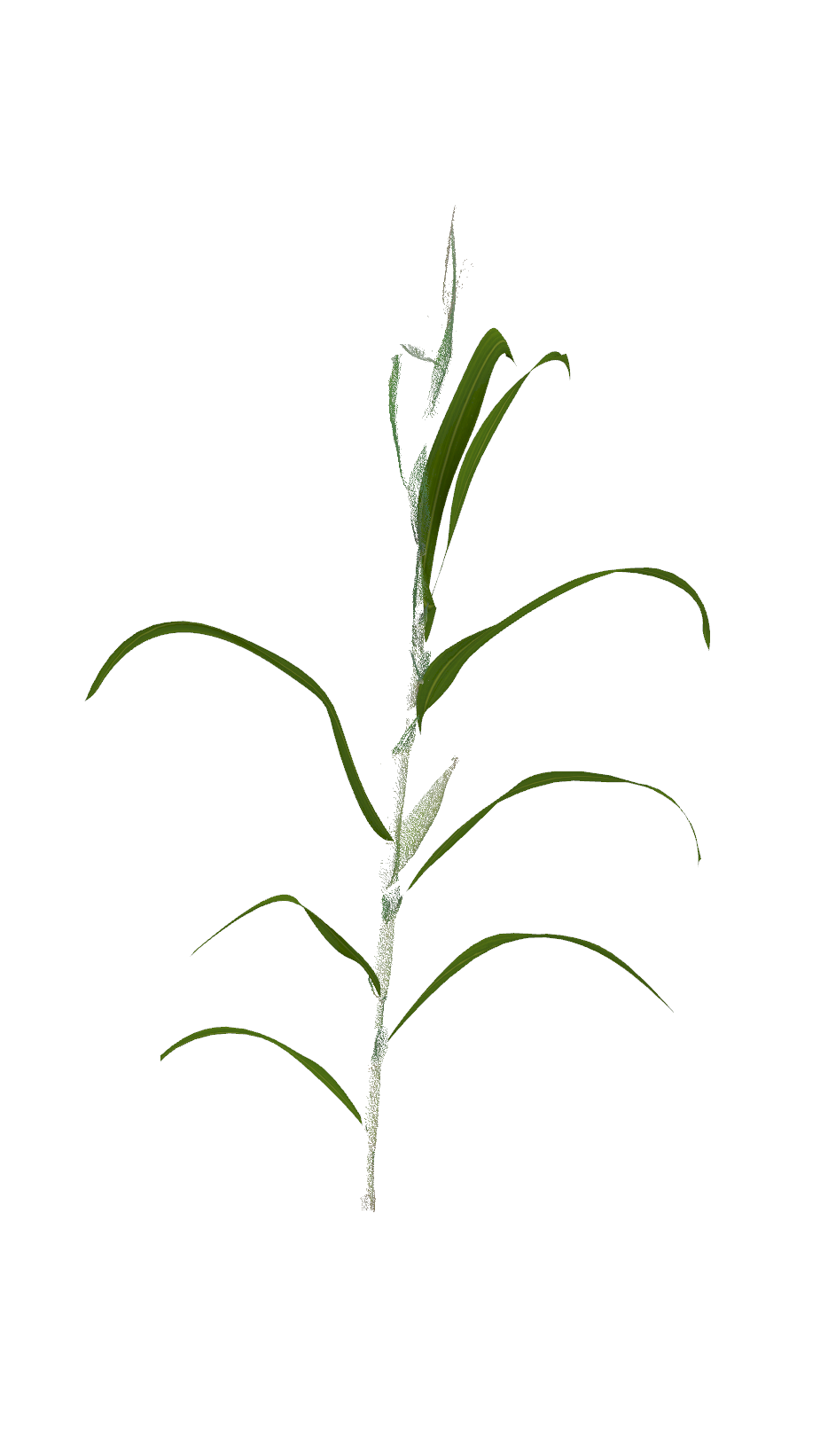} \\
    \hline
    CI90C &  
    \includegraphics[trim=2cm 6cm -1cm 7cm, clip, width=0.12\linewidth]{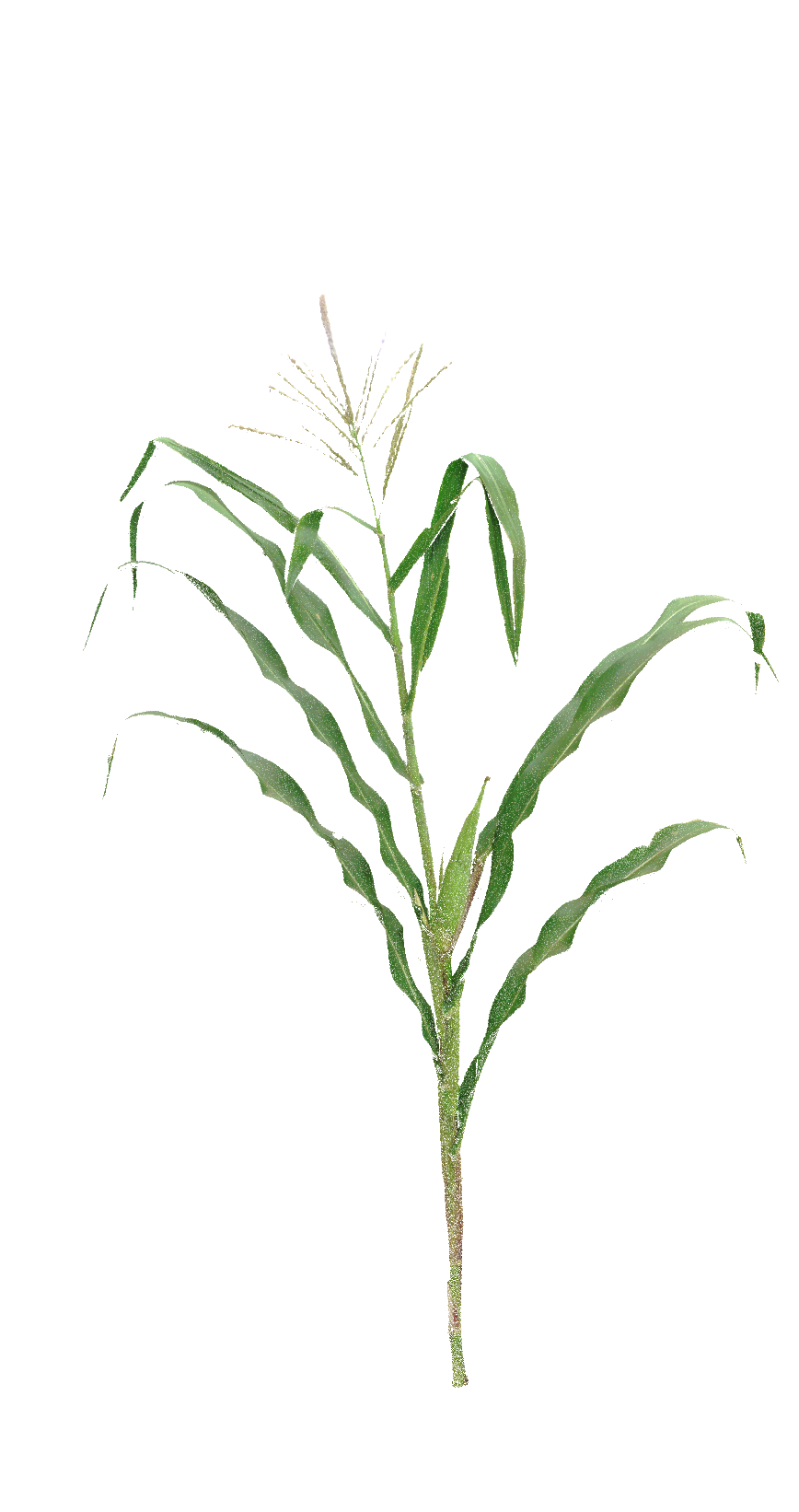} &
    \includegraphics[trim=2cm 6cm -1cm 7cm, clip, width=0.12\linewidth]{Figures_M/Plants_steps/CI90C_color.png} &
    \includegraphics[trim=2cm 6cm -1cm 7cm, clip, width=0.12\linewidth]{Figures_M/Plants_steps/CI90C_pso.png} &
    \includegraphics[trim=1cm 5cm 1cm 7cm, clip, width=0.12\linewidth]{Figures_M/Plants_steps/CI90C_nurbs1.png} \\
    \hline
    T8 &  
    \includegraphics[trim=2cm 10cm 2cm 5cm, clip, width=0.12\linewidth]{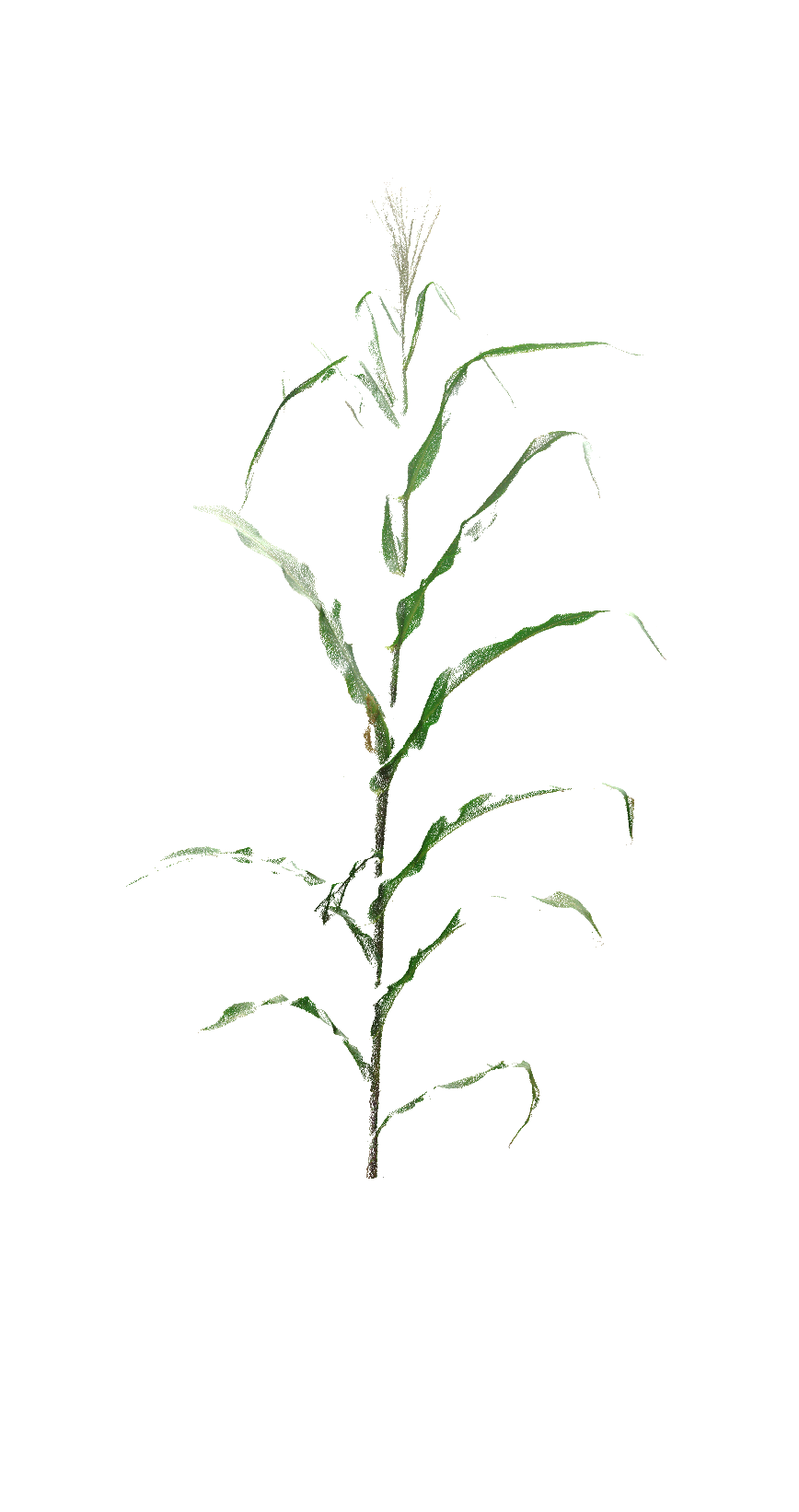} &
    \includegraphics[trim=2cm 10cm 2cm 5cm, clip, width=0.12\linewidth]{Figures_M/Plants_steps/T8_color.png} &
    \includegraphics[trim=2cm 10cm 2cm 5cm, clip, width=0.12\linewidth]{Figures_M/Plants_steps/T8_pso.png} &
    \includegraphics[trim=0cm 4cm 0cm 2cm, clip, width=0.12\linewidth]{Figures_M/Plants_steps/T8_nurbs1.png} \\
    \hline
\end{longtable}

\begin{figure}[t!]
    \centering
    \includegraphics[width=6in, height=3.5in,]{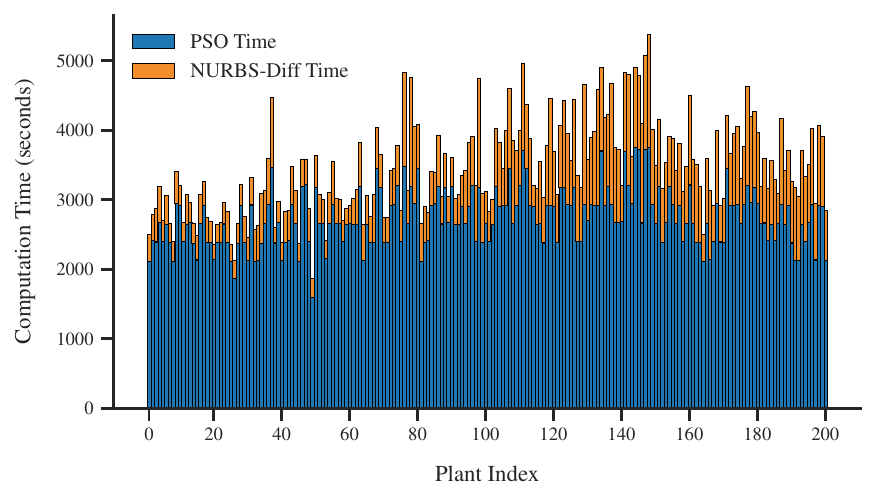}
    \caption{Total computational time for 200 maize plants. The plot shows the breakdown of computational time between PSO (in blue) and NURBS-Diff (in red). While PSO contributes to the majority of the runtime, NURBS-Diff adds a smaller, yet significant portion for achieving high-fidelity surface fitting.}
    \label{fig:computation_time}
\end{figure}

\end{document}